\documentclass[11pt]{article}

\usepackage[margin=1in]{geometry}
\usepackage{microtype}
\usepackage{needspace}

\usepackage{amsmath,amssymb,amsthm}
\usepackage{mathtools}
\usepackage{enumitem}

\usepackage[numbers,sort&compress]{natbib}
\usepackage{url}
\usepackage[hidelinks]{hyperref}

\newtheorem{theorem}{Theorem}
\newtheorem{lemma}[theorem]{Lemma}
\newtheorem{corollary}[theorem]{Corollary}
\newtheorem{proposition}[theorem]{Proposition}
\newtheorem{conjecture}{Conjecture}

\theoremstyle{definition}
\newtheorem{definition}{Definition}
\newtheorem{principle}{Principle}
\newtheorem{assumption}{Assumption}

\theoremstyle{remark}
\newtheorem{remark}{Remark}

\title{Predictive gains, causal losses:\\
why interventional competence cannot\\
be learned from data alone}

\author{
Marcus A. Thomas\thanks{Corresponding author. Email: \texttt{thomm15@mskcc.org}.}\\
Memorial Sloan Kettering Cancer Center\\
321 East 61st Street\\
New York, NY 10065, USA
}

\date{}

\begin{document}

\maketitle

\begin{abstract}

Retaining predictive gains can require persistent errors in a model's response to interventions. For deterministic predictors with finite second moments under squared-error loss, any observational advantage over the best predictor using only an outcome's direct causes requires dependence on other inputs, regardless of the learning rule. When admissible interventions vary these inputs while preserving the worldly outcome, this dependence yields a positive lower bound on root-mean-square intervention-response error. We identify conditions under which population gradient updates introduce this dependence from a parent-only baseline or damage a previously correct mechanism through shared parameters while improving other predictions.

These dependencies can also make model behavior vary with how the same intervention is realized. We characterize interventional alignment through a common output mapping across internal states and physical processes realizing the same intervention target and relevant context.

Correction is limited by the causal hypotheses currently admitted within inquiry. The relevant hypothesis class comprises these causal alternatives, rather than the function class the computational system could in principle represent. For predictions restricted to these alternatives, interventional mean error cannot fall below the best approximation they permit, whether evidence is observational, interventional, or both. Reducing error below a positive limit requires admitting better hypotheses, which may entail revising the causal variables, mechanism structure, or intervention interpretation. Interventional criticism constrains replacements without determining them. We conjecture that reliably acquiring and maintaining exact interventional competence when current causal commitments can fail requires autonomous participation in proposing alternatives, testing their consequences, and revising them in response to error.
\end{abstract}

\noindent\textbf{Keywords:} Causal reasoning; observational learning; interventional competence; no-go theorem; learning dynamics; causal mechanisms; hypothesis revision.

\section{Introduction}
\label{sec:intro}

Learning systems trained to predict observational data are often used to guide experiments, plans, and their own actions. These uses require predictions about interventions, including changes absent from training. A relationship that helps predict an outcome in the training data may produce an erroneous prediction when an intervention changes how an input is generated. The question is whether the learned computation respects the causal mechanism it is taken to represent.

A causal mechanism specifies both what can affect an outcome and what should leave it unchanged. These exclusions can be stipulated in a causal representation without being tested in the learned computation. Predictive success need not test them either. Our argument examines what predictive improvement and interventional evidence can establish about this computational requirement, and what correcting its failures requires.

Section~\ref{sec:setup} assesses interventional competence through correct predictions and agreement across realizations of the same intervention and across interventions preserving the world's outcome. It distinguishes an exact benchmark from practical adequacy. Section~\ref{sec:results} then shows why observational improvement need not secure that competence. Under specified conditions, gradient updates introduce dependence on inputs whose variation leaves the outcome unchanged, or damage a previously correct mechanism through shared parameters. For deterministic predictors with finite second moments under squared-error loss, retaining an advantage over the best predictor using only the outcome's direct causes requires retaining dependence on other inputs, independently of the learning rule. Further optimization therefore need not remove the dependency while preserving the gain.

These dependencies can affect predictions of the system's own actions. When the compared model runs supply the same parent inputs, we identify conditions under which policies preserving the worldly outcome produce different model outputs. When subsequent computations retain a specified minimum fraction of earlier deviations in expectation, we bound their cumulative discrepancy.

Access to interventional evidence does not by itself remove the remaining limits. Section~\ref{sec:instantiation-revision} distinguishes the causal alternatives currently admitted within inquiry from the mappings the computational system could represent. Predictions restricted to those alternatives cannot reduce error in interventional means below their best achievable approximation error, whether the evidence is observational, interventional, or both. Reducing error below a positive gap requires admitting better hypotheses, which can occur through changes in computational state with architecture and learned parameters held fixed.

Section~\ref{sec:scope} examines what interventional criticism contributes to that revision. When candidates predict different consequences, interventional tests can discriminate among them. Eliminating candidates cannot lower the class's approximation gap or supply hypotheses beyond it. These limits motivate, but do not establish, our conjecture that reliably acquiring and maintaining exact interventional competence when current causal commitments can fail requires autonomous participation in proposing alternatives, testing their consequences, and revising them in response to error.

\section{Relationship to Existing Approaches}
\label{sec:related}

{
Mechanism locality, formalized in Section~\ref{sec:lap}, occupies an unusual position between causal modeling and predictive learning. In a structural causal model, an equation specifies an outcome through its causal parents and exogenous conditions. Exclusion of other inputs is built into that representation, allowing identification, estimation, and intervention analysis to proceed with locality already stipulated. Predictive learning starts elsewhere, assessing computations by their predictive accuracy. Dependence on additional inputs can improve that accuracy without being assessed against causal exclusions. Locality can therefore remain outside explicit investigation in both settings, as a premise of the causal representation or as a property that predictive evaluation need not test.

We investigate locality as a property of the learned computation itself, asking whether it respects the exclusions of the mechanism it is taken to represent and what observational learning does to that requirement.

\paragraph{Causal identification and representation learning.}
The Causal Hierarchy Theorem \citep{bareinboim2022pch} establishes that observational data generically underdetermines interventional and counterfactual quantities. Structural assumptions and interventional evidence can resolve particular ambiguities. Xia et al.~\citep{xia2021causalneural} show that neural expressiveness does not remove this limitation. Their graph-constrained neural causal models preserve query identifiability from the graph and observational distribution, permitting estimation when identifiable. A class containing the true model still leaves a query unidentified when observationally consistent members disagree on it.

Causal representation learning seeks causal variables and relations in low-level observations \citep{scholkopf2021crl}. Under specified support and mixing assumptions, Ahuja et al.~\citep{ahuja2023interventional} identify latent factors without known intervention targets or structural assumptions on the latent dependency graph. CITRIS \citep{lippe2022citris} identifies multidimensional causal factors from temporal sequences with observed intervention targets under its assumptions. Brehmer et al.~\citep{brehmer2022weakly} establish identifiability using paired pre- and post-intervention observations with unknown targets, perfect interventions, and their other stated conditions. Ke et al.~\citep{ke2019learning} combine observational and interventional data with unknown targets for structure discovery.

\paragraph{Invariance across environments and learning dynamics.}
Invariant causal prediction \citep{peters2016icp} and invariant risk minimization \citep{arjovsky2019irm} seek invariant structure across environments, which may include interventions. The former tests invariance to constrain causal predictors under specified assumptions. The latter seeks representations supporting a common optimal predictor across training environments. Guarantees depend on how training environments relate to evaluation changes. Rosenfeld et al.~\citep{rosenfeld2021risks} exhibit failures of IRM to recover invariant predictors or generalize under distribution shift, including nonlinear settings.

Under Theorem~\ref{thm:obstruction}'s hypotheses, observational loss can decrease while interventional error increases. Extending the analysis to experimental training evidence requires establishing the corresponding loss, sensitivity, and causal conditions.

\paragraph{Causal abstraction and structural realization.}
Formal causal abstraction relates causal models at different levels of description by requiring consistency between corresponding interventions \citep{Rubenstein2017,BeckersHalpern2019}. Work on neural causal abstraction uses interchange interventions to investigate these relationships in learned computations \citep{Geiger2021,Geiger2025}. \citet{richens2024robust} recover an approximate causal model from low-regret policies across mixtures of local interventions under their stated assumptions. Their criterion concerns the causal information recoverable from policies. Our alignment condition concerns agreement of post-intervention outputs across internal realizations of the same target and relevant context at fixed system randomness.

Proposition~\ref{prop:realizability-gap} separately bounds correction within a hypothesis class with a positive interventional approximation gap.
}

\section{Formal Setup}
\label{sec:setup}

\subsection{Parametric mechanisms}

Different computational procedures can realize the same input--output mapping. For example, distinct algorithms can compute the sum of two numbers. Specifying the mapping therefore does not specify the procedure used to compute it. We distinguish a computational mechanism from the input--output mapping its execution induces. Calling a procedure a mechanism does not establish that it represents a causal process in the world. A mechanism attribution proposes how the computation is organized and, where applicable, which worldly causal process it represents.

{
\begin{definition}[Parametric Mechanism]
\label{def:mechanism}
A parametric computational mechanism $\mu_i$ associated with variable $X_i$ is an organized computational procedure with specified operations and dependencies. Given computational inputs, exogenous conditions, and mechanism parameters, its execution induces a mapping
\[
\mathcal{M}_i : \mathcal{X}_{\mathrm{In}(i)} \times \mathcal{U}_i
\times \Theta_i \to \mathcal{X}_i,
\]
where $\mathrm{In}(i)$ indexes the variables supplied to the computation, $\mathcal{U}_i$ represents exogenous conditions, $\Theta_i$ is the parameter space, and $\mathcal{X}_i$ is the output state space. Parameters may be shared with, or associated with, other mechanisms. We call $\mathcal M_i$ the mechanism-induced mapping. Distinct computational mechanisms may induce the same mapping.
\end{definition}

We reserve $\mathrm{PA}(i)$ for the parents of $X_i$ in the true causal
graph. The computational inputs $\mathrm{In}(i)$ need not coincide with
these parents. A learned mechanism can therefore use an input that helps
predict $X_i$ without being one of its direct causes.
}

\subsection{The Locality--Autonomy Principle}
\label{sec:lap}

Replacing one mechanism has a determinate effect only against a specified background of other mechanisms. The Locality--Autonomy Principle states which dependencies and mechanism rules must remain unchanged for the interventions being modeled.

A \emph{task representation} specifies the causal variables, their proposed mechanism structure, and how physical processes modify those mechanisms.

\begin{principle}[Locality--Autonomy Principle (LAP)]
\label{princ:LAP}

The Locality--Autonomy Principle specifies the dependencies and mechanism
invariances required by a causal representation of interventions.
\emph{Locality} requires that,
for a fixed mechanism and exogenous conditions, its output depends only
on the states of its causal parents. Varying other variable values at
fixed parent values leaves that output unchanged.

\emph{Autonomy}
requires that an intervention, carried out through a specified physical process, preserve the rules of mechanisms outside the changes admitted
by the causal representation. The representation must specify both the
mechanisms the process modifies directly and any further changes
permitted by legitimate dependencies between mechanisms. The remaining
rules provide the stable background against which the intervention is
interpreted; their outputs may nevertheless change through changes in
their causal inputs. A surgical intervention, as in Pearl's $\mathrm{do}$-operator, is the special case in which only the targeted mechanism is replaced and all other rules remain unchanged.

\end{principle}

{

This role of invariance is central to Pearl's structural causal models
and Woodward's interventionist account \citep{pearl2000models, woodward2003making}.

We use \emph{locality and autonomy} (L\&A) to assess whether the learning system's computational mechanisms respect the specified causal structure, reserving LAP for the domain. Computational locality requires the mechanism-induced mapping to ignore variation in inputs in $\mathrm{In}(i)\setminus\mathrm{PA}(i)$ at fixed parent inputs, mechanism parameters, and exogenous values. Computational autonomy requires a change representing an intervention on one mechanism to preserve the induced rules of mechanisms that the causal structure requires to remain unaffected. An induced rule specifies outputs as a function of legitimate inputs and exogenous conditions at fixed mechanism parameters. Comparing outputs under these conditions can witness a failure of rule invariance.

For example, entering ``\texttt{7 + 7; <enter>}'' into a calculator can select a particular addition algorithm and supply its two numerical inputs. In a programmable system, quantities supplied during operation can also specify the rule to be executed. Mechanism parameters therefore need not be learned weights: they may arrive through inputs, be computed internally, or be retained from earlier computation, in explicit or distributed form. Whether a quantity functions as an input to a rule or as a parameter specifying it depends on its role in the mechanism attribution, rather than where it is stored.

An intervention claim also concerns the consequences of a physically possible change to a system. A coherent replacement of equations alone does not establish that the change is physically possible. We therefore require interventions to have physically realizable processes, without requiring present technological feasibility or human control. This requirement is stronger than Woodward's allowance for scientifically grounded intervention counterfactuals whose antecedents may be physically impossible \citep{woodward2003making}.

Different physical processes classified as realizations of the same
intervention may affect other mechanisms
differently. When all admissible realizations nevertheless yield the
same worldly outcome in the relevant context, world realizer invariance
holds (Assumption~\ref{asm:world-ri}), and that common outcome supplies
the surgical target. Otherwise, the target must specify how the
intervention is carried out. Such dependence on realization can reveal
an inadequate task representation without implying that the world's
underlying mechanisms violate LAP.

Assessing a locality or autonomy violation requires a specified mechanism
attribution and intervention correspondence, held fixed during the
comparison.
The mechanism representation fixes the level and interpretation of an
L\&A assessment. When its variables carry attributed causal content,
tests of its dependencies and invariances assess whether the system
instantiates that interpreted mechanism.

When the mechanism-induced mappings are differentiable with respect to the quantities varied by an intervention, invariance can be tested along an intervention flow: a path through state or mechanism-parameter space generated by the specified intervention. If the intervention should leave the assessed output unaffected, that output must remain unchanged along the flow, independently of the coordinates used to describe it.

The \emph{Lie derivative} measures the rate of output change along this flow. For a scalar output $g$ and an intervention field $V$ generating the local flow $\varphi_t^V$, it is
$\mathcal{L}_V g(q)
:= \left.\frac{d}{dt}g(\varphi_t^V(q))\right|_{t=0}$.
In local coordinates this becomes
$\mathcal{L}_V g = \sum_a V^a\,\partial_a g$,
the directional derivative along the intervention field.
The geometric specification remains meaningful when parameters are shared and no canonical mechanism-specific blocks are available, provided the intervention field is specified.

Here $\mathrm{PA}(i)$ denotes the true causal parents of the state
variable $X_i$. By contrast, $\mathrm{MPA}^*(i)$ concerns dependence on
mechanism parameters: ``M'' refers to mechanisms, and the star marks the
worldly relation against which the model is assessed. It includes $i$
itself by convention, together with any other mechanisms whose parameter
settings legitimately influence mechanism $i$.

In this parametric setup, let
$W_j := \partial/\partial X_j$ vary input $X_j$ with the mechanism
specification fixed, and let $\Xi_A$ be a specified field implementing
a change to mechanism $A$ through the quantities representing its
parameters. Computational locality and autonomy respectively require

\[
\mathcal{L}_{W_j}\mathcal{M}_i = 0
\;\; (j \in \mathrm{In}(i)\setminus\mathrm{PA}(i)),
\qquad
\mathcal{L}_{\Xi_A}\mathcal{M}_i = 0
\;\; (A \notin \mathrm{MPA}^*(i)).
\]

These conditions test whether the mechanism-induced mapping preserves the required output invariances along the specified intervention directions. A nonzero Lie derivative witnesses a computational locality or autonomy
violation under the respective comparisons specified above. For autonomy,
the perturbation direction must implement a specified intervention on
one mechanism, while the assessed mechanism's rule is required to remain
unchanged at fixed legitimate inputs and exogenous values. Sensitivity
to learned weights alone does not establish this. The underlying
invariance requirements apply to both differentiable and
nondifferentiable mechanisms.

In the ordinary modular case, $\mathrm{MPA}^*(i)=\{i\}$, so autonomy
requires invariance under changes to other mechanisms' parameters.
In mechanised causal models, the set may also include other mechanisms
\citep{kenton2023discovering}. Their parameters
$\theta_{\mathrm{MPA}^*(i)}$ belong to the target description rather
than to the variation across realizations that must leave it unchanged.
The diagnostic concerns sensitivity to parameters outside this set.
}

{
\subsection{Realizers and levels of invariance}
\label{sec:structural-objects}

Testing invariance requires specifying what may vary while the intervention target and relevant context remain fixed. Relative to the task representation, we distinguish variation in the internal state representing them from variation in the physical process that sets the target variable.

\begin{definition}[Constitutive and process realizers]
\label{def:realizers}
Fix a task representation, task variables $(X,Y)$, a value $x$, and a relevant context $c$.
\begin{enumerate}[leftmargin=*,itemsep=2pt]
\item[(i)] A \emph{constitutive realizer} of $(x,c)$ is an internal state
$z$ in the learned system's state space $Z$ with $\pi_X(z) = x$ and $\pi_C(z) = c$, where $\pi_X$ and $\pi_C$
project internal states onto the task variable and onto the relevant context.
The constitutive realizers of $(x,c)$ form the fiber $F_{x,c}$.
\item[(ii)] 

A \emph{process realizer} $R$ of $\mathrm{do}(X{=}x)$ is a physical process that sets $X$ to $x$. It is represented as a modification of the tuple of mechanisms and written $\mathrm{do}_R(X{=}x)$.
Relative to the task representation, it is \emph{surgical} if it replaces only the mechanism for $X$ and has a \emph{fat hand} \citep{scheines2005similarity, baumgartner2016constitutive} if it modifies others.

A process realizer is \emph{admissible} relative to the task representation if it is physically available and satisfies the following graph condition. For each mechanism other than $X$'s that the process is represented as modifying, every directed path to $Y$ must pass through $X$.
\end{enumerate}
\end{definition}

Admissibility does not guarantee agreement of actual outcomes;
that agreement is the separate condition of world realizer invariance.
Additionally, invariance across internal states does not ensure
invariance across physical processes, or conversely. Both kinds of variation can occur
for externally specified or system-generated interventions.

Relevant context includes features of the initial situation whose
influence on the outcome is legitimate for the task. Starting states
may nevertheless differ, including in the initial value of the variable
being set. Outputs are compared after the intervention sets that
variable to the target value.
All world-side comparisons at context $c$ range over exogenous
values compatible with that context. World-side expectations use
the corresponding contextual law.

Equal mean outputs need not imply equal distributions, and equal
distributions need not imply equal outputs at fixed randomness.
We therefore distinguish mean, distributional, and pathwise invariance.

\begin{definition}[Levels of invariance]
\label{def:inv-hierarchy}
Let $\mathcal{R}$ be a family of realizers and, for $R \in \mathcal{R}$, let $Y_R(u)$ denote the outcome under $R$ at exogenous value $u$, where $u\in\mathcal{U}$ is a realization of a random variable $U$ with law $P_U$. The outcome is
invariant across $\mathcal{R}$
\begin{enumerate}[leftmargin=*,itemsep=2pt]
\item[(i)] \emph{pathwise} if $Y_R(u) = Y_{R'}(u)$ for all
$R, R' \in \mathcal{R}$ and all $u$;
\item[(ii)] \emph{distributionally} if $(Y_R)_* P_U = (Y_{R'})_* P_U$ for all
$R, R' \in \mathcal{R}$;
\item[(iii)] \emph{in mean} if $\mathbb{E}[Y_R] = \mathbb{E}[Y_{R'}]$ for all
$R, R' \in \mathcal{R}$.
\end{enumerate}
\end{definition}

Holding exogenous conditions fixed requires specifying those conditions and how they enter the mechanism. Different mechanisms can produce the same output distribution while assigning different outputs to the same fixed conditions. We therefore fix the mechanisms and exogenous variables used in pathwise comparisons.

\begin{definition}[Privileged representation]
\label{def:privileged}

For the world, the privileged representation consists of the LAP-level mechanisms $f_i^*$ and their exogenous conditions $U^*$. The law $P_{U^*}$ describes variation in those conditions across the cases under consideration. For a learned system, the privileged representation consists of its actual computational implementation, described by the state-transition map $\Phi$, and the exogenous conditions $U_{\mathrm{sys}}$ entering its execution, including those used by sampling procedures. Probability laws describe variation in these conditions and the resulting outputs; they are not additional causal mechanisms. All pathwise claims in this paper are relative to these representations.
\end{definition}

The task representation specifies the variables and process classifications being assessed. The privileged representation fixes the mechanisms and sources of randomness used for pathwise comparisons. Distinctions omitted by the task representation may still affect outcomes in the world.

\begin{lemma}
\label{lem:hierarchy}
Pathwise invariance implies distributional invariance, which implies
invariance in mean whenever the means exist. Equivalently, failure in mean
implies failure in distribution, which implies failure pathwise.
\end{lemma}

\emph{Proof.} If the outcome is invariant pathwise, each realizer induces the same map on the exogenous space, so that the pushforward distributions coincide. If these distributions have finite first moments, their means coincide as well. \hfill$\square$
}

\subsection{Observational and interventional competence}

{
Observational competence concerns prediction under the observed regime.
For an intervention specified by its target and relevant context alone,
structural interventional competence additionally requires a common
worldly target, agreement across the model's admissible realizations,
and correct predictions of that target.

\begin{assumption}[World realizer invariance]
\label{asm:world-ri}
Fix a task representation and its admissible process realizers (Definition~\ref{def:realizers}), with at least one available realizer for each evaluated target and context $(x,c)$. For every such $(x,c)$, every admissible process realizer $R$ of $\mathrm{do}(X{=}x)$, and every
exogenous value $u^*$ compatible with $c$ in the world's
privileged representation, the actual worldly outcome $Y^{*}_{R}(x,c,u^*)$ does not depend
on $R$. 

This common outcome defines the surgical target. Its expectation under the contextual law of the world's exogenous variables is $\mathbb{E}[Y \mid \mathrm{do}(X{=}x), c]$.
\end{assumption}

This assumption requires the task representation's grouping of physical processes to respect the world's response. Processes classified as realizers of the same intervention may nevertheless produce different outcomes because the representation omits causally relevant distinctions. For example, equal total cholesterol levels may have different consequences depending on their composition \citep{spirtes2004ambiguous}. The assumption is closely related to the condition of no multiple versions of treatment \citep{vanderweele2013versions}. It is also a cross-world claim, since it compares outcomes under different
realizers at the same exogenous value.

With a complete, correct recursive task representation and correctly
specified process modifications, admissibility implies world realizer
invariance: fixing $X$ blocks all routes from other modified mechanisms
to $Y$. The separate assumption has substantive content when the
representation or process classifications may be incomplete or incorrect.

Exact correctness has a further consequence under
Assumption~\ref{asm:world-ri}: admissible realizers of the same
intervention share a worldly target, so predictions correct under
every realizer must agree in mean. Writing $f_R(x,c)$ for the
model's mean prediction under realizer $R$,
\[
f_R(x,c)
=
\mathbb E[Y\mid\mathrm{do}_R(X=x),c]
=
\mathbb E[Y\mid\mathrm{do}(X=x),c].
\]
This agreement does not by itself establish equality of output
distributions or structural alignment at fixed system randomness.

}

{
\paragraph{Internal alignment.}

Internal alignment compares outputs reached from different starting states and through different admissible processes realizing the same intervention target and relevant context. Only starting states and processes available within the evaluation domain enter the comparison. Outputs are compared either in distribution or at each fixed value of the system's randomness.

\begin{definition}[Interventional alignment]
\label{def:alignment}
Let $\Phi:Z\times \mathcal{U}_{\mathrm{sys}}\to Z$ be the state-transition map induced by the system's actual computational implementation, with system randomness $U_{\mathrm{sys}}$
taking values in $\mathcal{U}_{\mathrm{sys}}$. Define
\[
P_{\Phi}(\cdot\mid z)
:= (\Phi(z,\cdot))_*P_{U_{\mathrm{sys}}},
\]
the distribution of resulting internal states obtained by holding
$z$ fixed and propagating the system's randomness through $\Phi$.
Here $*$ denotes pushforward of a probability distribution.
Let $\pi_X$, $\pi_Y$, and $\pi_C$ project onto the intervention
target, output, and relevant context.

Let $O\subseteq Z$ be the set of starting states reachable under the
initialization and interaction regimes included in the evaluation.
For each target $x$, let $\mathcal{R}_x$ be the specified family of
admissible internal process realizers. Each $R\in\mathcal{R}_x$
induces a physically realizable intervention map
\[
I_x^R:D_x^R\to Z,
\qquad D_x^R\subseteq O,
\]
where $D_x^R$ consists of the starting states from which that process
is available within the evaluation domain.
The intervention map must set the target to $x$ while preserving
the relevant context: for every $z\in D_x^R$,
\[
\pi_X(I_x^R(z))=x,
\qquad
\pi_C(I_x^R(z))=\pi_C(z).
\]
Other components of the internal state may change.
Every evaluated pair $(x,c)$ must admit at least one such realizer
and starting state with context $c$.

For every evaluated $(x,c)$, compare all
$R_1,R_2\in\mathcal{R}_x$ and all $z_j\in D_x^{R_j}$ satisfying
$\pi_C(z_1)=\pi_C(z_2)=c$. No restriction is placed on the initial
values $\pi_X(z_1)$ and $\pi_X(z_2)$.

\begin{enumerate}[leftmargin=*,itemsep=2pt]
\item[(i)] The system satisfies \emph{distributional alignment}
if, for every such comparison,
\[
(\pi_Y)_*P_{\Phi}(\cdot\mid I_x^{R_1}(z_1))
=
(\pi_Y)_*P_{\Phi}(\cdot\mid I_x^{R_2}(z_2)).
\]

\item[(ii)] The system satisfies \emph{structural alignment}
if, for every such comparison and every
$u_{\mathrm{sys}}\in \mathcal{U}_{\mathrm{sys}}$,
\[
\pi_Y\bigl(\Phi(I_x^{R_1}(z_1),u_{\mathrm{sys}})\bigr)
=
\pi_Y\bigl(\Phi(I_x^{R_2}(z_2),u_{\mathrm{sys}})\bigr).
\]
\end{enumerate}
\end{definition}

An aligned model can still be wrong. Exact structural competence
requires correct interventional means, alignment across realizations,
and agreement across evaluated interventions that preserve the world's
outcome.
}

{
\begin{definition}[Exact Observational Competence]
\label{def:obs-comp}
Fix an observational distribution $P_{\mathrm{obs}}$, a nonnegative
loss $L$, and a nonempty reference class $\mathcal G$ of predictors
with finite risks
\[
R_{\mathrm{obs}}(g)
:= \mathbb E_{(x,y)\sim P_{\mathrm{obs}}}[L(g(x),y)].
\]
A predictor $f\in\mathcal G$ has \emph{exact observational competence}
relative to this specification if
\[
R_{\mathrm{obs}}(f)
=\inf_{g\in\mathcal G}R_{\mathrm{obs}}(g).
\]
It is \emph{observationally adequate at tolerance $\delta_{\mathrm{obs}}$}
if its excess risk is at most $\delta_{\mathrm{obs}}\geq0$.
Optimality within $\mathcal G$ does not establish the adequacy of
$\mathcal G$ itself.
\end{definition}

\paragraph{Example: two kinds of invariance.}
Consider a model predicting outdoor temperature after an intervention on a desk lamp's switch. Assume that neither the lamp's operation nor the admissible processes used to operate its switch affect outdoor temperature. First, compare a person setting the switch to ``on'' with a robot setting it to ``on.'' The first comparison concerns different realizations of the same target. Second, compare setting the switch to ``on'' with setting it to ``off.'' The second concerns different targets whose worldly outcomes agree when the weather and other background conditions are held fixed. In both cases, the model's outputs should agree. Both model comparisons hold system randomness fixed.

\begin{definition}[Exact Interventional Competence]
\label{def:int-comp}
Fix a task $(X,Y)$, observational distribution $P_{\mathrm{obs}}$,
and a domain $\mathfrak E_{\mathrm{int}}\subseteq\mathfrak E$ of
admissible intervention specifications containing interventions absent
from the observational regime. A specification $e$ may describe a
fixed-target intervention or a policy. Write $M^{*,e}$ for the world
under $e$, with integrable outcome $Y^{*,e}(c,u^*)$ in relevant context
$c$ at world exogenous value $u^*$.
Target-only specifications require world realizer invariance
(Assumption~\ref{asm:world-ri}). If the worldly outcome depends on how
the intervention is implemented, $e$ must specify that implementation.
World comparisons use the fixed privileged representation.

For every evaluated $(e,c)$, specify a nonempty family of eligible
implementations and attainable internal states, indexed jointly by $r$.
Let $\widehat Y_{e,r}(c,u_{\mathrm{sys}})$ denote the model's output
under realization $r$. All comparisons use the system's actual
implementation and common randomness from its privileged representation.

A model $f$ has \emph{exact mean-level interventional competence} if,
for every evaluated $(e,c)$,
\[
f(e,c)=\mathbb E[Y\mid M^{*,e},c].
\]
It has \emph{exact structural interventional competence} if, additionally,
its outputs are integrable and realize this predicted mean,
\[
\mathbb E_{U_{\mathrm{sys}}}
\bigl[\widehat Y_{e,r}(c,U_{\mathrm{sys}})\bigr]=f(e,c)
\]
for every evaluated $(e,c)$ and eligible $r$, and satisfy both
requirements below:
\begin{enumerate}[leftmargin=*,itemsep=2pt]
\item[(i)] \emph{Agreement across realizations.} For each $(e,c)$,
outputs agree across all eligible realizations at every fixed value
of the system's randomness:
\[
\widehat Y_{e,r}(c,u_{\mathrm{sys}})
=\widehat Y_{e,r'}(c,u_{\mathrm{sys}}).
\]
\item[(ii)] \emph{Preservation of worldly outcome invariance.}
For any evaluated $e,e'$ in the same relevant context $c$, if,
for every world exogenous value $u^*$ compatible with $c$,
\[
Y^{*,e}(c,u^*)=Y^{*,e'}(c,u^*),
\]
then, for every eligible realization $r$ of $(e,c)$ and $r'$ of
$(e',c)$ and every $u_{\mathrm{sys}}$,
\[
\widehat Y_{e,r}(c,u_{\mathrm{sys}})
=\widehat Y_{e',r'}(c,u_{\mathrm{sys}}).
\]
\end{enumerate}
Correct means imply invariant predictions across contexts with the same
worldly interventional mean. Equality of worldly means alone does not
trigger requirement (ii), which concerns equality at every world
exogenous value compatible with the relevant context.
Unless otherwise qualified, formal claims about interventional
competence refer to this exact structural notion.
\end{definition}

For $e=\mathrm{do}(X=x)$, write $f(x,c)$ for $f(e,c)$.
An eligible realization is a pair $r=(R,z)$ from
Definition~\ref{def:alignment}, with
\[
\widehat Y_{e,r}(c,u_{\mathrm{sys}})
=\pi_Y\bigl(\Phi(I_x^R(z),u_{\mathrm{sys}})\bigr).
\]
Requirement (i) is then structural alignment; requirement (ii) also
covers different target values that leave the world's outcome unchanged.
For a policy specification $e=\pi$, write $f(\pi,c)$ and $M^{*,\pi}$.
Endogenous competence applies the same definition to an evaluated family
$\Pi_{\mathrm{int}}$ of the system's intervention-generating policies.

\paragraph{Practical interventional adequacy.}
Practical interventional adequacy is assessed against specified accuracy
and invariance tolerances. For a real-valued task output, one uniform
criterion requires
\[
\left|
\mathbb E_{U_{\mathrm{sys}}}\widehat Y_{e,r}(c,U_{\mathrm{sys}})
-\mathbb E[Y\mid M^{*,e},c]
\right|\leq\delta
\]
for every evaluated $(e,c)$ and eligible $r$, and
\[
|\widehat Y_{e,r}(c,u_{\mathrm{sys}})
-\widehat Y_{e',r'}(c,u_{\mathrm{sys}})|\leq\tau
\]
for every comparison required by Definition~\ref{def:int-comp}(i)--(ii),
with $\delta,\tau\geq0$. These include realizations of the same
intervention and evaluated interventions whose worldly outcomes agree
at every fixed world exogenous value compatible with the context.
Zero tolerances recover exact
structural competence, taking $f(e,c)$ to be the common output mean.
Mean-level adequacy imposes only the accuracy requirement.
Averaged criteria require an explicit evaluation measure and do not
by themselves establish uniform adequacy.

}

\clearpage
\section{Main Results}
\label{sec:results}

We first characterize agreement across realizations of an intervention. We then identify conditions under which observational learning introduces or strengthens dependencies that undermine intervention predictions, examine their consequences for policy comparisons, and bound their effects through subsequent computation.

\subsection{Alignment across intervention realizations}

{

Interventional alignment holds exactly when the model's output
distribution, or its output at fixed system randomness, depends only on
the intervention target and relevant context across the admissible
realizations being compared.

\begin{theorem}[Characterization of Interventional Alignment]
\label{thm:necessity}
In the setup of Definition~\ref{def:alignment}:

\begin{enumerate}[leftmargin=*,itemsep=4pt]
\item[(i)] \emph{Distributional characterization.}
The system satisfies distributional alignment if and only if
there is a map $\Psi$ assigning an output distribution to every
evaluated pair $(x,c)$ such that
\[
(\pi_Y)_*P_{\Phi}(\cdot\mid I_x^R(z))
=
\Psi(x,c)
\]
for every $R\in\mathcal{R}_x$ and $z\in D_x^R$ with
$\pi_C(z)=c$.

\item[(ii)] \emph{Structural characterization.}
The system satisfies structural alignment if and only if
there is a map $\psi$ such that
\[
\pi_Y\bigl(\Phi(I_x^R(z),u_{\mathrm{sys}})\bigr)
=
\psi(x,c,u_{\mathrm{sys}})
\]
for every evaluated $(x,c)$, every $R\in\mathcal{R}_x$,
every $z\in D_x^R$ with $\pi_C(z)=c$, and every
$u_{\mathrm{sys}}\in \mathcal{U}_{\mathrm{sys}}$.

\item[(iii)] Structural alignment implies distributional alignment,
but the converse fails in general.
\end{enumerate}
\end{theorem}

The characterization says that the output depends only on the intervention target and relevant context, with system randomness held fixed in the structural case. Differences between admissible realizations do not affect the resulting mapping. This factorization describes a dependency of the output; it need not correspond to separate stages of computation.

\emph{Proof sketch.}
For (i), distributional alignment makes the output law common to
all admissible starting-state and realizer pairs for $(x,c)$.
Define $\Psi(x,c)$ to be that common law; an admissible pair exists
by the setup. Conversely, a common map $\Psi$ gives equality across
all such pairs. Part (ii) repeats this argument at each fixed
$u_{\mathrm{sys}}$. For (iii), equal output maps have equal
pushforward distributions, whereas distinct output maps can induce
the same distribution. The full proof appears in Appendix~\ref{app:alignment-proof}.
\hfill$\square$

}

{

\paragraph{Grouping states by target and context.}
The states representing a fixed target $x$ and context $c$ form a set called a fiber:
\[
F_{x,c}
:= \{z\in Z:\pi_X(z)=x,\ \pi_C(z)=c\}.
\]

Its elements are the constitutive realizers of $(x,c)$. Only some may be attainable through the intervention processes included in the evaluation. These form the subset:
\[
F^{\circ}_{x,c}
:=
\left\{
I_x^R(z):
R\in\mathcal{R}_x,\ z\in D_x^R,\ \pi_C(z)=c
\right\}
\subseteq F_{x,c}.
\]
Thus $O$ describes available starting states, whereas
$F^{\circ}_{x,c}$ describes attainable post-intervention states.
The latter need not belong to the observational support, even
when the starting states do.

For an already-reached state $z\in F^{\circ}_{x,c}$, let $K_x(z)$ describe the output distribution and let $y_x(z,u_{\mathrm{sys}})$ denote the output at fixed exogenous conditions:
\[
K_x(z):=(\pi_Y)_*P_{\Phi}(\cdot\mid z),
\qquad
y_x(z,u_{\mathrm{sys}})
:=\pi_Y\bigl(\Phi(z,u_{\mathrm{sys}})\bigr).
\]
No second intervention is applied when evaluating these outputs.

The distributional and structural alignment conditions of
Definition~\ref{def:alignment} are equivalently
\[
K_x(z_1)=K_x(z_2)
\qquad
(z_1,z_2\in F^{\circ}_{x,c})
\]
and
\[
y_x(z_1,u_{\mathrm{sys}})
=
y_x(z_2,u_{\mathrm{sys}})
\qquad
(z_1,z_2\in F^{\circ}_{x,c},\
 u_{\mathrm{sys}}\in \mathcal{U}_{\mathrm{sys}}),
\]
respectively. These conditions compare states reached from
different starting states and through different admissible
realizers of the same target in the same relevant context.
A dependence on any remaining difference between these states can
therefore prevent the model from assigning a single answer to $(x,c)$.
}

To detect changes in output distributions, consider bounded measurable functions $\chi$ from a class that distinguishes those distributions. For each such function, define
\[
g_{\chi}(z)
:= \int \chi\,\mathrm{d}K_x(z)
= \mathbb{E}\bigl[\chi(y_x(z,U_{\mathrm{sys}}))\bigr]
\]
at an already-intervened state $z\in F^{\circ}_{x,c}$. If this expectation changes between two states, their output distributions differ.

Variation along an attainable fiber preserves the intervention target and relevant context. Write $\pi_{X,C}:=(\pi_X,\pi_C)$. Admissible fiber directions are tangent to curves of attainable post-intervention states and lie in $\ker(d\pi_{X,C})$. Under the smoothness and connectedness conditions of Proposition~\ref{prop:local-witness}, failure of distributional alignment is equivalent to a nonzero Lie derivative of $g_\chi$ along such a direction:
\[
\mathcal{L}_v\, g_{\chi}(z) \neq 0
\qquad \text{for some admissible } v \in \ker(d\pi_{X,C}(z)).
\]
For continuous outputs, the pathwise condition admits the same witness with $y_x(\cdot,u_{\mathrm{sys}})$ in place of $g_{\chi}$, provided this output map is continuously differentiable. For discrete outputs, such as tokens, the distributional derivative witness applies when the expected test-function outputs satisfy the stated differentiability assumptions.

{
\paragraph{From computational violations to alignment failure.}

Spurious dependence breaks alignment when it changes the output between
attainable realizations of the same target and context.

\noindent\textbf{Locality.} Take the parent variables as the target: varying a distinct nonparent
input while holding the mechanism specification and other arguments
fixed compares realizations of that target.

\noindent\textbf{Autonomy.} For the parameter-based comparison, take $A\notin\mathrm{MPA}^*(i)$ and compare
implementations of the same $X_A$ target at the same $Z$-state but
different $\theta_A$. In this parameterization, $\theta_A$ is external
to $Z$, so the comparison uses $Z\times\Theta_A$, with target and
context projections ignoring the parameter coordinate.

If the mechanism specification is instead encoded within $Z$, its
alternative realizations can already be represented there. To witness
alignment failure, they must be attainable realizations of the same
target and context; to witness autonomy failure, their difference must
also implement the specified change to another mechanism while
preserving the legitimate inputs to the mechanism being assessed.

Lower bounds on sensitivity quantify output disagreement in these
input-variable and mechanism-parameter comparisons.
Write $\varepsilon_{\mathrm{loc}}^{A,i}(\hat w)$ and
$\varepsilon_{\mathrm{aut}}^{A,i}(\hat v)$ for positive lower bounds on
$|\langle\partial_{X_A}\mathcal M_i,\hat w\rangle|$ and
$|\langle\partial_{\theta_A}\mathcal M_i,\hat v\rangle|$,
respectively, for unit directions $\hat w,\hat v$.
The bounds hold uniformly over the specified comparison segments
and evaluation domains, with other mechanism arguments fixed.

Suppose the two scalar outputs are evaluations of the same mechanism
at endpoints separated by $s$ times the specified unit direction.
If the mechanism-induced mapping is $C^1$ on the connecting segment in its domain,
the nonvanishing sensitivity cannot change sign along that segment.
Integrating it therefore gives an output difference of at least
$|s|\varepsilon$, where $\varepsilon$ is the corresponding bound,
at each exogenous value where the bound holds.
This establishes failure of structural alignment.
With a common randomness law, integrable outputs, and a common sign
across exogenous values, the same bound holds for the difference
in means, establishing distributional disagreement
(Corollary~\ref{cor:la-alignment}).

\paragraph{Relationship to causal abstraction.}
Causal abstraction requires consistency between corresponding
interventions at different levels of description
\citep{Rubenstein2017,BeckersHalpern2019}. Interchange interventions
in neural networks test such relationships by substituting internal
activations \citep{Geiger2021,Geiger2025}. Here, task projections
specify the target, context, and output without requiring a separate
high-level SCM. Agreement across admissible realizations determines
whether a single task-level interventional mapping exists.

We distinguish agreement in distribution from agreement at fixed
system randomness. The core results of
\citet{BeckersHalpern2019} concern deterministic models, with
probabilistic extensions developed separately by
\citet{BeckersEberhardtHalpern2019}.

}

\subsection{When observational learning increases interventional error}
\label{sec:gradient}
{

Observational loss can decrease while interventional error increases. We begin with a locality example, establish conditions for introducing and retaining nonlocal dependence, and examine damage through shared parameters before giving the general response-error bounds.

For locality, observational learning can reward an input that predicts
residual variation without causing the outcome. Consider a task with
structural equations $Y=\alpha P+U$ and $S=U+N$. Under the observational
regime, let $P,U,N$ be independent, centered variables with unit variance.
The shared exogenous $U$ associates $S$ with $Y$. An intervention replacing
the equation for $S$ leaves the equation for $Y$ unchanged.
Fit $\widehat Y=\alpha P+\beta S$, learning only $\beta$.
Its expected squared-error loss is
\[
L_{\mathrm{obs}}(\beta)=(1-\beta)^2+\beta^2,
\qquad L_{\mathrm{obs}}'(\beta)=4\beta-2.
\]
For $0\leq\beta<1/2$, sufficiently small gradient-descent steps increase
$\beta$ and decrease loss. At fixed $P$, the spurious response magnitude
$\beta|s_1-s_0|$ vanishes at $\beta=0$ but equals
$\tfrac12|s_1-s_0|$ at the observational optimum $\beta^*=1/2$,
for $s_0\ne s_1$.

Thus observational optimization moves the predictor from zero response to interventions on $S$ toward a nonzero response, although those interventions leave $Y$ unchanged.

}

\paragraph{Gradient updates and nonlocal dependence.}
Even the best predictor using only an outcome's direct causes may improve its
observational predictions by using other inputs. The following result shows when
gradient updates introduce this dependence, even though changing those inputs
leaves the worldly outcome unchanged.

Fix distinct $A,i$ with $i\notin\mathrm{Desc}(A)$, and let
$P=X_{\mathrm{PA}(i)}$. The evaluated interventions on $X_A$ leave $X_i$ unchanged
at fixed world exogenous values. Let $\rho$ be the observational law of $(P,X_A)$,
with $\mathbb E_{\mathrm{obs}}[X_i^2]<\infty$. Consider a deterministic predictor
$\mathcal M_i(P,X_A;\theta_i)$ whose baseline satisfies, throughout the evaluated
input domain,
\[
\mathcal M_i(p,a;\theta_i^0)=h(p),
\qquad h(P)=\mathbb E_{\mathrm{obs}}[X_i\mid P].
\]
The conditional mean specifies $h$ up to observational null sets. Its correctness
as an interventional mean additionally requires agreement with the worldly target.

Use squared-error loss and define the residual and its predictable component:
\[
L_i(\theta_i):=\mathbb E_{\mathrm{obs}}
[(X_i-\mathcal M_i(P,X_A;\theta_i))^2],
\]
\[
r_i:=X_i-h(P),\qquad
\gamma(P,X_A):=\mathbb E_{\mathrm{obs}}[X_i\mid P,X_A]-h(P).
\]
Here $\gamma$ is the additional predictive signal available from $X_A$.
To measure dependence on $X_A$, define the conditional-centering operator $C$, which subtracts the average prediction at fixed $P$:
\[
Cf:=f-\mathbb E_{\mathrm{obs}}[f\mid P],\qquad
\mathcal V_{\mathrm{loc}}(\theta_i):=
\|C\mathcal M_i(\cdot;\theta_i)\|_{L^2(\rho)}.
\]
The quantity $\mathcal V_{\mathrm{loc}}$ measures root-mean-square departures from the average prediction at fixed parent inputs, under the observational law $\rho$. It is zero exactly when the predictor equals a function of $P$ alone, $\rho$-almost surely.

Assume $\theta_i$ ranges over an open subset of $\mathbb R^d$ and the map
$\theta_i\mapsto\mathcal M_i(\cdot;\theta_i)$ is twice continuously
Fr\'echet differentiable into $L^2(\rho)$. Updates use the full gradient in
$\theta_i$, holding other parameters and the observational law fixed. At the baseline, $J$ measures how predictions change with the learned parameters. To obtain $J_c$, subtract the conditional average of $J$ at fixed parent inputs:
\[
J:=\nabla_{\theta_i}\mathcal M_i(P,X_A;\theta_i^0),\qquad J_c:=CJ,
\]
The following quantity measures whether the parameter gradient can exploit the additional predictive signal:

\[
a_{\mathrm{obs}}:=\mathbb E_{\mathrm{obs}}[r_iJ]
=\mathbb E_{\mathrm{obs}}[\gamma J_c].
\]
The condition $a_{\mathrm{obs}}\ne0$ means that the parameter gradient can exploit
the additional signal from $X_A$.

\begin{proposition}[Gradient-driven emergence of nonlocal dependence]
\label{prop:locality-emergence}
Under the preceding assumptions, suppose $a_{\mathrm{obs}}\ne0$.
Then $J_c^{\mathsf T}a_{\mathrm{obs}}\ne0$ in $L^2(\rho)$.
For gradient flow $\dot\theta_i=-\nabla L_i(\theta_i)$ starting at
$\theta_i^0$,
\[
\begin{aligned}
\mathcal V_{\mathrm{loc}}(\theta_i(t))
&=2t\|J_c^{\mathsf T}a_{\mathrm{obs}}\|_{L^2(\rho)}+o(t),\\
L_i(\theta_i(t))
&=L_i(\theta_i^0)-4t\|a_{\mathrm{obs}}\|^2+o(t).
\end{aligned}
\]
For all sufficiently small $t>0$, the locality discrepancy strictly increases
from zero and observational loss strictly decreases. The same expansions hold
with $t$ replaced by $\eta>0$ for a full gradient step
$\theta_i^\eta=\theta_i^0-\eta\nabla L_i(\theta_i^0)$.
\end{proposition}

Appendix~\ref{app:locality-emergence} proves the result and specifies the comparison law under which admissible interventions with unchanged worldly outcomes give root-mean-square response error $\sqrt{2}\,\mathcal V_{\mathrm{loc}}$.

\paragraph{Retained improvement requires nonlocal dependence.}
A predictor cannot retain an observational advantage over the best predictor using
only $P$ while eliminating its dependence on $X_A$. The following bound quantifies
this requirement independently of the learning rule.

Keep the observational law $\rho$ and baseline $h$. For any square-integrable
deterministic predictor $\mathcal M(P,X_A)$, let
\[
G(\mathcal M):=L_i(h)-L_i(\mathcal M),
\qquad g:=\|\gamma\|_{L^2(\rho)},
\]
where $L_i$ is the squared-error risk. Thus $G$ measures the reduction in observational squared-error risk achieved by $\mathcal M(P,X_A)$ relative to the best predictor using only $P$, while $g$ measures the additional predictive signal available from $X_A$.
Use the locality discrepancy
$\mathcal V_{\mathrm{loc}}(\mathcal M):=\|C\mathcal M\|_{L^2(\rho)}$
defined above.

\begin{lemma}[Observational improvement bounds nonlocal dependence]
\label{lem:locality-gain}
For every such predictor,
\[
G(\mathcal M)
\leq 2g\,\mathcal V_{\mathrm{loc}}(\mathcal M)
-\mathcal V_{\mathrm{loc}}(\mathcal M)^2.
\]
If $G(\mathcal M)>0$, then $g>0$ and
\[
\mathcal V_{\mathrm{loc}}(\mathcal M)
\geq \frac{G(\mathcal M)}{2g}>0.
\]
If $g=0$, no predictor using these inputs improves on $h$.
\end{lemma}

For fixed predictive signal $g$, retaining a specified positive observational gain requires a positive amount of dependence on $X_A$. This restriction applies independently of how the predictor was learned. The proof appears in Appendix~\ref{app:locality-emergence}.

\begin{remark}[No-go form and correction]
This lemma has the form of a no-go result: under stated assumptions, specified requirements cannot be jointly satisfied. Under the admissibility and worldly-invariance conditions for the comparison law $\nu$ in Appendix~\ref{app:locality-emergence}, the same predictor cannot achieve observational gain $G>0$ without root-mean-square intervention-response error of at least $G/(\sqrt{2}\,g)$ under $\nu$, which compares values drawn from the observational conditional support. Whereas the Causal Hierarchy Theorem \citep{bareinboim2022pch} concerns observational underdetermination, this bound establishes an incompatibility between observational advantage and correct intervention responses. Correction preserving the gain requires changing the prediction setup, which may involve revising the task representation or distinguishing observational from interventional computations. Neither change by itself establishes interventional adequacy; observationally equivalent causal alternatives may still imply different intervention predictions (Theorem~\ref{thm:passive-bias-app}).
\end{remark}

\paragraph{Learning can damage a correct mechanism.}
An update that improves predictions elsewhere in a system can disturb a mechanism
that already gives correct answers. The following result identifies when parameter
coupling produces this damage.

Fix $A\notin\mathrm{MPA}^*(i)$. Under a fixed joint evaluation law $Q$ for parent
inputs and paired system and world randomness, assume
\[
\mathcal M_i(\cdot;\theta_A^0)=f_i^*
\quad Q\text{-almost surely}.
\]
Hold the worldly target and other parameters fixed. Assume that the mechanism-induced mapping is
$C^1$ into $L^2(Q)$ and that the system observational loss
\[
L_{\mathrm{obs}}^{\mathrm{sys}}=L_i+L_{\mathrm{rest}}
\]
is $C^2$ near $\theta_A^0$. Here $L_i$ evaluates mechanism $i$'s predictions, while $L_{\mathrm{rest}}$ evaluates the remaining predictions. Assume
\[
\nabla_{\theta_A}L_i(\theta_A^0)=0,
\qquad
g_A:=\nabla_{\theta_A}L_{\mathrm{rest}}(\theta_A^0)\ne0.
\]

We additionally assume that the gradient of $L_i$ is zero at the baseline. Correctness under the evaluation law $Q$ alone does not imply this condition, although exact fit to observational squared-error targets does. The initial update is therefore driven entirely by the remaining predictions.

Define
\[
J_A:=\nabla_{\theta_A}\mathcal M_i(\cdot;\theta_A^0),
\qquad G_A:=\mathbb E_Q[J_AJ_A^{\mathsf T}].
\]
The condition $g_A^{\mathsf T}G_Ag_A>0$ means that this update changes the correct
mechanism's outputs.

\begin{proposition}[Damage through an observationally driven parameter update]
\label{prop:autonomy-gradient-damage}
Under the preceding assumptions, suppose $g_A^{\mathsf T}G_Ag_A>0$.
Then gradient flow in the block $\theta_A$ satisfies
\[
\begin{aligned}
\|\mathcal M_i(\cdot;\theta_A(t))-f_i^*\|_{L^2(Q)}
&=t\sqrt{g_A^{\mathsf T}G_Ag_A}+o(t),\\
L_{\mathrm{obs}}^{\mathrm{sys}}(\theta_A(t))
&=L_{\mathrm{obs}}^{\mathrm{sys}}(\theta_A^0)-t\|g_A\|^2+o(t).
\end{aligned}
\]
For sufficiently small positive times, mechanism error strictly increases and
system loss strictly decreases. The same expansions hold for a gradient step
$\theta_A^\eta=\theta_A^0-\eta g_A$ with $t$ replaced by $\eta$.
\end{proposition}

The proof appears in Appendix~\ref{app:autonomy-learning-conditions}.

\paragraph{Conditions for increasing interventional error.}
Theorem~\ref{thm:obstruction} below collects three directional results in which observational loss decreases while interventional error increases. Its two autonomy cases and one locality case require the sensitivity conditions stated there.

In the first autonomy case, learning changes a parameter block $\theta_A$ associated with mechanism $A$. Mechanism $i$ also depends on this block, although the represented intervention should leave its worldly counterpart unchanged. If mechanism $i$ initially gives the correct output and remains spuriously sensitive to changes in $\theta_A$, a learning direction that improves system observational loss can move it away from that output. Its sensitivity to $\theta_A$ need not increase.

In the second case, two implementations of the same intervention remain fixed while learning changes how strongly the outcome mechanism responds to their differences. The implementations preserve the world's outcome. Training evaluates one fixed implementation, where predictions can improve even as the model's response to implementation differences grows. This case requires an additional sensitivity-growth assumption and does not require a correct baseline output.

To verify the sensitivity conditions in case (II) for the earlier predictor $\widehat Y=\alpha P+\beta S$, take $\beta=\beta_0+\phi$ near any $\beta_0\in(0,1/2)$. Sensitivity to $S$ is $\beta$ and increases at rate one. Thus case (II) below has observational pressure and sensitivity growth with $\varepsilon_{\mathrm{loc}}=\beta_0/2$ and $\kappa=1$ on a sufficiently small neighborhood.

{
\begin{theorem}[Gradient Misalignment Under Computational L\&A Violations]
\label{thm:obstruction}

Consider a world with causal graph $G=(V,E)$ and parametric model mechanisms with induced mappings $\{\mathcal{M}_i:i\in V\}$. Fix $A,i\in V$ and a real-valued output $X_i$. Write $p$ for the parent inputs. The following three cases summarize the corresponding appendix theorems and assume their full hypotheses, including regularity and integrability conditions. Throughout, parent inputs and evaluation laws remain fixed along the specified learning direction. Autonomy has two variants, (I-a) and (I-b). Locality is case (II).
\begin{enumerate}[leftmargin=*,itemsep=6pt]
\item[(I-a)] \emph{Damage to a correct mechanism}
(Theorem~\ref{thm:autonomy-app}). Suppose $A\notin\mathrm{MPA}^*(i)$,
mechanism $i$ is correct at $\theta_A=0$, and it has nonvanishing
spurious sensitivity along $\theta_A=\phi\hat v$.

Define the expected mechanism output error by
\[
\varepsilon_{\mathrm{int}}^{A,i}(\theta;a)
:=\mathbb E_a|\mathcal M_i(p,U_{\mathrm{sys}};\theta)-f_i^*(p,U^*)|,
\]
where $f_i^*$ is the true mechanism mapping and $\mathbb E_a$ uses a fixed joint law for $(p,U_{\mathrm{sys}},U^*)$ at intervention value $a$.
If system observational loss has a negative derivative at $\phi=0$, then
\[
\varepsilon_{\mathrm{int}}^{A,i}(\phi\hat v;a)
\geq|\phi|\,\varepsilon_{\mathrm{aut}}^{A,i}(\hat v)
\]
for sufficiently small $|\phi|$ and every evaluated $a$.

\item[(I-b)] \emph{Growth of implementation-response error}
(Theorem~\ref{thm:autonomy-response-app}). Suppose
$A\notin\mathrm{MPA}^*(i)$. Fix two admissible realizers of the same
target and context, with parameters $\theta^1=\theta^0+s\hat{v}$,
$s\neq0$, and a disjoint learning path $\eta_\phi=\eta^0+\phi w$.
The endpoints stay fixed and the realizers remain admissible for that
target throughout learning. Assume nonvanishing spurious sensitivity
along the connecting segment, with its magnitude increasing at rate
at least $\kappa_{\mathrm{aut}}>0$, and a negative initial derivative
of system observational loss. The world's mechanism response is zero
at fixed parents because $A\notin\mathrm{MPA}^*(i)$. Write $m_\phi(\theta,u):=\mathcal M_i(p,u;\theta,\eta_\phi)$. Define the expected magnitude of the implementation-response error by
\[
V_{\mathrm{aut}}(\phi):=\mathbb E|
m_\phi(\theta^1,U_{\mathrm{sys}})
-m_\phi(\theta^0,U_{\mathrm{sys}})|
\]
under the fixed system randomness law. Then
\[
V_{\mathrm{aut}}(\phi)\geq
|s|\,\varepsilon_{\mathrm{aut}}^{A,i}(\hat{v}),
\qquad
V_{\mathrm{aut}}'(\phi)\geq|s|\kappa_{\mathrm{aut}}>0.
\]

\item[(II)] \emph{Growth of locality-response error}
(Theorem~\ref{thm:locality-app}). Suppose $i\notin\mathrm{Desc}(A)$.
Fix intervention values $a_1=a_0+\ell\hat{w}$, $\ell>0$, and a
learning path $\theta_i=\theta_i^0+\phi v$. Assume nonvanishing
spurious sensitivity along this state segment, with its magnitude
increasing at rate at least $\kappa>0$, and a negative initial
derivative of mechanism $i$'s observational prediction loss. The world's
response is zero because $i$ is a non-descendant. Write $m_\phi(a,u):=\mathcal M_i(p,a,u;\theta_i^0+\phi v)$. Define the expected magnitude of the locality-response error by
\[
V(\phi):=\mathbb E|
m_\phi(a_1,U_{\mathrm{sys}})-m_\phi(a_0,U_{\mathrm{sys}})|
\]
under the fixed system randomness law. Then
\[
V(\phi)\geq\ell\,\varepsilon_{\mathrm{loc}}^{A,i}(\hat{w}),
\qquad V'(\phi)\geq\ell\kappa>0.
\]
\end{enumerate}
In each case, for sufficiently small $\phi>0$, the corresponding
observational loss decreases while the specified error increases.
Cases (I-b) and (II) require no correct baseline output.
\end{theorem}
}

Case (I-a) measures departure from a correct baseline as learning
changes $\theta_A$. This error can occur wherever the evaluated inputs
occur. Case (I-b) measures growing response differences between two fixed
implementation endpoints as learning changes $\eta$. Its additional
sensitivity-growth assumption bounds response error, not the absolute
error of either endpoint prediction. Applying (I-b) to complete policy
runs requires both runs to supply the specified parent inputs throughout
learning, with world realizer invariance holding for the compared
implementations.

{
\emph{Proof sketch.} For (I-a), fix the parent inputs and exogenous value and let $e(\phi)$ be the model output minus the true output. Baseline correctness gives $e(0)=0$. Its continuous, nonzero derivative has constant sign, so integration gives $|e(\phi)|\geq|\phi|\varepsilon_{\mathrm{aut}}^{A,i}(\hat{v})$ and, for $\phi>0$, $\partial_\phi|e|=|\partial_\phi e|\geq\varepsilon_{\mathrm{aut}}^{A,i}(\hat{v})$. The integrable bound in Theorem~\ref{thm:autonomy-app} permits differentiation under the expectation. Expected error therefore increases, while the system observational-pressure assumption makes system loss decrease locally.

For (I-b) and (II), the world's response is zero at fixed parent
inputs. At each exogenous value, continuous nonvanishing sensitivity
has constant sign along the comparison segment. The absolute response
error therefore equals the integral of the sensitivity's magnitude,
with the factor $|s|$ in the parameterization of (I-b).
The regularity and domination assumptions permit differentiation
under the integrals. Sensitivity growth then gives
\[
V_{\mathrm{aut}}'(\phi)\geq |s|\kappa_{\mathrm{aut}},
\qquad
V'(\phi)\geq \ell\kappa,
\]
while the respective observational-pressure assumptions give local
loss decrease. \hfill$\square$

The full proofs appear in Appendix~\ref{app:proofs}. The error bounds concern pathwise discrepancies and their expected absolute magnitudes; they do not by themselves establish differences between output distributions.
}

\subsection{Endogenous Interventions and Policy-Dependent Model Outputs}
\label{sec:endogenous}
{

When a system uses a policy to bring about an intervention target, the policy specifies a process for realizing that target. Dependence on how the intervention is realized can therefore affect the consequences the system predicts for its own actions.

For autonomy, compare two policies replacing the $A$-generating
mechanism and producing the same target value. Assume the world's
outcome is invariant across these realizers. The policy parameters
$\theta_A$ describe the compared implementations and need not be learned
network weights. If the model's outcome mechanism depends spuriously on
these parameters, it can assign different outcomes to the two ways of
achieving the target. Although its formula
$\mathcal M_i(p,u;\theta_A)$ is unchanged, different policy settings can
induce different rules as functions of $(p,u)$. The autonomy test concerns
whether those rules agree over the comparison domain.

For locality, compare policies that set different values of $X_A$ while
leaving a non-descendant $X_i$ unchanged in the world. A sufficient
condition for this worldly invariance is that neither physical policy
modifies $X_i$ or any of its ancestors in the world's mechanism graph.
Admissibility under the task representation alone does not ensure this.
Spurious model dependence on $X_A$ can then predict an outcome change
where the world has none.

In either case, changing a policy may also change the inputs supplied
to the outcome mechanism, potentially offsetting its direct response.
For the bound on actual policy outputs, assume that the two model runs
supply the same parent inputs. Under this additional condition,
the mechanism's sensitivity produces a difference between the runs.

\begin{theorem}[Endogenous Interventions and Policy-Dependent Model Outputs]
\label{thm:endogenous}
Consider a model with parametric mechanisms with induced mappings $\{\mathcal{M}_i : i \in V\}$ and admissible policies $\Pi_A$ acting on $A$. Write $M^{\pi_A}$ for the model obtained by substituting policy $\pi_A$ for the $A$-generating mechanism. The formulas for the other induced mappings remain fixed, including any dependence they have on the policy parameters $\theta_A$; parameters outside that block are held fixed.

Fix the relevant context and a real-valued output $X_i$ with $i\neq A$. Compare two policies $\pi_A,\pi_A'$ at each common realization $u$ of the system's randomness, and assume their parent inputs to $i$ agree:
\[
X_{\mathrm{PA}(i)}^{M^{\pi_A}}(u)
= X_{\mathrm{PA}(i)}^{M^{\pi_A'}}(u) =: p(u).
\]
In each case below, the connecting segment at fixed $(p(u),u)$ must lie in the mapping's domain, where the mapping is continuously differentiable and its directional sensitivity has magnitude uniformly bounded below by the indicated $\varepsilon>0$.

\smallskip
\noindent\textbf{(a) Autonomy.} Suppose $A\notin\mathrm{MPA}^*(i)$, the policies both realize $X_A=a$ at every $u$, and
$\theta_A(\pi_A)-\theta_A(\pi_A')=s\hat{v}$ for a unit vector $\hat{v}$ and a fixed $s\neq0$. If $\mathcal{M}_i(p,u;\theta_A)$ has directional sensitivity at least $\varepsilon_{\mathrm{aut}}^{A,i}(\hat{v})$ in magnitude along this parameter segment, then
\[
\bigl|X_i^{M^{\pi_A}}(u)-X_i^{M^{\pi_A'}}(u)\bigr|
\ge |s|\,\varepsilon_{\mathrm{aut}}^{A,i}(\hat{v}).
\]

\smallskip
\noindent\textbf{(b) Locality.} Suppose $i\notin\mathrm{Desc}(A)$ and
$X_A^{\pi_A}(u)-X_A^{\pi_A'}(u)=r\hat{w}$ for a unit vector $\hat{w}$ and a fixed $r\neq0$ across realizations. In this comparison the formula and parameters of $\mathcal{M}_i(p,x_A,u)$ are fixed; the policies affect its inputs. If its directional sensitivity has magnitude at least $\varepsilon_{\mathrm{loc}}^{A,i}(\hat{w})$ along the segment between the two values of $X_A$, then
\[
\bigl|X_i^{M^{\pi_A}}(u)-X_i^{M^{\pi_A'}}(u)\bigr|
\ge |r|\,\varepsilon_{\mathrm{loc}}^{A,i}(\hat{w}).
\]

Both bounds hold pathwise and for the expected absolute discrepancy under a common law for $U_{\mathrm{sys}}$. If the outputs have finite expectations and the signed output difference has the same sign across values of $U_{\mathrm{sys}}$, the same bound holds for the absolute difference in means. The output distributions then differ as well.
\end{theorem}

\emph{Proof sketch.} Fix $u$ and the common parent inputs $p(u)$. Along the connecting segment, the continuous, nonzero directional derivative has a constant sign. Integrating its magnitude gives the stated bound on the mechanism's output difference. Equal parent inputs identify the two endpoint evaluations with the actual policy outputs. Taking expectations gives the expected absolute discrepancy; a common sign prevents cancellation and gives the bound on the difference in means. Full proofs appear in Appendix~\ref{app:proofs}. \hfill$\square$

When policies preserve the world's outcome at every fixed world
exogenous value, differing model outputs violate that invariance
pathwise. The comparison applies both to externally selected policies
and to policies through which an acting system generates its own
interventions.
}

{
\subsection{Cumulative impact under sequential composition}

A deviation can affect later computations when they retain some of its effects. We bound cumulative discrepancy by assuming that each stage retains a specified minimum fraction of the preceding discrepancy in expectation. This retention condition is an additional assumption; computational locality or autonomy violations alone do not establish it.

\begin{proposition}[Cumulative Impact of a Persistent Deviation]
\label{prop:cumulative-impact}

Fix a finite horizon $T$ and a time $t\in\{1,\dots,T\}$ of the initial deviation.
Let
$X_s^0$ and $X_s^1$ be reference and perturbed trajectories on a common
probability space, with identical histories before $t$ and a possible
deviation at $t$. For $s>t$, both evolve by the same mechanism-induced mappings
and exogenous values:
\[
X_s^j=\mathcal{M}_s(X_{<s}^j,U_s),\qquad j\in\{0,1\}.
\]
There are no further externally introduced differences between the runs.
Fix measurable task-relevant distances $d_s$ and a joint evaluation law
for the initial conditions, perturbation, and computational randomness.
Define
\[
D_s:=d_s(X_s^1,X_s^0),\qquad
\delta_t:=D_t,\qquad
D_{\mathrm{cum}}^{(t)}:=\sum_{s=t}^{T}D_s,
\]
and assume that each $D_s$ has finite expectation.

Suppose deterministic retention bounds $q_s\in[0,1]$ satisfy
\begin{equation}
\label{eq:expected-retention}
\mathbb{E}[D_{s+1}]\geq q_s\,\mathbb{E}[D_s],
\qquad s=t,\dots,T-1.
\end{equation}
Then
\begin{equation}
\label{eq:cumulative-impact}
\mathbb{E}[D_{\mathrm{cum}}^{(t)}]
\geq
\mathbb{E}[\delta_t]
\left(1+\sum_{s=t+1}^{T}\prod_{r=t}^{s-1}q_r\right).
\end{equation}
An empty sum is zero, so the bound includes $t=T$.
\end{proposition}

\begin{proof} Iterating~\eqref{eq:expected-retention} gives
\[
\mathbb{E}[D_s]
\geq \mathbb{E}[\delta_t]\prod_{r=t}^{s-1}q_r
\qquad (s>t).
\]
By linearity of expectation, summing these inequalities and the identity
$\mathbb{E}[D_t]=\mathbb{E}[\delta_t]$ proves the result.
\end{proof}

\paragraph{Example: a scalar chain.}
Let the subsequent mechanisms be
$X_{s+1}^j=a_sX_s^j+U_{s+1}$ with deterministic $|a_s|\leq1$ and shared
noise. For $d_s(x,x')=|x-x'|$, subtraction gives
\[
D_{s+1}=|a_s|D_s.
\]
Thus $q_s=|a_s|$ satisfies~\eqref{eq:expected-retention}, and the bound
in~\eqref{eq:cumulative-impact} is attained. For constant $|a_s|=q$
with $0<q<1$, the cumulative multiplier is
$(1-q^{T-t+1})/(1-q)$. With $a_s=1/2$, it approaches $2$ although
each discrepancy shrinks. A reset with $a_s=0$ removes the discrepancy
and all its later effects in this memoryless example.

The bound concerns the sum of discrepancies across stages; individual
discrepancies need not grow. It measures interventional error when the reference
trajectory supplies the correct task behavior and the distances measure
the corresponding errors.
}

\clearpage
\section{Mechanism Instantiation and Hypothesis Revision}
\label{sec:instantiation-revision}

To specify a causal correction, we must distinguish changes in a system's
computation, changes in its embodied commitments, and changes in the alternatives
admitted for consideration. Mechanism attribution specifies the organization and
commitments being assessed, whereas hypothesis revision concerns changes within the
system's inquiry.

\subsection{Structural dispositions and mechanism instantiation}

Architecture and learned parameters constrain which computational procedures a system can realize as its state and inputs change. We call these capacities \emph{structural dispositions} for mechanism instantiation. Such dispositions can be deterministic: under specified operating conditions, a calculator may reliably select and execute the same addition algorithm for the same expression.

During operation, computational state and incoming inputs can select or configure a procedure while architecture and learned parameters remain fixed. The current state $z\in Z$ may retain effects of earlier exogenous inputs and of the system's own computation. It can determine which information a procedure uses and which operations it subsequently performs, so instantiation and execution may be interleaved. Here, inputs are exogenous relative to the system boundary under consideration; they are not thereby identified with system randomness $U_{\mathrm{sys}}$.

A fixed overall computational map, such as $\Phi$, can support the realization of different procedures as state and inputs change. A mechanism attribution specifies which organization is proposed and what would count as its persistence or alteration.

On this account, learning occurs when experience changes the system's dispositions to instantiate mechanisms in context. These dispositions can change through learned parameters or retained computational state. A change in context alone can nevertheless exercise an existing disposition without changing it.

An attribution of a causal mechanism can propose that an instantiated computational procedure embodies commitments about worldly interventional or counterfactual outcomes through its role in prediction, planning, or action.

\subsection{Conjecturing and testing causal mechanism attributions}

A causal mechanism attribution proposes both how a computation is organized and what it represents. It specifies the variables, inputs, parameters, dependencies, their computational realization, and intervention correspondences. It also specifies their interpretation, the relevant context and exogenous comparison conditions, and the domain of states and operations.

Assessment distinguishes three questions:
\begin{enumerate}[leftmargin=*,itemsep=3pt]
\item Whether the system implements the specified operations and dependencies (\emph{mechanism instantiation}).
\item Whether their use in prediction, planning, or action supports the attributed causal commitments (\emph{representational significance}).
\item Whether those commitments are correct about the represented process (\emph{worldly adequacy}).
\end{enumerate}
Evidence about a state's computational role can bear on both instantiation and representational significance. Each comparison holds the attribution fixed. Revising it proposes a new account.
Constitutive realizers specify states representing a task value and context, whereas mechanism instantiation concerns the operations and dependencies through which the system computes with those states.

An output difference can refute a mechanism attribution that requires output invariance under the specified comparison. Agreement supports that functional requirement. Identifying the procedure responsible additionally requires evidence about its operations and dependencies.
When an autonomy test changes computational state, it should support interpreting that change as an intervention on an attributed mechanism. It must also establish that the legitimate inputs and exogenous values of the mechanism being assessed remain fixed, while the causal representation requires that mechanism's rule to remain unchanged.
Changing only the inputs to an unchanged mechanism does not establish
an autonomy violation.
Applying the earlier autonomy results on learning and policy-dependent
outputs to changes in computational state requires identifying which
state quantities serve as mechanism parameters. The comparison must
also satisfy the other assumptions of the particular result.

Relations between represented and expressed commitments are treated in Appendix~\ref{app:semantic}.

If an attributed organization requires output invariance under a specified comparison, an output difference rules that organization out. Other candidates may remain possible under the same variable binding and intervention interpretation. If every candidate in the permitted class requires the violated invariance, the comparison rules out the entire class on the specified domain. Excluding arbitrarily accurate approximation additionally requires a positive discrepancy bound under the chosen comparison criterion.

For example, the alignment characterization
(Theorem~\ref{thm:necessity}) rules out a common task-level mapping when
outputs disagree across attainable realizations of the same target and
context at the assessed level.
Failure at fixed system randomness need not imply failure in
distribution, and output disagreement need not imply disagreement in
expressed content.
Conversely, existence of a common mapping does not identify a complete
causal organization.

Comparisons testing an unchanged mechanism must be distinguished from
transitions intended to revise it. Changed evidence or relevant context
can legitimately change an answer. Correction at fixed target and context
can also pass through stages lacking competence. The evaluation domain
specifies which stages and realizations must satisfy the invariance
requirements.

\subsection{Causal commitments and hypothesis revision}
\label{sec:commitment-revision}

Revising an account of the system and revising the system's causal
commitments are different operations. An investigator can change a
mechanism attribution without changing the computation. By contrast,
revision within the system changes the commitments embodied or the
alternatives admitted through its operation. A system may embody a
causal commitment without being able to formulate it for comparison
and revision. Its outputs may also imply causal structure without
establishing that it maintains a corresponding collection of hypotheses.
Neither content attribution nor mechanism attribution requires the
system to compute an interpretation of its own states.

Function-approximation capacity concerns the
range of mappings a network can realize under a specified architecture
and parameterization.
At inquiry stage $t$, the \emph{current hypothesis class} $\mathcal H_t$
comprises the causal alternatives formulated or admitted within the ongoing
inquiry. Admission makes an alternative available
for consideration without requiring its acceptance as true. The
possibility of instantiating a hypothesis under some combination of
input and internal state does not establish that it has been formulated
or admitted within that inquiry. These alternatives concern the world.
They are distinct from an investigator's candidate accounts of what
the system instantiates, unless the inquiry itself concerns the system.

A system may possess computational resources capable of realizing many
mappings while its inquiry admits only causal alternatives that fail
to reproduce the relevant interventional consequences. Expressive
capacity alone establishes neither the adequacy of those alternatives
nor the system's ability to formulate an adequate replacement.

Revision can occur through changes in computational state with architecture and
learned parameters held fixed. Eliminating a candidate, modifying a
commitment, and admitting a new alternative are distinct operations.
The class can contract as well as expand. Its expansion need not enlarge
the network's function-approximation capacity, and a state change need
not constitute a revision of causal commitments.

Correction may therefore require revising a commitment within a task
representation, revising the representation and its intervention
interpretation, or changing the computation that is supposed to realize
them.

\subsection{Limits of correction within an admitted class}
\label{sec:admitted-class}

At any fixed inquiry stage $t$, write $\mathcal H=\mathcal H_t$
for the nonempty class of admitted causal models to which predictions
are restricted, and let $M^*$ denote the true causal model. The restriction concerns the models actually
used to produce predictions, not merely the alternatives named during
inquiry.
It does not assume that the class exhausts the causal models
the underlying computational system could realize. The bound concerns
these models' adequacy for the world, rather than how well candidate
mechanism attributions describe the system's computation.

Containing $M^*$ is sufficient for adequacy, but not necessary: another
model may reproduce its task-relevant interventional mapping. Even when
no member reproduces that mapping exactly, members may approximate it
arbitrarily closely. A positive lower bound on error requires a positive
distance from the true mapping to the set of available mappings.

\paragraph{Approximation within the admitted class.}
Fix a task with intervention variable $X$, real-valued outcome $Y$,
intervention domain $\mathfrak E_{\mathrm{int}}$, and context space
$\mathcal C$. An intervention
specification $e$ may describe a fixed-target intervention or a policy.
Consider candidate models satisfying the agreement-across-realizations
requirement of Definition~\ref{def:int-comp}(i) on this domain, without
assuming correctness. This condition restricts the candidates covered
by the bound; computational instantiation alone does not guarantee it.
Write $M^e$ for candidate model $M$ under $e$, and define
\[
f_M(e,c):=\mathbb E[Y\mid M^e,c].
\]
For $e=\mathrm{do}(X=x)$, this reduces to
$f_M(x,c)=\mathbb E_M[Y\mid\mathrm{do}(X=x),c]$.
Write $\mathcal F_{\mathcal H}:=\{f_M:M\in\mathcal H\}$ for the
available class of interventional mean functions.

Let $\mu_{\mathrm{eval}}$ be a probability measure on the evaluated
intervention--context pairs, with support covering the evaluation
domain. Assume that the candidate mean functions and the worldly
mean function $f_{M^*}$ belong to $F:=L^2(\mu_{\mathrm{eval}})$, and define
\[
d_{\mathrm{int}}(M,M'):=\lVert f_M-f_{M'}\rVert_{L^2(\mu_{\mathrm{eval}})}.
\]
The measure $\mu_{\mathrm{eval}}$ specifies how intervention--context pairs are weighted
in the evaluation. The same class may approximate the world well on
some pairs and poorly on others, so its adequacy gap depends on this
evaluation regime.

\begin{definition}[Model-Class-Adequacy Gap]
\label{def:realizability-gap}
The \emph{model-class-adequacy gap} of $\mathcal H$ relative to $M^*$
under $\mu_{\mathrm{eval}}$ is
\[
\Delta(M^*,\mathcal H;\mu_{\mathrm{eval}})
:=\inf_{M\in\mathcal H}\lVert f_M-f_{M^*}\rVert_{L^2(\mu_{\mathrm{eval}})}.
\]
Dependence on $\mu_{\mathrm{eval}}$ is suppressed where unambiguous.
\end{definition}

\begin{proposition}[Access-Mode Independence of the Model-Class-Adequacy Gap]
\label{prop:realizability-gap}
Let $\mathcal H$ be a nonempty class of causal models with the
interventional mean functions specified above, and let $M^*$ denote
the true causal model. For any protocol $\Lambda$ that receives
observational data, interventional queries, or any combination thereof,
and outputs a model $M_\Lambda\in\mathcal H$ almost surely,
\[
\mathbb E\bigl[\lVert f_{M_\Lambda}-f_{M^*}\rVert_{L^2(\mu_{\mathrm{eval}})}\bigr]
\geq\Delta(M^*,\mathcal H;\mu_{\mathrm{eval}}).
\]
No protocol restricted to outputs in $\mathcal H$ can achieve expected
interventional error below the model-class-adequacy gap, regardless
of data access mode.
\end{proposition}

\emph{Proof.} Membership of $M_\Lambda$ in $\mathcal H$ gives
\[
\lVert f_{M_\Lambda}-f_{M^*}\rVert_{L^2(\mu_{\mathrm{eval}})}
\geq\Delta(M^*,\mathcal H;\mu_{\mathrm{eval}})
\]
almost surely by the definition of the infimum. Taking expectations
preserves the inequality. \hfill$\square$

\paragraph{Observational ambiguity within the class.}
Even when the admitted class contains the true model, observational
equivalence can prevent identification of its interventional behavior.

\begin{definition}[Observational Equivalence Class]
\label{def:obs-equiv}
For a model class $\mathcal H$, let $X$ here denote the joint collection
of observed variables, rather than only the intervention target.
Given their observational distribution $P_{\mathrm{obs}}$, define
\[
\mathcal E_{\mathcal H}(P_{\mathrm{obs}})
:=\{M\in\mathcal H:P_M(X)=P_{\mathrm{obs}}(X)\}.
\]
\end{definition}

Let $\mathcal B\subseteq\mathcal H$ be the models admitted under a specified
inductive bias, assume $M^*\in\mathcal B$, and define the observationally
consistent candidates by
\[
\mathcal A:=\mathcal B\cap\mathcal E_{\mathcal H}(P_{\mathrm{obs}}).
\]
Let $\mathfrak E$ be the evaluated intervention family, with
$P_M(X\mid e)$ denoting the distribution predicted by $M$ under intervention
$e$. A passive learner receives only observational data.

\begin{theorem}[Passive Bias Insufficiency]
\label{thm:passive-bias-app}
Assume $|\mathcal{A}| > 1$, and that there exist distinct $M', M'' \in \mathcal{A}$ and an intervention $e \in \mathfrak{E}$ such that
\[
P_{M'}(X \mid e) \neq P_{M''}(X \mid e).
\]
Then the interventional behavior of $M^*$ is not identifiable from
$P_{\mathrm{obs}}$ within $\mathcal B$.
\end{theorem}

The proof appears in Appendix~\ref{app:passive-bias-proof}.

For a finite
candidate set containing $M^*$, pairwise interventional separability
and exact observation of interventional distributions permit criticism
to identify the true interventional mapping
(Proposition~\ref{prop:criticism-sufficiency-app}). Here inclusion of
$M^*$ gives $\Delta=0$; criticism resolves the remaining ambiguity.
If every candidate is false, tests may eliminate them all without
finding an adequate replacement or the best approximation.

\paragraph{Consequences for correction.}

A positive $\Delta$ rules out exact mean-level, hence exact structural,
competence within $\mathcal H$. Since
$\mu_{\mathrm{eval}}$ is a probability measure, a uniform mean-prediction error bound
$\delta$ also bounds the $L^2(\mu_{\mathrm{eval}})$ error by $\delta$. Thus
$\Delta>\delta$ rules out both $L^2(\mu_{\mathrm{eval}})$ accuracy at tolerance
$\delta$ and the uniform $(\delta,\tau)$ adequacy criterion,
regardless of the invariance tolerance $\tau$.

For fixed $M^*$, evaluation measure $\mu_{\mathrm{eval}}$, and intervention interpretation, eliminating candidates cannot reduce the class's best achievable approximation error:
\[
\varnothing\ne\mathcal H_{t+1}\subseteq\mathcal H_t
\quad\Longrightarrow\quad
\Delta(M^*,\mathcal H_{t+1};\mu_{\mathrm{eval}})
\geq\Delta(M^*,\mathcal H_t;\mu_{\mathrm{eval}}).
\]
Elimination can improve the selected prediction within a class.
Reducing the class's gap requires admitting alternatives whose
interventional mean functions approximate the target more closely.
The same bound applies to each stage's class. 

\clearpage
\section{Interventional Criticism and the Problem of Correction}
\label{sec:scope}

{

Observational equivalence can leave competing intervention predictions
unresolved. Under the identified
learning conditions, observational improvement can increase causal error. When the available model class has a
positive interventional approximation gap, no data-access protocol
restricted to that class can remove it. Correction therefore requires
distinguishing inadequate commitments and, where the class itself is
inadequate, proposing alternatives beyond it.
Interventional criticism can constrain these revisions.

\subsection{What interventional criticism can eliminate}
\label{sec:active-criticism}

A causal hypothesis commits the learner to consequences of possible
interventions. Two explanations of a fault may fit the same
observations yet predict different consequences of an attempted
repair. Testing that disagreement makes the competing commitments
answerable to evidence that observational fit alone cannot supply.

Consider an elimination sequence beginning with the observationally
consistent candidate class $\mathcal H_0$.
An intervention on which candidates disagree permits criticism:
given its exact outcome distribution, candidates predicting a different
distribution are eliminated. If $\mathcal H_0$ is finite, contains the
true model, and is pairwise interventionally separable, successive
separating interventions leave that model as the sole survivor in at
most $|\mathcal H_0|-1$ steps
(Proposition~\ref{prop:criticism-sufficiency-app}).

Pairwise interventional separability means that every pair of distinct candidates predicts different outcome distributions for at least one evaluated intervention.
Candidates agreeing on every interventional distribution remain
indistinguishable by these tests. 
The guarantee concerns elimination within the assumed class. It does not establish that the class contains the truth.

\subsection{From detected failure to revision}

Eliminating a candidate does not specify how a learner should change \citep{popper1959logic}.
A failed prediction can implicate several commitments without determining which should be revised \citep{duhem1954aim}.
Revision requires attributing failure to a commitment, modifying that
commitment, and checking the consequences of the modification.
Revision can change a causal commitment or the account attributing that
commitment to the system.

For agents, comparing outputs at fixed system randomness across
admissible realizations of the same target and relevant context can
expose a dependence that should not affect the answer. Where the required worldly invariance holds, disagreement shows
that the model has failed to preserve it. The comparison does not
identify which answer is correct or which computational dependency
should change. Correction requires evidence or constraints that guide
the revision, followed by tests of its consequences.
Architectural constraints can prevent spurious dependencies,
interventional evidence can expose errors, and corrective operations can
prevent their propagation. 

Repair can also require new hypotheses. If every candidate is false,
criticism may exhaust the class without constructing a replacement.
Correction beyond the current class's approximation limit requires
formulating new causal hypotheses whose interventional predictions
approximate the target more closely.

\subsection{What successful criticism leaves unresolved}

Even exact agreement with the world's interventional distributions
would not establish agreement across realizations at fixed system
randomness. Outputs $U$ and $-U$, with $U=\pm1$ equally likely, have
identical distributions but disagree at every fixed realization.

On the model side, fixing the system's randomness permits tests across
attainable internal realizers and admissible policies. These tests
can expose structural misalignment over the comparisons tried.

World realizer invariance instead
compares worldly outcomes across realizers at the same world
exogenous value. The same distinction applies: at fixed context,
let an unobserved $U^*$ take values $\pm1$ with equal probability.
For two realizers $R_1$ and $R_2$, one candidate world may assign outcomes $U^*$ and $U^*$, respectively, whereas another assigns $U^*$ and $-U^*$. Both worlds give the same outcome distribution under each realizer, but only the first satisfies realizer invariance. Interventional outcome distributions
alone therefore do not identify this cross-world property.
Surviving distributional tests does not certify structural competence.

Criticism can address the mechanism attribution itself: whether the
system instantiates the proposed organization and whether its operation
supports the attributed causal content.
Evidence about internal dependencies and their use in prediction and
action tests these claims. Testing those attributions is distinct from testing worldly correctness.

\subsection{Acquiring and maintaining competence}

A system may implement an adequate causal mechanism without being able
to recognize or repair its failures. Maintaining competence when existing
commitments cease to be adequate raises a further question: must the
system itself be able to formulate alternatives, compare their
consequences, and propose revisions? The conjecture concerns acquisition
and maintenance under conditions that can invalidate current commitments.
The causal alternatives admitted within inquiry can be revised while
learned parameters remain fixed.

Autonomous participation means that the system can formulate alternative
causal commitments, seek or use evidence that discriminates among them,
and propose revisions in response to detected failures.
This conception of inquiry draws on the critical method of proposing conjectures and subjecting them to criticism \citep{popper1963conjectures,deutsch2011beginning}. This requirement
assigns comparison and the proposal of revisions to the system itself.
People and tools may supply observations and execute interventions.
Participation in a process where others supply all comparisons and
replacement commitments would not satisfy this requirement. Surviving a test need not
change a commitment, and a proposed revision remains subject to further
criticism.

Criticism may concern not only predictions within a task representation but also the variables and admissibility criteria that define its interventions. The learner may revise these commitments in response to predictions and evidence. Such revision changes which processes are compared and which invariance claims must be assessed.

\begin{conjecture}[Acquisition and Maintenance of Exact Interventional Competence]
\label{conj:competence-maintenance}
In a domain admitting LAP and world realizer invariance, reliable
acquisition and maintenance of exact interventional competence across
conditions that can invalidate the system's current causal commitments
requires autonomous participation in conjecture and criticism.
This process must support correction of erroneous intervention
predictions and failures of required invariance, including proposals
beyond the current hypothesis class when its members are inadequate.
\end{conjecture}

The performance requirement is specified independently by
Definition~\ref{def:int-comp}. Autonomous participation is the proposed
means necessary for acquiring and maintaining it under the stated
conditions. No preceding theorem depends on this conjecture.

A separate empirical conjecture concerns acquisition and restoration
of mean-level interventional adequacy at a specified tolerance.
Evaluation must fix the intervention domain, error measure and
weighting, tolerance, regime changes, available resources, and recovery
time independently of the learner's method. A counterexample would
reliably restore that adequacy after failure without autonomously
formulating and comparing causal alternatives or receiving replacement
commitments from elsewhere. Retaining adequacy under changes already
handled by an existing mechanism would not address this claim.

Mean-level recovery leaves structural alignment unresolved. An in-context or adaptive learner would refute the exact conjecture if it reliably acquired and restored exact interventional competence (Definition~\ref{def:int-comp}) after its existing commitments failed. It would need to recover both correct interventional means and the required structural invariances, without autonomously formulating and comparing alternative causal commitments.

}

\clearpage
\section*{Funding}

This research did not receive any specific grant from funding agencies in the public, commercial, or not-for-profit sectors.

\section*{Competing interests}

The author declares no competing interests.

\section*{Data availability}

No datasets were generated or analyzed in this study.

\section*{Declaration of generative AI and AI-assisted technologies in the manuscript preparation process}

During the preparation of this work, the author used AI models from OpenAI (o-series and GPT-series models accessed through ChatGPT and Codex), Anthropic (Claude Sonnet and Opus), and Google (Gemini) in an iterative process of conjecture and criticism that covered mathematical development, narrative exposition, and understanding the relevant literature. The author originated the work's conceptual framework and research directions, directed the development of its theorem statements, proof strategies, and exposition, and takes full responsibility for the manuscript's claims, proofs, and conclusions.

\bibliographystyle{unsrtnat}
\bibliography{references}

\newpage

\appendix

\section{Causal Content, Alignment, and Correctness}
\label{app:semantic}

Consider a system tasked with predicting whether a lamp will be lit
after its switch is closed in a specified context. Different internal
states realizing the same intervention target and context may produce
different verbal predictions. ``Closing the switch lights the lamp''
and ``The lamp turns on when the switch is closed'' may express the
same prediction. ``Closing the switch leaves the lamp off'' expresses
a conflicting prediction. Determining what an output difference
establishes about causal content therefore requires specifying which
outputs express the same commitment.

An output may also agree with an internal model without being generated
through its use. This appendix distinguishes the content of a prediction,
its relationship to represented content, and the computation through
which it is produced. It then relates these assessments to computational
alignment and the prediction's correctness about the world.

\subsection{Content assignments and satisfaction conditions}
\label{app:content-attribution}
\label{app:anchor}

To make claims about the content of outputs and internal states precise,
let $\iota(y)$ denote the content attributed to output $y$ under the
specified interpretation. Similarly, let $\sigma(z)$ denote the content
attributed to internal state $z$. An output may express a single
prediction while an internal state represents a model of several causal
dependencies. We therefore allow their contents to belong to different
spaces:
\[
\iota:\mathcal Y\to\mathcal C_{\mathrm{out}},
\qquad
\sigma:Z\to\mathcal C_{\mathrm{rep}}.
\]
For the lamp task, an internal model might describe dependencies among
the switch, power supply, and bulb, while the output predicts only
whether the lamp will light. We can ask both whether the prediction is
supported by the model and whether the system generated it by using the
model. To formulate the content relationship, introduce a relation
\[
\mathrm{R}_{x,c}\subseteq
\mathcal C_{\mathrm{out}}\times\mathcal C_{\mathrm{rep}}.
\]
Here $\mathrm{R}_{x,c}(a,b)$ means that expressed content $a$ faithfully
reflects a task-relevant claim supported by represented content $b$,
for intervention target $x$ and context $c$. What counts as such support
is specified by the interpretation. The notation makes three comparisons
precise:
\begin{enumerate}[leftmargin=*,itemsep=2pt,topsep=4pt]
\item Two outputs express the same content:
$\iota(y_1)=\iota(y_2)$.
\item Two internal states represent the same content:
$\sigma(z_1)=\sigma(z_2)$.
\item An output faithfully reflects content represented by an internal
state: $\mathrm{R}_{x,c}(\iota(y),\sigma(z))$.
\end{enumerate}
The two kinds of content can concern a common worldly subject matter.
Let $\mathcal W$ denote the possible worldly outcomes or conditions
about which they make claims. A claimed outcome or condition may fail
to obtain in the actual world. Represented commitments may be explicit
or distributed across computational structures.

Evidence from the use of outputs and states in prediction, planning, or
action can support or challenge the content assignments and the relation
$\mathrm{R}_{x,c}$ together with the attributed organization.

What would justify assigning the same content to two different outputs?
One approach compares the circumstances in which the outputs meet a
specified task success criterion. These \emph{satisfaction conditions}
provide a basis for identifying shared content
\citep{millikan1984language,papineau1987reality,dretske1988explaining}.
For a fixed intervention $\mathrm{do}(X=x)$ and context $c$, specify a class of candidate worlds satisfying LAP and world realizer invariance, together with a task success criterion. Associate each output with the set of candidate worlds in which it meets that criterion. This satisfaction set specifies a worldly condition in $\mathcal W$. At this level of interpretation, two outputs express the same content in $\mathcal C_{\mathrm{out}}$ exactly when their satisfaction sets coincide. An output meets the success criterion in the actual world exactly when that world belongs to its satisfaction set.

For the lamp task, a prediction succeeds in a candidate world if it
correctly states whether the lamp is lit after the switch is closed in
the specified context. The two affirmative phrasings are correct in
exactly the same candidate worlds: those in which closing the switch
leaves the lamp lit. Under this construction, they therefore receive
the same content assignment.

For scalar predictions, satisfaction conditions supply a quantitative
way to establish distinct content. Suppose success means lying within
$\tau\geq0$ of the same interventional target. If two outputs are more
than $2\tau$ apart, the triangle inequality prevents both from succeeding
in the same candidate world. If at least one succeeds in some candidate
world, that world belongs to one satisfaction set and not the other,
establishing different contents.

Represented content can be assessed through the success conditions of
a state's uses while preserving the additional structure attributed to
its internal model. In the lamp example, $\mathrm{R}_{x,c}$ could require
that the expressed prediction follow from that model under the specified
intervention and context. The model and prediction can satisfy this
relation even if the prediction is false. The candidate worlds, success
criterion, and relation between the two kinds of content are substantive
parts of the interpretation to be justified.

\subsection{Represented content and output generation}
\label{app:comp-sem-la}
\label{app:propagation}

A system might contain a model of the switch, power supply, and bulb
but generate its answer through a separate learned association. Its
answer could agree with the model without using it. Conversely, the
system could use that model to generate a prediction faithfully, while
the model's mistaken assumptions lead to a false prediction. It could
also use the model but produce an output that misstates its prediction.
These cases distinguish causal use of a representation, faithful
expression of its content, and worldly correctness.

To assess \emph{causal use}, an attribution must identify the internal
representation and the operations through which it contributes to
output generation. Schematically, a candidate computation has the form
\[
y=G(z_{\mathrm{rep}},v,u),
\]
where $z_{\mathrm{rep}}$ is the internal state at the stage where the
representation is alleged to guide the subsequent computation, $v$
collects other relevant inputs, and $u$ denotes system randomness.
The attribution specifies which state variables bear the representation
and how their use affects the output. The content maps $\sigma$ and
$\iota$ describe what that state represents and what the output expresses;
$G$ describes the intervening computation. Establishing the attributed
dependencies requires evidence about this computation.

Interventions can test those dependencies. For example, suppose the
internal model represents the bulb as functioning and predicts that
closing the switch will light it. Changing that represented condition
to a broken bulb should alter the prediction in the way the attributed
computation specifies, with the intervention query and other relevant
conditions held fixed. This intervention changes the internal
representation of the bulb, rather than the bulb itself. The test must
justify that the change targets the attributed representation and
preserves the other conditions of the comparison. The informative
comparison concerns the predicted change in the output's content;
a change in wording or arbitrary disruption of the computation would
not establish the attributed use.

We can also ask whether represented and expressed content remain
invariant across realizations of the same intervention target and
context, and whether they stand in $\mathrm{R}_{x,c}$. The last condition
is \emph{semantic faithfulness}.\footnote{
Causal (or Markov) faithfulness concerns whether every conditional
independence in a distribution is implied by d-separation in its
causal DAG \citep{spirtes2000causation}.}

Holding the content assignments, $\mathrm{R}_{x,c}$, and the stage of
assessment fixed, we compare attainable post-intervention states
$z\in F^\circ_{x,c}$ at a fixed target $x$ and context $c$. Write
$\rho(z,u)\in Z$ for the state reached during execution at the identified
stage before the final output is generated. The assignment
$\sigma(\rho(z,u))$ describes the representation at that stage. The
attribution may identify a distributed representation through several
state variables; the choice of stage specifies which such state is
being assessed. During learning comparisons, the computational maps and
kernels may change while the content assignments, relation, and rule
for selecting that stage remain fixed.

Using the system's computational maps, write
\[
y_x(z,u)=\pi_Y(\Phi(z,u)),
\qquad
K_x(z)=(\pi_Y)_*P_\Phi(\cdot\mid z),
\]
where $u$ denotes actual system randomness $u_{\mathrm{sys}}$, held fixed
throughout each pathwise comparison. In the schematic computation above,
$z_{\mathrm{rep}}=\rho(z,u)$. We compare the expressed content
$\iota(y_x(z,u))$ with the earlier represented content
$\sigma(\rho(z,u))$. For distributional comparisons, assume the maps
are measurable with respect to specified sigma-algebras and define
\[
P_\rho(\cdot\mid z)
:=\bigl(\rho(z,\cdot)\bigr)_*P_{U_{\mathrm{sys}}}.
\]
The laws of expressed and represented content are then $\iota_*K_x(z)$
and $\sigma_*P_\rho(\cdot\mid z)$, respectively.

\begin{definition}[Expressed-content invariance]
\label{def:osl}
The system satisfies expressed-content invariance \emph{pathwise} if,
for every evaluated $(x,c)$, all $z_1,z_2\in F^\circ_{x,c}$, and every $u$,
\[
\iota(y_x(z_1,u))=\iota(y_x(z_2,u)).
\]
It satisfies expressed-content invariance \emph{in distribution} if,
for every evaluated $(x,c)$,
\[
\iota_*K_x(z_1)=\iota_*K_x(z_2)
\qquad (z_1,z_2\in F^\circ_{x,c}).
\]
\end{definition}

\begin{definition}[Represented-content invariance]
\label{def:rsl}
The system satisfies represented-content invariance \emph{pathwise} if,
for every evaluated $(x,c)$, $\sigma(\rho(z,u))$ is constant over
$z\in F^\circ_{x,c}$ at every fixed $u$. It satisfies represented-content
invariance \emph{in distribution} if $\sigma_*P_\rho(\cdot\mid z)$ is
constant over that fiber for every evaluated $(x,c)$.
\end{definition}

\begin{definition}[Semantic faithfulness]
\label{def:faith}
The system is semantically faithful if, for every evaluated $(x,c)$,
every $z\in F^\circ_{x,c}$, and every $u$,
\[
\mathrm{R}_{x,c}\bigl(\iota(y_x(z,u)),\sigma(\rho(z,u))\bigr).
\]
\end{definition}

Semantic faithfulness assesses the relation between the two contents.
Evidence that the output is generated through use of the representation
additionally concerns the computational dependencies described above.

A richer representation may faithfully support several expressed
claims, and different representations may support the same claim.
An invariance implication therefore requires an additional restriction
on the relation. One useful case is a task for which represented content
determines a unique expressed commitment.

\begin{lemma}[Relations among the semantic conditions]
\label{lem:sem-relations}
Semantic faithfulness alone implies neither direction between
represented-content invariance and expressed-content invariance, at
either level.
Suppose additionally that, for each evaluated $(x,c)$, there is a map
$p_{x,c}:\mathcal C_{\mathrm{rep}}\to\mathcal C_{\mathrm{out}}$ such that
\[
\mathrm{R}_{x,c}(a,b)\ \Longrightarrow\ a=p_{x,c}(b).
\]
Under semantic faithfulness, represented-content invariance then implies
expressed-content invariance pathwise, and also in distribution when
$p_{x,c}$ is measurable. The converse need not hold.
\end{lemma}

\begin{proof}
For an unrestricted relation, one represented content can be related
to two distinct expressed contents. Keeping the former constant while
varying the latter gives a faithful system with represented-content
invariance but without expressed-content invariance. Relating two distinct
represented contents to one expressed content gives the reverse example. Both
examples can be deterministic, so they apply at either level.
Under the additional restriction, faithfulness gives
\[
\iota(y_x(z,u))=p_{x,c}\bigl(\sigma(\rho(z,u))\bigr).
\]
Equal represented contents therefore yield equal expressed contents;
equal laws of represented content yield equal laws of expressed content
under a measurable $p_{x,c}$. Distinct represented contents can have the same
image under $p_{x,c}$, so the converse fails in general.
\end{proof}

\begin{lemma}[Alignment implies expressed-content invariance]
\label{lem:alignment-osl}
Structural alignment implies expressed-content invariance pathwise for
every content map $\iota$. Distributional alignment implies
expressed-content invariance in distribution for every measurable $\iota$.
\end{lemma}

\begin{proof}
Equal outputs have equal images under $\iota$; equal output laws have equal
pushforwards under $\iota$.
\end{proof}

Corollary~\ref{cor:semantic-instability} connects computational output
discrepancies to failures of expressed-content invariance. Under its
computational hypotheses, injectivity of $\iota$ or the satisfaction-set
separation criterion in Section~\ref{app:anchor} establishes a content difference and hence
failure of pathwise invariance. Distributional failure requires different
laws $\iota_*K_x$. The corollary also covers the corresponding
parameterized autonomy comparisons.

Under semantic faithfulness and the additional assumption that
$\mathrm{R}_{x,c}$ determines expressed content through $p_{x,c}$,
different expressed contents imply different represented contents.
Different expressed-content laws likewise imply different
represented-content laws when $p_{x,c}$ is measurable.

\newpage
\section{Complete Proofs}
\label{app:proofs}

\subsection{Passive Bias Insufficiency for Causal Identification}
\label{app:passive-bias-proof}

Use the candidate set $\mathcal A$, bias-restricted class $\mathcal B$,
and intervention family $\mathfrak E$ introduced before
Theorem~\ref{thm:passive-bias-app}. This notation also applies to the
identification and candidate-elimination results below.

\begin{proof}[Proof of Theorem~\ref{thm:passive-bias-app}]
Both $M'$ and $M''$ lie in $\mathcal A$, so
$P_{M'}(X)=P_{M''}(X)=P_{\mathrm{obs}}$.
The observational data have the same distribution under both models.
Because the models disagree under intervention $e$, those data do not
uniquely determine the interventional prediction within $\mathcal B$.
\end{proof}

\begin{theorem}[Existence of an Identifying Intervention Sequence]
\label{thm:identifying-sequence-app}
Assume $|\mathcal{A}| < \infty$, and that for every distinct pair $M_1, M_2 \in \mathcal{A}$ there exists $e \in \mathfrak{E}$ with $P_{M_1}(X \mid e) \neq P_{M_2}(X \mid e)$. Assume further that interventional distributions are observed without noise. Then there exists a finite sequence of interventions, of length at most $|\mathcal{A}| - 1$, whose induced distributions identify $M^*$ within $\mathcal{B}$.
\end{theorem}

\begin{proof}
Proceed by induction on $|\mathcal{A}|$. If $|\mathcal{A}| = 1$, the claim holds with the empty sequence. Suppose $|\mathcal{A}| = k > 1$. Pick any $M \in \mathcal{A}$ with $M \neq M^*$. By pairwise separability, some $e \in \mathfrak{E}$ satisfies $P_{M}(X \mid e) \neq P_{M^*}(X \mid e)$. Observing the true interventional distribution under $e$ eliminates $M$, yielding an updated candidate set of size at most $k - 1$. Apply the inductive hypothesis. The total length is at most $|\mathcal{A}| - 1$.
\end{proof}

\begin{proof}[Alternative proof using candidate elimination]
Begin with the candidate set $\mathcal A$.
While more than one candidate remains, choose any two distinct survivors
$M_1,M_2$. Pairwise separability supplies an intervention $e$ for which
\[
P_{M_1}(X\mid e)\ne P_{M_2}(X\mid e).
\]
Observe the true interventional distribution and eliminate all candidates
that disagree with it. The two selected predictions differ, so at least
one of those candidates is eliminated.

The true model belongs to the initial set and always agrees with the
observed distribution, so it is never eliminated. Each intervention thus
removes at least one candidate while retaining $M^*$.
After at most $|\mathcal A|-1$ interventions, the sole survivor is $M^*$.
Choosing each separating intervention requires only the candidates'
predictions, without knowing which candidate is true.
\end{proof}

\begin{corollary}[Extension to Structural Causal Models]
Let $\mathcal{H}$ be the class of SCMs over endogenous variables $(X_1,\dots,X_n)$, with interventions defined by $\mathrm{do}(X_S = x_S)$ for arbitrary subsets $S$. If there exist multiple observationally equivalent SCMs differing under at least one intervention, then the conclusions of Theorem~\ref{thm:passive-bias-app} apply unchanged. If in addition the pairwise separability and finiteness conditions of Theorem~\ref{thm:identifying-sequence-app} hold, so does its conclusion.
\end{corollary}

The impossibility claim in Theorem~\ref{thm:passive-bias-app} relates to classical results on Markov equivalence~\citep{verma1990equivalence, andersson1997characterization} and to the underdetermination arguments in~\citep{spirtes2000causation, peters2017elements}. Those results establish that observational data identifies causal structure only up to a Markov equivalence class, under causal Markov and faithfulness assumptions over DAGs. Theorem~\ref{thm:passive-bias-app} differs in three respects: it assumes no graph structure, imposes neither faithfulness nor causal Markov conditions, and states the result for an abstract model class $\mathcal{H}$ with an explicit inductive bias $\mathcal{B}$. The existence claim in Theorem~\ref{thm:identifying-sequence-app} relates to~\citep{hauser2012characterization, eberhardt2005number}, which characterize the number of interventions needed to orient all edges in a DAG. The present bound of $|\mathcal{A}| - 1$ is loose by comparison, reflecting the absence of graph structure: without a combinatorial object to exploit, each intervention is credited with eliminating only a single hypothesis.

\subsection{Candidate Elimination by Disagreement-Maximizing Criticism}
\label{app:candidate-elimination}

For an inquiry proceeding by candidate elimination, let
$\mathcal H_t\subseteq\mathcal A$ denote the candidates retained at stage
$t$, with $\mathcal H_0=\mathcal A$. For an intervention $e \in \mathfrak{E}$, define the disagreement measure
\[
D(e; \mathcal{H}_t) := \bigl|\{\, P_{M}(X \mid e) : M \in \mathcal{H}_t \,\}\bigr| - 1,
\]
the number of distinct interventional distributions induced by $\mathcal{H}_t$ under $e$, minus one. A \emph{disagreement-maximizing policy} selects, at each step, an intervention $e_t \in \arg\max_{e \in \mathfrak{E}} D(e; \mathcal{H}_t)$.

\begin{proposition}[Candidate Elimination by Disagreement-Maximizing Criticism]
\label{prop:criticism-sufficiency-app}
Assume the hypotheses of Theorem~\ref{thm:identifying-sequence-app}: $|\mathcal{A}| < \infty$, pairwise interventional separability on $\mathcal{A}$, and noise-free observation of interventional distributions. Then any disagreement-maximizing policy identifies $M^*$ within $\mathcal{B}$ in at most $|\mathcal{A}| - 1$ steps.
\end{proposition}

\begin{proof}

If $|\mathcal{H}_t| > 1$, pick any distinct $M_1, M_2 \in \mathcal{H}_t$. By pairwise separability, some $e \in \mathfrak{E}$ satisfies $P_{M_1}(X \mid e) \neq P_{M_2}(X \mid e)$, so $D(e; \mathcal{H}_t) \geq 1$. The policy selects such an intervention. Observing the true interventional distribution under $e_t$ eliminates every $M \in \mathcal{H}_t$ whose induced distribution under $e_t$ disagrees with the observed one. Since $D(e_t; \mathcal{H}_t) \geq 1$, at least one hypothesis is eliminated, giving $|\mathcal{H}_{t+1}| \leq |\mathcal{H}_t| - 1$.

The candidate set strictly shrinks at each step and remains a subset of $\mathcal{A}$. Termination occurs when $|\mathcal{H}_t| = 1$, at which point the sole remaining hypothesis is $M^*$. The bound $|\mathcal{A}| - 1$ follows.
\end{proof}

{
\begin{remark}[Identification up to interventional equivalence]
Pairwise separability individuates hypotheses in $\mathcal{A}$ by their interventional distributions. Structurally distinct hypotheses that induce the same interventional distributions under every $e \in \mathfrak{E}$ violate this assumption. When such hypotheses are present, Proposition~\ref{prop:criticism-sufficiency-app} holds with $\mathcal{A}$ replaced by its quotient under interventional equivalence and identifies $M^*$ only up to that equivalence. Criticism accordingly suffices for distributional identification rather than structural identification.
\end{remark}
}

The measure $D$ counts distinct interventional distributions. Alternative measures, such as worst-case or average pairwise divergence between candidates, yield policies with the same termination guarantee under the stated hypotheses. The guarantee depends only on the policy selecting an intervention that separates at least one pair whenever $|\mathcal{H}_t| > 1$. Finer measures matter when observations are noisy or when the cost of interventions varies, neither of which is modeled here.

Proposition~\ref{prop:criticism-sufficiency-app} restates a standard elimination guarantee. Maintaining a candidate set, selecting queries that induce disagreement, and eliminating inconsistent hypotheses are central to version-space learning and query-by-committee \citep{mitchell1982generalization, seung1992query}. Sequential experiment selection has a longer history \citep{chernoff1959sequential}, with causal applications to Bayesian network discovery \citep{tong2001active, he2008active, hauser2012characterization}. Here, the candidate set is an abstract subset of an inductive bias, and finiteness is assumed explicitly. The bound $|\mathcal{A}|-1$ follows by eliminating at least one candidate per query. Its role is to establish what interventional criticism can identify under the stated assumptions.
{
\subsection{Autonomy Violations and Departure from a Correct Output}

Under autonomy violations, changing the parameters of one mechanism can damage another mechanism that was initially correct. The observational objective can still improve when gains elsewhere in the system outweigh that damage.
This damage does not require the spurious sensitivity to increase.

\begin{theorem}[System Observational Descent Can Damage a Correct Mechanism]
\label{thm:autonomy-app}
Consider a causal system with true graph $G=(V,E)$ and a model with parametric mechanisms with induced mappings $\{\mathcal{M}_i:i\in V\}$. Fix $A\in V$ and $i$ with $A\notin\mathrm{MPA}^*(i)$, and suppose the output $X_i$ is real-valued. Under the true causal semantics, changing the parameters of mechanism $A$ does not alter the mechanism for $i$.

For an intervention $\mathrm{do}(X_A=a)$, compare the model and true
mechanism at the same parent inputs $p=x_{\mathrm{PA}(i)}$. This theorem
uses a fixed common-noise evaluation: write $u=(u_{\mathrm{sys}},u^*)$
for the paired system and world exogenous values under its evaluation
law. In this subsection only, $\mathcal{M}_i(p,u;\theta_A)$ abbreviates
$\mathcal{M}_i(p,u_{\mathrm{sys}};\theta_A)$. The pairing is fixed as
parameters vary; no independence or identification of the pairing from
output distributions is assumed. Then
\[
\begin{aligned}
X_i^{\mathrm{model}}(a;\phi\hat{v})&=\mathcal{M}_i(p,u;\phi\hat{v}),\\
X_i^{\mathrm{true}}(a)&=f_i^*(p,u^*).
\end{aligned}
\]
These are mechanism-level comparisons with the parent inputs held fixed;
they do not include changes to those inputs caused by other model
mechanisms. At fixed $(p,u)$, the output discrepancy has no explicit
dependence on $a$. The joint evaluation law for $(p,u)$ may depend on $a$, but
is held fixed as $\theta_A$ varies, so the expected error can depend on
$a$ through that law. Under this law, define
\[
\varepsilon_{\mathrm{int}}^{A,i}(\theta;a)
:=\mathbb{E}\!\left[|X_i^{\mathrm{model}}(a;\theta)-X_i^{\mathrm{true}}(a)|\right].
\]
The system observational objective is the sum of squared-error prediction losses under the observational distribution,
\[
L_{\mathrm{obs}}^{\mathrm{sys}}(\theta)=L_i(\theta)+L_{\mathrm{rest}}(\theta),
\]
where $L_i$ evaluates mechanism $i$ and $L_{\mathrm{rest}}$ the remaining predictions. The parameter block $\theta_A$ is associated with mechanism $A$; changes to it may affect both terms.

Assume:
\begin{enumerate}
\item[(i)] $\mathcal{M}_i$ is $C^1$ in $\theta_A$ with uniformly continuous derivative. Along $\theta_A=\phi\hat{v}$ near $\phi=0$, the directional derivative is bounded in magnitude by a function $B(p,u)$ integrable under the fixed evaluation law for the parent inputs and exogenous variables at each intervention value $a$.
\item[(ii)] \emph{(Directional violation magnitude)} There exist a unit vector $\hat{v}$ in parameter space and a scalar $\varepsilon_{\mathrm{aut}}^{A,i}(\hat{v})>0$ such that
\[
\inf_{(p,u),\,\theta_A}
\bigl|\langle\partial_{\theta_A}\mathcal{M}_i(p,u;\theta_A),\hat{v}\rangle\bigr|
\geq\varepsilon_{\mathrm{aut}}^{A,i}(\hat{v}).
\]
\item[(iii)] \emph{(Causal baseline correctness)} At $\theta_A=0$,
\[
\mathcal{M}_i(p,u;0)=f_i^*(p,u^*)
\]
for every parent configuration $p$ and exogenous value $u$ in the evaluation domain. Hence $\varepsilon_{\mathrm{int}}^{A,i}(0;a)=0$ for every evaluated $a$. Correctness of mechanism $i$ does not imply optimality of the system objective, because other predictions may remain incorrect.
\item[(iv)] \emph{(Local observational pressure along $\hat{v}$)} The system observational loss is continuously differentiable along $\theta_A=\phi\hat{v}$ near $\phi=0$, and its directional derivative along $\hat{v}$ is strictly negative at the reference parameter:
\[
D_{\hat{v}}L_{\mathrm{obs}}^{\mathrm{sys}}(\theta)\big|_{\theta_A=0}<0.
\]
The system objective locally rewards increasing $\phi$ from the reference parameter.
\end{enumerate}
Then, along $\theta_A=\phi\hat{v}$:
\begin{enumerate}
\item[(a)] For all sufficiently small $|\phi|$, every evaluated intervention value $a$, and every $(p,u)$ in the evaluation domain,
\[
|X_i^{\mathrm{model}}(a;\phi\hat{v})-X_i^{\mathrm{true}}(a)|
\geq |\phi|\,\varepsilon_{\mathrm{aut}}^{A,i}(\hat{v}),
\]
and hence
\[
\varepsilon_{\mathrm{int}}^{A,i}(\phi\hat{v};a)
\geq |\phi|\,\varepsilon_{\mathrm{aut}}^{A,i}(\hat{v}).
\]
\item[(b)] For all sufficiently small $\phi>0$,
\[
\begin{aligned}
\partial_\phi\varepsilon_{\mathrm{int}}^{A,i}(\phi\hat{v};a)
&\geq\varepsilon_{\mathrm{aut}}^{A,i}(\hat{v})>0,\\
\partial_\phi L_{\mathrm{obs}}^{\mathrm{sys}}(\phi\hat{v})&<0.
\end{aligned}
\]
Thus the directional derivatives of system observational loss and mechanism output error have opposite signs.
\end{enumerate}
Along the specified direction, improving system observational loss introduces output error into an initially correct mechanism, including under the evaluated interventions.
\end{theorem}

\begin{proof}
\textbf{Step 1: Departure from the correct output.}
Fix $(p,u)$ and define
\[
e(\phi;p,u):=\mathcal{M}_i(p,u;\phi\hat{v})-\mathcal{M}_i(p,u;0).
\]
By baseline correctness, $e$ is exactly $X_i^{\mathrm{model}}(a;\phi\hat{v})-X_i^{\mathrm{true}}(a)$, and $e(0;p,u)=0$. Conditions (i) and (ii) make $\partial_\phi e$ continuous and nonzero, so its sign is constant on the segment from $0$ to $\phi$. The fundamental theorem of calculus therefore gives, for either sign of sufficiently small $\phi$,
\[
|e(\phi;p,u)|
=\int_{\min(0,\phi)}^{\max(0,\phi)}|\partial_s e(s;p,u)|\,ds
\geq |\phi|\,\varepsilon_{\mathrm{aut}}^{A,i}(\hat{v}).
\]
Taking expectations proves part (a).

\textbf{Step 2: Compare error growth with system-loss improvement.}
For $\phi>0$, $e(\phi;p,u)$ has the same sign as its derivative, so
\[
\partial_\phi|e(\phi;p,u)|
=|\partial_\phi e(\phi;p,u)|
\geq\varepsilon_{\mathrm{aut}}^{A,i}(\hat{v}).
\]
The integrable bound in condition (i) permits differentiation under the expectation, giving
\[
\partial_\phi\varepsilon_{\mathrm{int}}^{A,i}(\phi\hat{v};a)
=\mathbb{E}|\partial_\phi e(\phi;p,u)|
\geq\varepsilon_{\mathrm{aut}}^{A,i}(\hat{v})>0.
\]
By condition (iv), the system-loss derivative is negative at $0$ and continuous nearby. It therefore remains negative for all sufficiently small $\phi>0$, proving part (b).
\end{proof}

The policy-comparison result bounds discrepancies between realizations
under its equal-parent-input, segment, and sensitivity conditions,
without requiring either output to be correct. In Theorem~\ref{thm:autonomy-app},
baseline correctness identifies the reference output with the world's
output. Training then moves the mechanism away from that reference.
This error is not specific to intervention, although the bound holds
for every evaluated intervention value $a$. It does not measure growth
in the mechanism's response to a change of intervention or implementation.
}

{
\subsection{Autonomy Violations and Growth of Implementation-Response Error}
\label{app:autonomy-response}

Different implementations of the same intervention can expose a spurious
dependence on mechanism parameters. Learning may strengthen that
dependence even when the compared implementations remain fixed. The
error is the model's predicted response to changing the implementation,
where the worldly response is zero.

Fix $A\notin\mathrm{MPA}^*(i)$, a real-valued output $X_i$, legitimate
parent inputs $p$, and a relevant context $c$. Here $u$ abbreviates the
system exogenous value $u_{\mathrm{sys}}$; world values are denoted $u^*$. Let $R_0,R_1$ be two
admissible process realizers of the same target $X_A=a$ in this context.
Their post-intervention implementation parameters are
\[
\theta^0,\qquad \theta^1=\theta^0+s\hat{v},
\qquad s\neq0,\quad \|\hat{v}\|=1.
\]
These endpoints are fixed properties of the compared realizers. A
disjoint block of learned parameters $\eta$ controls the outcome
mechanism's dependence on $\theta_A$. Along a fixed learning direction
$w$, write
\[
\eta_\phi=\eta^0+\phi w,\qquad
m_\phi(\theta,u):=\mathcal{M}_i(p,u;\theta,\eta_\phi),
\qquad \phi\in J,
\]
where $J$ is an open interval containing $0$ and the parameter path lies
in its domain. All other mechanism inputs and parameters are held fixed.
Assume that both realizers remain admissible and realize $(a,c)$
throughout $J$.

These are mechanism-level comparisons at the same parent inputs, which
do not vary with $\phi$. The world and the evaluation regime also remain
fixed. Since $A\notin\mathrm{MPA}^*(i)$, changing the parameters associated
with mechanism $A$ leaves the world's mechanism $i$ unchanged. At fixed
parent inputs and world exogenous value $u^*$, its response difference
$\Delta_{\mathrm{world}}(u^*)$ is therefore zero. This is a statement about
the mechanism at fixed inputs; interpreting it as the outcome difference
between complete intervention processes additionally requires world
realizer invariance. Define the model's mechanism response error by
\begin{equation}
\label{eq:autonomy-response-error}
\begin{aligned}
e_\phi^{\mathrm{aut}}(u)
&:=\bigl[m_\phi(\theta^1,u)-m_\phi(\theta^0,u)\bigr]
-\Delta_{\mathrm{world}}(u^*)\\
&=m_\phi(\theta^1,u)-m_\phi(\theta^0,u).
\end{aligned}
\end{equation}
Equip the system exogenous space with a fixed probability law $P_{U_{\mathrm{sys}}}$,
common to both realizers and independent of $\phi$, and set
\begin{equation}
\label{eq:autonomy-response-magnitude}
V_{\mathrm{aut}}(\phi):=\mathbb{E}_{P_{U_{\mathrm{sys}}}}|e_\phi^{\mathrm{aut}}(U_{\mathrm{sys}})|.
\end{equation}
No baseline correctness is assumed. The fixed target $a$ and context $c$
specify which realizers are compared and are suppressed in this notation.

\begin{theorem}[Observational Descent Can Increase Implementation-Response Error]
\label{thm:autonomy-response-app}
In the setting above, define the directional sensitivity along the
implementation-parameter segment by
\[
q_\phi(t,u):=
\bigl\langle\partial_{\theta_A}
\mathcal{M}_i(p,u;\theta^0+ts\hat{v},\eta_\phi),\hat{v}\bigr\rangle,
\qquad 0\leq t\leq1.
\]
Assume the following, with pointwise conditions holding on one common
set of full $P_{U_{\mathrm{sys}}}$-measure for all $\phi\in J$ and $t\in[0,1]$:
\begin{enumerate}[leftmargin=*,itemsep=2pt]
\item[(i)] \emph{Regularity.} The connecting parameter segment lies in the
mapping's domain for every $\phi\in J$. The mapping is measurable in
$u$ and continuously differentiable in $\theta_A$ on a neighborhood of
that segment. The sensitivity $q_\phi(t,u)$ is continuous in $(\phi,t)$
and continuously differentiable in $\phi$. There is a function
$B\in L^1(P_{U_{\mathrm{sys}}})$ such that
\[
|q_\phi(t,u)|+|\partial_\phi q_\phi(t,u)|\leq B(u)
\]
throughout the interval and segment.
\item[(ii)] \emph{Autonomy violation.} For some
$\varepsilon_{\mathrm{aut}}^{A,i}(\hat{v})>0$,
\[
|q_\phi(t,u)|\geq\varepsilon_{\mathrm{aut}}^{A,i}(\hat{v})
\]
throughout the same interval and segment.
\item[(iii)] \emph{Increasing spurious sensitivity.} For some
$\kappa_{\mathrm{aut}}>0$,
\[
\partial_\phi|q_\phi(t,u)|\geq\kappa_{\mathrm{aut}}
\]
throughout the same interval and segment.
\item[(iv)] \emph{Observational pressure.} The system squared-error
observational objective $L_{\mathrm{obs}}^{\mathrm{sys}}(\phi)$ is evaluated
along $\eta_\phi$, with the observational distribution and remaining
parameters fixed. It is continuously differentiable near $0$, with
\[
(L_{\mathrm{obs}}^{\mathrm{sys}})'(0)<0.
\]
\end{enumerate}
Then:
\begin{enumerate}[leftmargin=*,itemsep=2pt]
\item[(a)] For every $\phi\in J$ and $P_{U_{\mathrm{sys}}}$-almost every $u$,
\[
|e_\phi^{\mathrm{aut}}(u)|\geq|s|\,\varepsilon_{\mathrm{aut}}^{A,i}(\hat{v}),
\qquad
V_{\mathrm{aut}}(\phi)\geq|s|\,\varepsilon_{\mathrm{aut}}^{A,i}(\hat{v}).
\]
\item[(b)] For every $\phi\in J$,
\[
V_{\mathrm{aut}}'(\phi)\geq|s|\kappa_{\mathrm{aut}}>0.
\]
For all sufficiently small $\phi\geq0$,
\[
(L_{\mathrm{obs}}^{\mathrm{sys}})'(\phi)
V_{\mathrm{aut}}'(\phi)<0.
\]
Thus movement along $w$ decreases system observational loss while
increasing the expected magnitude of the erroneous response to changing
the intervention implementation.
\end{enumerate}
\end{theorem}

\begin{proof}
\emph{Step 1: Integrate along the implementation-parameter segment.}
Fix $\phi\in J$ and $u$ in the common full-measure set. Since the world's
response is zero, the fundamental theorem of calculus gives
\[
e_\phi^{\mathrm{aut}}(u)=s\int_0^1q_\phi(t,u)\,\mathrm{d}t.
\]
At fixed $(\phi,u)$, continuity and nonvanishing make the sign of
$q_\phi(t,u)$ constant along the segment. Consequently,
\[
|e_\phi^{\mathrm{aut}}(u)|=|s|\int_0^1|q_\phi(t,u)|\,\mathrm{d}t
\geq|s|\,\varepsilon_{\mathrm{aut}}^{A,i}(\hat{v}).
\]
The integrable bound ensures that $V_{\mathrm{aut}}$ is finite. Taking
expectations proves (a).

\emph{Step 2: Differentiate the response magnitude.}
The segment, parent inputs, and law $P_{U_{\mathrm{sys}}}$ are fixed as $\phi$ varies.
Nonvanishing sensitivity makes $|q_\phi|$ differentiable in $\phi$, with
$\bigl|\partial_\phi|q_\phi|\bigr|\leq|\partial_\phi q_\phi|\leq B(u)$.
Differentiation under both integrals is therefore justified, giving
\[
\begin{aligned}
V_{\mathrm{aut}}'(\phi)
&=|s|\int_{\mathcal{U}_{\mathrm{sys}}}\int_0^1
\partial_\phi|q_\phi(t,u)|\,\mathrm{d}t\,\mathrm{d}P_{U_{\mathrm{sys}}}(u)\\
&\geq|s|\kappa_{\mathrm{aut}}.
\end{aligned}
\]

\emph{Step 3: Compare with observational loss.}
Continuity of $(L_{\mathrm{obs}}^{\mathrm{sys}})'$ and its negativity at
$0$ imply negativity for sufficiently small $\phi\geq0$. Combining
this with the positive derivative of $V_{\mathrm{aut}}$ proves (b).
\end{proof}

The bound concerns a response difference, not the absolute error of
either endpoint prediction. It requires no common sign across exogenous
values because the expectation is taken after the absolute value.
Differences at fixed randomness can cancel in distribution, so growth
of $V_{\mathrm{aut}}$ alone does not establish different output laws.
The learning conclusion concerns the specified direction, not the effect
of a full gradient-descent update. Increasing sensitivity is an additional
assumption; neither parameter sharing nor an autonomy violation alone
implies it.

For this mechanism-level comparison to describe complete policy runs,
both runs must supply the specified parent inputs $p$ at every evaluated
$\phi$. World realizer invariance must also hold for the compared
implementations, so their complete worldly outcomes agree at fixed
world exogenous value. Equal parent inputs between the two model runs
at each $\phi$ are not
sufficient if their common value changes with training: that change can
also affect the derivative of the discrepancy. Intermediate points of
the connecting segment need only lie in the differentiable mechanism's
domain; only its endpoints must represent the compared admissible realizers.
}

{
\subsection{Locality Violations and Interventional Response Error}

A model can misrepresent an intervention by predicting a change where the world produces none. To measure this error, compare the model's predicted change between two interventions with the corresponding change in the world.

Fix distinct variables $A$ and $i$, with $i \notin \mathrm{Desc}(A)$, and hold the legitimate parent inputs of mechanism $i$ fixed at $p$. Along a parameter direction $v$, write
\[
m_\phi(a,u) := \mathcal{M}_i(p,a,u;\theta_i^0+\phi v),
\]
where $\phi$ indexes the parameter displacement and $u$ abbreviates
$u_{\mathrm{sys}}$. The mechanism is deterministic at each fixed system
exogenous value. Let $y^*(a;p,u^*)$ denote the corresponding world outcome
under $\mathrm{do}(X_A=a)$ at world exogenous value $u^*$.

For two intervention values $a_0,a_1$, define
\[
\begin{aligned}
\Delta_{\mathrm{model}}(\phi,u)
&:= m_\phi(a_1,u)-m_\phi(a_0,u),\\
\Delta_{\mathrm{world}}(u^*)
&:= y^*(a_1;p,u^*)-y^*(a_0;p,u^*).
\end{aligned}
\]
The \emph{interventional response error} is
\begin{equation}
\label{eq:local-response-error}
e_\phi^{\mathrm{loc}}(u) := \Delta_{\mathrm{model}}(\phi,u)-\Delta_{\mathrm{world}}(u^*).
\end{equation}
Because $i \notin \mathrm{Desc}(A)$, changing $A$ leaves the world's outcome unchanged at the same world exogenous value:
\[
y^*(a_1;p,u^*)=y^*(a_0;p,u^*).
\]
The worldly difference is zero for every $u^*$, so the response error does not require a pairing of world and system randomness. Hence $e_\phi^{\mathrm{loc}}(u)=m_\phi(a_1,u)-m_\phi(a_0,u)$. Neither model output is assumed correct; the error concerns the model's response to changing $A$. Equip the exogenous space with a fixed probability measure $P_{U_{\mathrm{sys}}}$, common to both interventions, and define the expected response-error magnitude
\begin{equation}
\label{eq:local-response-magnitude}
V(\phi):=\mathbb{E}_{P_{U_{\mathrm{sys}}}}|e_\phi^{\mathrm{loc}}(U_{\mathrm{sys}})|.
\end{equation}

Observational association can reward a dependence that is causally inappropriate. Under squared-error loss, when differentiation under the expectation is justified,
\[
D_v L_{\mathrm{obs}}
=-2\,\mathbb{E}_{\mathrm{obs}}\bigl[(X_i-\mathcal{M}_i)D_v\mathcal{M}_i\bigr].
\]
The expectation is taken under the observational pushforward, with the mechanism and its parameter derivative evaluated at the observed inputs. A parameter direction decreases observational loss when its induced prediction change has positive expected product with the observational residual. A noncausal input can therefore receive increased weight when it predicts residual variation in the observations. The theorem separately assumes that the rewarded direction increases sensitivity to that input throughout the intervention segment.

\begin{theorem}[Observational Descent Can Amplify Spurious Interventional Responses]
\label{thm:locality-app}
In the setting above, suppose the output is real-valued. Let
\[
a_1=a_0+\ell\hat{w},\qquad \ell>0,\qquad \|\hat{w}\|=1,
\]
and assume the segment $a_0+t\hat{w}$, $0\leq t\leq\ell$, lies in the mapping's domain. Define its directional sensitivity by
\[
q_\phi(t,u):=\bigl\langle\partial_a m_\phi(a_0+t\hat{w},u),\hat{w}\bigr\rangle.
\]
Assume the following conditions on an open parameter interval $J$ containing $0$, with the pointwise conditions holding on one common set of full $P_{U_{\mathrm{sys}}}$-measure:
\begin{enumerate}[leftmargin=*,itemsep=2pt]
\item[(i)] \emph{Regularity.} The mechanism-induced mapping is continuously differentiable along the intervention segment. The sensitivity $q_\phi(t,u)$ is continuous in $(\phi,t)$ and continuously differentiable in $\phi$. There is a function $B\in L^1(P_{U_{\mathrm{sys}}})$ such that
\[
|q_\phi(t,u)|+|\partial_\phi q_\phi(t,u)|\leq B(u)
\]
for all $\phi\in J$ and $t\in[0,\ell]$.
\item[(ii)] \emph{Locality violation.} For some $\varepsilon_{\mathrm{loc}}^{A,i}(\hat{w})>0$,
\[
|q_\phi(t,u)|\geq\varepsilon_{\mathrm{loc}}^{A,i}(\hat{w})
\]
throughout the parameter interval and intervention segment.
\item[(iii)] \emph{Increasing spurious sensitivity.} For some $\kappa>0$,
\[
\partial_\phi|q_\phi(t,u)|\geq\kappa
\]
throughout the same interval and segment.
\item[(iv)] \emph{Observational pressure.} The squared-error observational prediction loss for mechanism $i$, $L_{\mathrm{obs}}(\phi):=L_{\mathrm{obs}}(\theta_i^0+\phi v)$ is continuously differentiable near $0$, with $L'_{\mathrm{obs}}(0)<0$.
\end{enumerate}
Then:
\begin{enumerate}[leftmargin=*,itemsep=2pt]
\item[(a)] The pathwise response error satisfies
\[
|e_\phi^{\mathrm{loc}}(u)|\geq\ell\,\varepsilon_{\mathrm{loc}}^{A,i}(\hat{w})
\quad\text{for }P_{U_{\mathrm{sys}}}\text{-almost every }u,
\]
and consequently $V(\phi)\geq\ell\,\varepsilon_{\mathrm{loc}}^{A,i}(\hat{w})$ for every $\phi\in J$.
\item[(b)] The expected response-error magnitude satisfies
\[
V'(\phi)\geq\ell\kappa>0\qquad(\phi\in J).
\]
For all sufficiently small $\phi\geq0$,
\[
L'_{\mathrm{obs}}(\phi)V'(\phi)<0.
\]
Thus movement along $v$ decreases observational loss while increasing the expected magnitude of the spurious interventional response.
\end{enumerate}
\end{theorem}

\begin{proof}
\emph{Step 1: Express the response through the mechanism's sensitivity.}
Fix $\phi\in J$ and an exogenous value $u$ satisfying the assumptions. The fundamental theorem of calculus gives
\[
e_\phi^{\mathrm{loc}}(u)
=m_\phi(a_1,u)-m_\phi(a_0,u)
=\int_0^\ell q_\phi(t,u)\,\mathrm{d}t.
\]
The function $t\mapsto q_\phi(t,u)$ is continuous and never zero, so its sign is constant along the segment. Therefore,
\[
|e_\phi^{\mathrm{loc}}(u)|=\int_0^\ell|q_\phi(t,u)|\,\mathrm{d}t
\geq\ell\,\varepsilon_{\mathrm{loc}}^{A,i}(\hat{w}).
\]
Taking expectations proves part (a).

\emph{Step 2: Differentiate the response magnitude.}
The preceding equality expresses the response magnitude exactly as an integral of sensitivity magnitudes. The regularity and integrable bound permit differentiation under both integrals:
\[
\begin{aligned}
V'(\phi)
&=\int_{\mathcal{U}_{\mathrm{sys}}}\int_0^\ell
\partial_\phi|q_\phi(t,u)|\,\mathrm{d}t\,\mathrm{d}P_{U_{\mathrm{sys}}}(u)\\
&\geq\int_{\mathcal{U}_{\mathrm{sys}}}\int_0^\ell\kappa\,\mathrm{d}t\,\mathrm{d}P_{U_{\mathrm{sys}}}(u)
=\ell\kappa.
\end{aligned}
\]
Hence the expected response-error magnitude strictly increases along $v$.

\emph{Step 3: Compare with observational loss.}
Since $L'_{\mathrm{obs}}(0)<0$ and $L'_{\mathrm{obs}}$ is continuous, $L'_{\mathrm{obs}}(\phi)<0$ for sufficiently small $\phi\geq0$. Combining this inequality with $V'(\phi)>0$ proves part (b).
\end{proof}

The conclusion concerns movement along the specified parameter direction; other directions may affect response error differently. Moreover, $V$ averages absolute pathwise response errors. A positive $V$ does not by itself imply that the two output distributions differ.
}

{
\subsection{Cumulative Impact of a Persistent Deviation}
\label{app:cumulative-impact}

Proposition~\ref{prop:cumulative-impact} requires no independence between
discrepancies at different stages.

\paragraph{Occurrence and uniform retention.}
If $r_t:=\Pr(\delta_t>0)>0$ and
$\varepsilon_t:=\mathbb{E}[\delta_t\mid\delta_t>0]$, then
$\mathbb{E}[\delta_t]=r_t\varepsilon_t$; if $r_t=0$, the bound is zero.
When $q_s\geq q$ for a common $0<q<1$, the bound yields
\[
\mathbb{E}[D_{\mathrm{cum}}^{(t)}]
\geq \mathbb{E}[\delta_t]\,
\frac{1-q^{T-t+1}}{1-q}.
\]

\paragraph{Connection to interventional error.}
It measures discrepancy from a reference
computation. It bounds interventional error when that reference supplies
the correct task behavior under the evaluated interventions and the
distances measure the corresponding errors. Applying it to a local
discrepancy identified by the earlier results requires such a trajectory
comparison and the additional retention conditions. Computational L\&A
violations alone do not establish persistence. Without a correct
reference, the result measures computational sensitivity rather than
error relative to the world.

\paragraph{A sufficient conditional retention condition.}

The expected-retention condition assumes average retention under
the evaluation law, rather than retention after every history. A stronger
sufficient condition bounds the conditional probability of retaining a
specified fraction of the current discrepancy.

\begin{proposition}[Conditional Survival and Retention]
\label{prop:conditional-retention}
Let $\mathcal{F}_s$ record the history of the paired computations through
stage $s$, including the injected deviation, and let $D_s$ be
$\mathcal{F}_s$-measurable. Assume the discrepancies are nonnegative and
integrable. Suppose deterministic $p_s,\beta_s'\in[0,1]$ satisfy
\[
\Pr\!\left(D_{s+1}\geq\beta_s'D_s\mid\mathcal{F}_s\right)
\geq p_s
\quad\text{almost surely on }\{D_s>0\}.
\]
Then
\[
\mathbb{E}[D_{s+1}\mid\mathcal{F}_s]
\geq p_s\beta_s'D_s,
\qquad
\mathbb{E}[D_{s+1}]
\geq p_s\beta_s'\mathbb{E}[D_s].
\]
Thus $q_s=p_s\beta_s'$ supplies the expected-retention bound.
\end{proposition}

\begin{proof}
Write $A_s:=\{D_{s+1}\geq\beta_s'D_s\}$.
Nonnegativity gives $D_{s+1}\geq\beta_s'D_s\mathbf{1}_{A_s}$.
On $\{D_s>0\}$, conditional expectation therefore yields
\[
\mathbb{E}[D_{s+1}\mid\mathcal{F}_s]
\geq\beta_s'D_s\Pr(A_s\mid\mathcal{F}_s)
\geq p_s\beta_s'D_s.
\]
On $\{D_s=0\}$, the same inequality follows from nonnegativity.
Taking expectations proves the average-retention claim.
\end{proof}

\begin{remark}[Interpretation of Retention Bounds]
\label{rem:beta-regimes}
The bounds $q_s$ are relative to the specified comparison and evaluation
law. They permit correction on some histories and persistence on others.
They are lower bounds on expected retention: $q_s\leq1$ does not establish
that actual discrepancies contract, and a small lower bound does not show
rapid dissipation. A zero $q_s$ supplies no positive guarantee through
that transition. Ordinary Lipschitz continuity provides upper sensitivity
bounds and does not establish the retention inequalities.

If $q_s\geq q\in(0,1)$, the geometric multiplier in the lower bound
approaches $1/(1-q)$ as the remaining horizon increases; this is not an
upper bound on actual cumulative discrepancy. A large multiplier requires
$q$ close to one and a sufficiently long horizon. Establishing that early
errors dominate other errors additionally requires comparing their
occurrence rates, magnitudes, and downstream effects. For $q_s=1$ at every
stage, the bound is $(T-t+1)\mathbb{E}[\delta_t]$.

In systems with memory, a discrepancy can reappear after the current
outputs agree because earlier histories still differ. The distances
$d_s$ must specify the stage quantities being assessed; repeatedly
measuring an entire growing history can count an earlier discrepancy
again without establishing a new downstream effect.
\end{remark}

\paragraph{Extension to several error origins.}

For a run containing several fresh errors, actual trajectory error
need not equal the sum of their isolated effects. Assume a nonnegative
additive decomposition of the stage errors. The evaluation law concerns
the run with multiple errors and need not coincide with the law of an
isolated-perturbation experiment.

\begin{proposition}[Cumulative Error Under Additive Contributions]
\label{prop:multiple-origin-impact}
Fix a finite horizon $T$ and nonnegative, integrable initial deviation
magnitudes $\delta_t$. Suppose the actual stage errors admit
\[
E_s:=d_s(X_s,X_s^*)
=\sum_{t=1}^{s}C_{t\to s},
\qquad C_{t\to s}:=\alpha_{t\to s}\delta_t,
\]
almost surely, with nonnegative coefficients $\alpha_{t\to s}$,
$\alpha_{t\to t}=1$, and integrable contributions $C_{t\to s}$.
Set $D_{\mathrm{total}}:=\sum_{s=1}^{T}E_s$.

For each origin with $A_t:=\{\delta_t>0\}$ of positive probability,
write $\varepsilon_t:=\mathbb{E}[\delta_t\mid A_t]$ and assume:
\begin{enumerate}[leftmargin=*,itemsep=2pt]
\item[(i)] There are nested survival events
$S_{t\to t}=A_t$ and $S_{t\to s+1}\subseteq S_{t\to s}$.
For a common $p_{\min}\in(0,1]$,
\[
\Pr(S_{t\to s+1}\mid S_{t\to s})\geq p_{\min},
\qquad t\leq s<T.
\]
The conditioning includes occurrence of the initial deviation because
every survival event is a subset of $A_t$.
\item[(ii)] For a common $\beta'\in(0,1)$, on $S_{t\to s+1}$,
\[
\alpha_{t\to s+1}\geq\beta'\alpha_{t\to s}.
\]
\item[(iii)] Survival does not change the conditional mean initial
magnitude:
\[
\mathbb{E}[\delta_t\mid S_{t\to s}]=\varepsilon_t,
\qquad t\leq s\leq T.
\]
\end{enumerate}
Then
\begin{equation}
\label{eq:amp-bound}
\mathbb{E}[D_{\mathrm{total}}]
\geq
\sum_{t=1}^{T}\mathbb{E}[\delta_t]\,
\frac{1-(p_{\min}\beta')^{T-t+1}}{1-p_{\min}\beta'}.
\end{equation}
For origins of positive occurrence probability,
$\mathbb{E}[\delta_t]=\Pr(A_t)\varepsilon_t$; other origins contribute zero.
\end{proposition}

\begin{proof}
Fix an origin with $\Pr(A_t)>0$. Nested survival events and (i) give
\[
\Pr(S_{t\to s}\mid A_t)
=\prod_{r=t}^{s-1}\Pr(S_{t\to r+1}\mid S_{t\to r})
\geq p_{\min}^{s-t}.
\]
By (ii) and $\alpha_{t\to t}=1$, on $S_{t\to s}$ one has
$\alpha_{t\to s}\geq(\beta')^{s-t}$. Consequently, (iii) implies
\[
\begin{aligned}
\mathbb{E}[C_{t\to s}]
&\geq (\beta')^{s-t}
       \mathbb{E}[\delta_t\mathbf{1}_{S_{t\to s}}]\\
&=(\beta')^{s-t}\Pr(A_t)
       \Pr(S_{t\to s}\mid A_t)\varepsilon_t\\
&\geq (p_{\min}\beta')^{s-t}\mathbb{E}[\delta_t].
\end{aligned}
\]
The same lower bound is zero for an origin with $\Pr(A_t)=0$.
Using the additive decomposition and summing the finite contributions,
\[
\begin{aligned}
\mathbb{E}[D_{\mathrm{total}}]
&=\sum_{t=1}^{T}\sum_{s=t}^{T}\mathbb{E}[C_{t\to s}]\\
&\geq\sum_{t=1}^{T}\mathbb{E}[\delta_t]
       \sum_{s=t}^{T}(p_{\min}\beta')^{s-t},
\end{aligned}
\]
which is~\eqref{eq:amp-bound}.
\end{proof}

\begin{remark}[Isolated Perturbations and Interacting Errors]
\label{rem:single-origination}
Proposition~\ref{prop:cumulative-impact} compares two actual trajectories after
one controlled perturbation. It does not identify the isolated effect
of one origin with a lower bound on error in a run containing other
deviations. In such a run, deviations can cancel: an initial displacement
of $+1$ can propagate as $+1/2$ and be offset by a new displacement of
$-1/2$, leaving cumulative actual error $1$ rather than the isolated
value $3/2$. Nonnegative magnitudes do not prevent this cancellation.
Dropping the additive decomposition in
Proposition~\ref{prop:multiple-origin-impact} therefore requires a
replacement condition relating the selected contribution to actual
error, such as $E_s\geq C_{t\to s}$ for a selected origin $t$.
Nonlinear interactions in attention-based systems require this
connection to be justified separately.
\end{remark}
}

{
\subsection{Endogenous Interventions under Autonomy Violations}

The two results below compare actual policy runs. Fix the relevant context, a model with parametric mechanisms with induced mappings $\{\mathcal{M}_i:i\in V\}$, and policies $\pi_A,\pi_A'\in\Pi_A$ admissible relative to the specified task representation. Write $M^{\pi_A}$ for the model obtained by substituting $\pi_A$ for the $A$-generating mechanism. The formulas for the other induced mappings remain fixed, including any dependence on $\theta_A$; parameters outside that block are held fixed. Let $u$ abbreviate $u_{\mathrm{sys}}$, a common realization of the system's randomness, with each mechanism using its relevant components, and suppose $X_i$ is real-valued with $i\neq A$. The bounds concern the model runs; their interpretation as erroneous responses additionally requires the world-side invariances stated after each proof.

\emph{Equal parent inputs.} Throughout these two results, assume
\[
X_{\mathrm{PA}(i)}^{M^{\pi_A}}(u)
= X_{\mathrm{PA}(i)}^{M^{\pi_A'}}(u) =: p(u)
\]
for every $u$. The common value may vary with $u$. This is an assumption about the two policy runs, not an additional intervention clamping the parents. Without it, changes in parent inputs can offset a mechanism's direct response to the policy change.

\begin{theorem}[Policy-Dependent Model Outputs under Autonomy Violations]
\label{thm:endo-autonomy-app}
Under the common setup above, fix $A\notin\mathrm{MPA}^*(i)$ and write its induced mapping as $\mathcal{M}_i(p,u;\theta_A)$. Assume:
\begin{enumerate}[leftmargin=*,itemsep=2pt]
\item[(i)] \emph{Policy realizations.} Both policies realize the same target $X_A=a$ at every $u$, and
\[
\theta_A(\pi_A)-\theta_A(\pi_A')=s\hat{v},
\qquad s\neq0,\quad \|\hat{v}\|=1.
\]
\item[(ii)] \emph{Segment and regularity.} For every $u$, the segment
\[
\theta_A(t):=\theta_A(\pi_A')+ts\hat{v},\qquad 0\le t\le1,
\]
lies in the domain of $\mathcal{M}_i(p(u),u;\cdot)$, which is $C^1$ in $\theta_A$ on a neighborhood of that segment.
\item[(iii)] \emph{Directional violation magnitude.} For some $\varepsilon_{\mathrm{aut}}^{A,i}(\hat{v})>0$ and every $u,t$,
\[
\bigl|\langle\partial_{\theta_A}\mathcal{M}_i(p(u),u;\theta_A(t)),\hat{v}\rangle\bigr|
\ge\varepsilon_{\mathrm{aut}}^{A,i}(\hat{v}).
\]
\end{enumerate}
Then, for every $u$,
\[
\bigl|X_i^{M^{\pi_A}}(u)-X_i^{M^{\pi_A'}}(u)\bigr|
\ge |s|\,\varepsilon_{\mathrm{aut}}^{A,i}(\hat{v}).
\]
The same lower bound holds for the expected absolute discrepancy under a common law $P_{U_{\mathrm{sys}}}$. If both outputs are integrable and their discrepancy has a common sign almost surely, it also holds for the absolute difference in means, and the output distributions differ.
\end{theorem}

\begin{proof}
\emph{Step 1: Discrepancy at fixed parents.} Fix $u$, write $p=p(u)$, and define
\[
g(t):=\mathcal{M}_i(p,u;\theta_A(t)).
\]
The segment and regularity conditions make $g$ continuously differentiable, with
\[
g'(t)=s\,\langle\partial_{\theta_A}\mathcal{M}_i(p,u;\theta_A(t)),\hat{v}\rangle.
\]
Its derivative is continuous and has magnitude at least $|s|\,\varepsilon_{\mathrm{aut}}^{A,i}(\hat{v})>0$, so its sign is constant on $[0,1]$. The fundamental theorem of calculus gives
\[
|g(1)-g(0)|
=\int_0^1|g'(t)|\,dt
\ge |s|\,\varepsilon_{\mathrm{aut}}^{A,i}(\hat{v}).
\]

\emph{Step 2: Identify the policy outputs.} Equal parent inputs give
\[
g(1)=X_i^{M^{\pi_A}}(u),\qquad
g(0)=X_i^{M^{\pi_A'}}(u).
\]
The bound in Step~1 is therefore the claimed pathwise bound for the actual runs.

\emph{Step 3: Expectations and distributions.} Write
\[
\Delta_i(u):=X_i^{M^{\pi_A}}(u)-X_i^{M^{\pi_A'}}(u).
\]
Taking expectations gives
\[
\mathbb{E}|\Delta_i(U_{\mathrm{sys}})|\ge |s|\,\varepsilon_{\mathrm{aut}}^{A,i}(\hat{v}).
\]
If the outputs are integrable and $\Delta_i$ has a common sign almost surely, then
\[
\bigl|\mathbb{E}X_i^{M^{\pi_A}}-\mathbb{E}X_i^{M^{\pi_A'}}\bigr|
=\mathbb{E}|\Delta_i(U_{\mathrm{sys}})|
\ge |s|\,\varepsilon_{\mathrm{aut}}^{A,i}(\hat{v}).
\]
Distinct finite means imply distinct output distributions.
\end{proof}

The policies realize the same target. Under world realizer invariance, the world's outcome is the same under both, whereas the model outputs differ pathwise. This is a failure of process invariance; neither model output is assumed correct.

\subsection{Endogenous Interventions under Locality Violations}

\begin{theorem}[Policy-Dependent Model Outputs under Locality Violations]
\label{thm:endo-locality-app}
Under the common setup above, fix $i\notin\mathrm{Desc}(A)$ and write its induced mapping as $\mathcal{M}_i(p,x_A,u)$. Its formula and parameters are fixed across the two runs; the policies affect its inputs. Assume:
\begin{enumerate}[leftmargin=*,itemsep=2pt]
\item[(i)] \emph{Policy realizations.} For a fixed $r\neq0$ and unit vector $\hat{w}$, the realized values satisfy
\[
X_A^{\pi_A}(u)-X_A^{\pi_A'}(u)=r\hat{w}
\]
for every $u$.
\item[(ii)] \emph{Segment and regularity.} For every $u$, the segment
\[
X_A(t):=X_A^{\pi_A'}(u)+tr\hat{w},\qquad 0\le t\le1,
\]
lies in the domain of $\mathcal{M}_i(p(u),\cdot,u)$, which is $C^1$ in $X_A$ on a neighborhood of that segment.
\item[(iii)] \emph{Directional violation magnitude.} For some $\varepsilon_{\mathrm{loc}}^{A,i}(\hat{w})>0$ and every $u,t$,
\[
\bigl|\langle\partial_{X_A}\mathcal{M}_i(p(u),X_A(t),u),\hat{w}\rangle\bigr|
\ge\varepsilon_{\mathrm{loc}}^{A,i}(\hat{w}).
\]
\end{enumerate}
Then, for every $u$,
\[
\bigl|X_i^{M^{\pi_A}}(u)-X_i^{M^{\pi_A'}}(u)\bigr|
\ge |r|\,\varepsilon_{\mathrm{loc}}^{A,i}(\hat{w}).
\]
The same lower bound holds for the expected absolute discrepancy under a common law $P_{U_{\mathrm{sys}}}$. If both outputs are integrable and their discrepancy has a common sign almost surely, it also holds for the absolute difference in means, and the output distributions differ.
\end{theorem}

\begin{proof}
\emph{Step 1: Discrepancy at fixed parents.} Fix $u$, write $p=p(u)$, and define
\[
h(t):=\mathcal{M}_i(p,X_A(t),u).
\]
The segment and regularity conditions give
\[
h'(t)=r\,\langle\partial_{X_A}\mathcal{M}_i(p,X_A(t),u),\hat{w}\rangle.
\]
This derivative is continuous and has magnitude at least $|r|\,\varepsilon_{\mathrm{loc}}^{A,i}(\hat{w})>0$, so its sign is constant. Hence
\[
|h(1)-h(0)|
=\int_0^1|h'(t)|\,dt
\ge |r|\,\varepsilon_{\mathrm{loc}}^{A,i}(\hat{w}).
\]

\emph{Step 2: Identify the policy outputs.} Equal parent inputs and the fixed mechanism give
\[
h(1)=X_i^{M^{\pi_A}}(u),\qquad
h(0)=X_i^{M^{\pi_A'}}(u).
\]
Thus Step~1 bounds the discrepancy between the actual policy runs.

\emph{Step 3: Expectations and distributions.} With
$\Delta_i(u):=X_i^{M^{\pi_A}}(u)-X_i^{M^{\pi_A'}}(u)$,
taking expectations gives
\[
\mathbb{E}|\Delta_i(U_{\mathrm{sys}})|\ge |r|\,\varepsilon_{\mathrm{loc}}^{A,i}(\hat{w}).
\]
If both outputs are integrable and $\Delta_i$ has a common sign almost surely, then
\[
\bigl|\mathbb{E}X_i^{M^{\pi_A}}-\mathbb{E}X_i^{M^{\pi_A'}}\bigr|
=\mathbb{E}|\Delta_i(U_{\mathrm{sys}})|
\ge |r|\,\varepsilon_{\mathrm{loc}}^{A,i}(\hat{w}).
\]
The means, and therefore the output distributions, differ.
\end{proof}

These policies set different values of $X_A$. To interpret their model discrepancy as a spurious response, assume that their physical implementations leave the world outcome $X_i$ unchanged at each fixed world exogenous value. This holds, for example, if neither policy modifies $X_i$ or any of its ancestors in the world's mechanism graph. Admissibility relative to the task representation alone does not imply this world-side condition, nor does realizer invariance across processes setting the same value of $X_A$ establish invariance across different values. Under the stated world-side condition, the model responds where the world's outcome is unchanged.

In either proof, a sufficient condition for the mean-level conclusion is that the directional sensitivity has a common sign throughout the segment and across realizations of the randomness. Because $s$ or $r$ is fixed, the output discrepancy then has a common sign as well. Without this condition, pathwise discrepancies can cancel in distribution, as the counterexample in the proof of Corollary~\ref{cor:la-alignment} demonstrates.

The segment conditions justify integration of the mechanism at fixed inputs. Intermediate points need not themselves correspond to admissible policies. Equal parent inputs are required only for the two policy runs whose outputs are compared.
}

\subsection{Interventional Alignment and Local Witnesses}
\label{app:alignment-proof}

{
\noindent\textbf{Proof of Theorem~\ref{thm:necessity}.}

Use the computational map, projections, starting-state set
$O$, and admissible intervention maps $I_x^R:D_x^R\to Z$ of
Definition~\ref{def:alignment}. For every evaluated $(x,c)$, define
\[
F_{x,c}
:=\{z\in Z:\pi_X(z)=x,\ \pi_C(z)=c\},
\]
\[
F^{\circ}_{x,c}
:=
\left\{
I_x^R(z):
R\in\mathcal{R}_x,\ z\in D_x^R,\ \pi_C(z)=c
\right\}.
\]
Evaluation is restricted to pairs for which
$F^{\circ}_{x,c}\neq\emptyset$.

For an already-intervened state $z\in F^{\circ}_{x,c}$, write
\[
K_x(z):=(\pi_Y)_*P_{\Phi}(\cdot\mid z),
\qquad
y_x(z,u_{\mathrm{sys}})
:=\pi_Y\bigl(\Phi(z,u_{\mathrm{sys}})\bigr).
\]
In this notation, distributional alignment means that $K_x$ is constant
on each attainable fiber $F^{\circ}_{x,c}$. Structural alignment means
that $y_x(\cdot,u_{\mathrm{sys}})$ is constant on each such fiber for every
fixed $u_{\mathrm{sys}}$. These are the conditions used in parts (i) and
(ii) of the proof below.

\begin{proof}
By construction, every admissible starting-state and realizer
pair produces a state in $F^{\circ}_{x,c}$, and every state in
$F^{\circ}_{x,c}$ arises from such a pair. Consequently,
constancy over this attainable set is equivalent to agreement
across all admissible pairs.

\emph{(i), necessity.} Suppose $\Psi$ exists, and let
$z_1, z_2 \in F^{\circ}_{x,c}$. Then
$K_x(z_1) = \Psi(x,c) = K_x(z_2)$.

\emph{(i), sufficiency.} Suppose the condition holds. For each $(x,c)$ with
$F^{\circ}_{x,c} \neq \emptyset$, choose any $z \in F^{\circ}_{x,c}$ and set
$\Psi(x,c) := K_x(z)$. By the condition, this value does not depend on the
choice of $z$, so that $\Psi$ is a function of $(x,c)$ alone.

\emph{(ii).} The same argument applies at each fixed $u_{\mathrm{sys}}$ with
$y_x(\cdot, u_{\mathrm{sys}})$ in place of $K_x$, setting
$\psi(x, c, u_{\mathrm{sys}}) := y_x(z, u_{\mathrm{sys}})$ for any
$z \in F^{\circ}_{x,c}$.

\emph{(iii).} If the condition in (ii) holds, then for
$z_1, z_2 \in F^{\circ}_{x,c}$ the maps $y_x(z_1, \cdot)$ and
$y_x(z_2, \cdot)$ on $\mathcal{U}_{\mathrm{sys}}$ coincide, so that their pushforwards
under $P_{U_{\mathrm{sys}}}$ coincide as well.
The counterexample in the proof of Corollary~\ref{cor:la-alignment}
shows that the converse fails, even under a locality violation.
\end{proof}

\paragraph{Lie-derivative witness.} To obtain a local witness of
distributional alignment failure, consider expected values of test
functions of the output. The differentiability assumptions below concern
these expectations as functions of the post-intervention state.

Fix $(x,c)$. For a bounded measurable $\chi : \mathcal{Y} \to \mathbb{R}$, define
\[
g_{\chi}(z) := \int \chi \, \mathrm{d}K_x(z)
= \mathbb{E}\bigl[\chi\bigl(y_x(z, U_{\mathrm{sys}})\bigr)\bigr],
\]
the expected value of $\chi$ after computation from the already-intervened state $z$. Let $\mathcal{T}$ be a
class of bounded measurable functions that separates distributions on
$\mathcal{Y}$, in the sense that $\int \chi \, \mathrm{d}P_1 = \int \chi \, \mathrm{d}P_2$
for all $\chi \in \mathcal{T}$ implies $P_1 = P_2$. When $\mathcal{Y}$ is finite,
the indicators of single outputs form such a class, and $g_{\mathbf{1}\{y\}}(z)$
is the probability of output $y$. When $\mathcal{Y}$ is a metric space with its
Borel $\sigma$-algebra, the bounded continuous functions form such a class.
Since $\mathcal{T}$ separates distributions, $K_x(z_1) = K_x(z_2)$ if and only
if $g_{\chi}(z_1) = g_{\chi}(z_2)$ for all $\chi \in \mathcal{T}$. Differentiation is with respect to $z$, so that $\chi$ need not be smooth.

A vector $v \in T_z Z$ at $z \in F^{\circ}_{x,c}$ is an \emph{admissible fiber
direction} if $v = \gamma'(0)$ for some $C^1$ curve
$\gamma : (-\delta, \delta) \to F^{\circ}_{x,c}$ with $\gamma(0) = z$. For $g$ differentiable at $z$,
$\mathcal{L}_v g(z) := \mathrm{d}g(z)[v]$ is the Lie derivative of $g$ along $v$. Because $g$ is a function rather than a
vector field, it reduces to the directional derivative of $g$ along $v$.

\begin{proposition}[Lie-derivative witness of alignment failure]
\label{prop:local-witness}
Fix $(x,c)$ and a separating class $\mathcal{T}$ as above, and assume:
\begin{enumerate}[leftmargin=*,itemsep=2pt]
\item[(H0)] $Z$ is a smooth manifold, for instance an open subset of
$\mathbb{R}^n$, and $\pi_X$, $\pi_C$ are $C^1$;
\item[(H1)] any two points of $F^{\circ}_{x,c}$ can be joined by a piecewise
$C^1$ path in $F^{\circ}_{x,c}$;
\item[(H2)] for every $\chi \in \mathcal{T}$, $g_{\chi}$ is $C^1$ on an open set
containing $F^{\circ}_{x,c}$.
\end{enumerate}
Then:
\begin{enumerate}[leftmargin=*,itemsep=2pt]
\item[(a)] Under (H0) and (H2), if $\mathcal{L}_v g_{\chi}(z) \neq 0$ for some
$\chi \in \mathcal{T}$, some $z \in F^{\circ}_{x,c}$, and some admissible fiber
direction $v$ at $z$, then the alignment condition of
Theorem~\ref{thm:necessity}(i) fails at $(x,c)$.
\item[(b)] Under (H0)--(H2), if the alignment condition of
Theorem~\ref{thm:necessity}(i) fails at $(x,c)$, then there exist
$\chi \in \mathcal{T}$, $z \in F^{\circ}_{x,c}$, and an admissible fiber direction
$v$ at $z$ with $\mathcal{L}_v g_{\chi}(z) \neq 0$.
\end{enumerate}
Consequently, under (H0)--(H2), the alignment condition holds at $(x,c)$ if and
only if $\mathcal{L}_v g_{\chi}(z) = 0$ for every $\chi \in \mathcal{T}$, every
$z \in F^{\circ}_{x,c}$, and every admissible fiber direction $v$ at $z$.
Every admissible fiber direction at $z$ lies in
$\ker(\mathrm{d}\pi_{X,C}(z))$, where $\pi_{X,C} := (\pi_X, \pi_C)$, and every
vector in the kernel is admissible when $\pi_{X,C}$ is a submersion near
$F_{x,c}$ and $F^{\circ}_{x,c}$ is relatively open in $F_{x,c}$.
\end{proposition}

\begin{proof}
\emph{Admissible directions.} A $C^1$ curve in $F^{\circ}_{x,c}$ keeps
$\pi_{X,C}$ constant, so that its velocity lies in the kernel by the chain rule.
The converse under the submersion assumption is the regular value theorem.

\emph{(a).} Let $\gamma$ be a $C^1$ curve in $F^{\circ}_{x,c}$ with
$\gamma(0) = z$ and $\gamma'(0) = v$, and set $h := g_{\chi} \circ \gamma$. By the
chain rule, $h'(0) = \mathrm{d}g_{\chi}(z)[v] = \mathcal{L}_v g_{\chi}(z) \neq 0$, so
that $h$ is not constant on any neighbourhood of $0$. Hence there is $t$ with
$g_{\chi}(\gamma(t)) \neq g_{\chi}(z)$. Since $\gamma(t)$ and $z$ both lie in
$F^{\circ}_{x,c}$ and $K_x(\gamma(t)) \neq K_x(z)$, the alignment condition
fails.

\emph{(b).} If the alignment condition fails at $(x,c)$, there are
$z_1, z_2 \in F^{\circ}_{x,c}$ with $K_x(z_1) \neq K_x(z_2)$, and hence some
$\chi \in \mathcal{T}$ with $g_{\chi}(z_1) \neq g_{\chi}(z_2)$. By (H1), there is a
piecewise $C^1$ path $\gamma : [0,1] \to F^{\circ}_{x,c}$ from $z_1$ to $z_2$,
with breakpoints $0 = t_0 < t_1 < \dots < t_k = 1$. The function
$h := g_{\chi} \circ \gamma$ satisfies $h(0) \neq h(1)$. Since
$h(1) - h(0) = \sum_{j=1}^{k} \bigl(h(t_j) - h(t_{j-1})\bigr)$, some piece has
$h(t_j) \neq h(t_{j-1})$. On $[t_{j-1}, t_j]$, $h$ is $C^1$ by (H2) and the
chain rule, so that the mean value theorem gives
$t^* \in (t_{j-1}, t_j)$ with
\[
h'(t^*) = \frac{h(t_j) - h(t_{j-1})}{t_j - t_{j-1}} \neq 0 .
\]
By the chain rule, $h'(t^*) = \mathcal{L}_v g_{\chi}(z)$ with
$z := \gamma(t^*) \in F^{\circ}_{x,c}$ and $v := \gamma'(t^*)$, which is an
admissible fiber direction because it is the velocity at $z$ of the $C^1$
curve $\gamma|_{(t_{j-1}, t_j)}$ in $F^{\circ}_{x,c}$.
\end{proof}

For continuous outputs $\mathcal{Y} \subseteq \mathbb{R}^m$, the same argument gives a pathwise witness and characterizes failure of structural alignment (Theorem~\ref{thm:necessity}(ii)). Apply the argument coordinatewise to $y_x(\cdot, u_{\mathrm{sys}})$ at each fixed $u_{\mathrm{sys}}$, replacing (H2) with continuous differentiability of $z \mapsto y_x(z, u_{\mathrm{sys}})$. For discrete
outputs, a change in the output at fixed randomness need not admit a
derivative witness. The distributional witness instead applies when
the expected test-function outputs $g_\chi$ satisfy (H2), even if the
individual output maps are discontinuous.
}

{
\begin{corollary}[L\&A Violations Imply Alignment Failure]
\label{cor:la-alignment}
Let the output be the real-valued $X_i$, and write $u$ for
$u_{\mathrm{sys}}$. Let this system exogenous value be independent of the internal state, so that it has the same distribution under
every post-intervention state, and assume that $\mathcal{M}_i$ is integrable
in $u$, so that the means below exist.
\begin{enumerate}[leftmargin=*,itemsep=2pt]
\item[(a)] \emph{Locality.} Fix distinct $A,i$ with
$A\notin\mathrm{PA}(i)$ and assume that $\mathcal M_i$ depends on
$X_A$ and is $C^1$ in $X_A$. Take $X_{\mathrm{PA}(i)}$ as the
intervention target.
Suppose the attainable post-intervention fiber
$F^{\circ}_{x,c}$ contains states $z_1,z_2$ from which the output
is $\mathcal{M}_i(x,x_A(z_j),u)$, with
\[
x_A^{(1)}-x_A^{(2)}=s\hat{w},
\qquad
x_A^{(j)}:=x_A(z_j),
\]
for a unit vector $\hat{w}$ and some $s\neq0$.
Assume that the segment
$S := \{x_A^{(2)} + t s\hat{w} : t \in [0,1]\}$ lies in the domain of
$\mathcal{M}_i$.
\begin{enumerate}[leftmargin=*,itemsep=2pt]
\item[(a1)] \emph{Pathwise.} If the bound in assumption (ii) of
Theorem~\ref{thm:locality-app} holds at the fixed parameter value on $S$, that is,
\[
\bigl|\langle \partial_{X_A}\mathcal{M}_i(x, x_A, u), \hat{w}\rangle\bigr| \;\ge\;
\varepsilon_{\mathrm{loc}}^{A,i}(\hat{w}) > 0
\]
for all $x_A \in S$ and $P_{U_{\mathrm{sys}}}$-almost every $u$, then for $P_{U_{\mathrm{sys}}}$-almost every $u$
\[
\bigl|\mathcal{M}_i(x, x_A^{(1)}, u) - \mathcal{M}_i(x, x_A^{(2)}, u)\bigr|
\;\ge\; |s|\,\varepsilon_{\mathrm{loc}}^{A,i}(\hat{w}),
\]
and the structural alignment condition of
Theorem~\ref{thm:necessity}(ii) fails.
\item[(a2)] \emph{In mean.} If, in addition, the sensitivity has a common
sign, that is, there is $\zeta \in \{-1, +1\}$ such that, for all
$x_A \in S$ and $P_{U_{\mathrm{sys}}}$-almost every $u$,
\[
\text{(iii}^{+}\text{)} \qquad
\zeta\,\langle \partial_{X_A}\mathcal{M}_i(x, x_A, u), \hat{w}\rangle
\;\ge\; \varepsilon_{\mathrm{loc}}^{A,i}(\hat{w}),
\]
then the means of $X_i$ from the post-intervention states $z_1$ and $z_2$ differ by at least
$|s|\,\varepsilon_{\mathrm{loc}}^{A,i}(\hat{w})$. Alignment then fails in mean
and hence in the distributional form of
Theorem~\ref{thm:necessity}(i).
\end{enumerate}
\item[(b)] \emph{Autonomy.} Fix $(A,i)$ with $A \notin \mathrm{MPA}^*(i)$ and
assume regularity condition (i) of Theorem~\ref{thm:autonomy-app}, writing the
output as $\mathcal{M}_i(x_{\mathrm{PA}(i)}, u; \theta_A)$. Because $\theta_A$
is a parameter rather than a coordinate of $Z$, no fiber of $Z$ registers this
dependence. Work instead on $\tilde{Z} := Z \times \Theta_A$, with $\pi_X$ and
$\pi_C$ ignoring the $\Theta_A$ coordinate and $u$ independent of $\theta_A$.
Take $X_A$ as the intervention target. Suppose the attainable
post-intervention fiber in $\tilde{Z}$ over $(a,c)$ contains
states $(z,\theta^{(1)})$ and $(z,\theta^{(2)})$ with the same
$Z$-coordinate and
\[
\theta^{(1)}-\theta^{(2)}=s\hat{v},
\qquad s\neq0,\quad \|\hat{v}\|=1.
\]
Their outputs are
$\mathcal{M}_i(x_{\mathrm{PA}(i)}(z),u;\theta^{(j)})$.
Assume that the segment joining $\theta^{(2)}$ to
$\theta^{(1)}$ lies in $\Theta_A$. Then (a1) and (a2) hold with $(\theta_A, \hat{v},
\varepsilon_{\mathrm{aut}}^{A,i}(\hat{v}))$ in place of $(x_A, \hat{w},
\varepsilon_{\mathrm{loc}}^{A,i}(\hat{w}))$, using condition (ii) of
Theorem~\ref{thm:autonomy-app} and the corresponding common-sign condition
(iii$^{+}$) for $\partial_{\theta_A}\mathcal{M}_i$, and with
$x_{\mathrm{PA}(i)}$ the parent value of the common post-intervention state $z$.
\end{enumerate}
Condition (iii$^{+}$) follows from the nonvanishing-sensitivity bound when the sensitivity does not depend
on $u$, as under additive noise.
\end{corollary}

\begin{proof}
(a1) Fix $u$ outside a $P_{U_{\mathrm{sys}}}$-null set and define
$h(t) := \mathcal{M}_i(x, x_A^{(2)} + t s\hat{w}, u)$ for $t \in [0,1]$.
Because $S$ lies in the domain, continuous differentiability makes $h$ a $C^1$ function
with
$h'(t) = s\,\langle \partial_{X_A}\mathcal{M}_i(x, x_A^{(2)} + t s\hat{w}, u),
\hat{w}\rangle$, and the nonvanishing-sensitivity bound gives
$|h'(t)| \ge |s|\,\varepsilon_{\mathrm{loc}}^{A,i}(\hat{w}) > 0$ on $[0,1]$.
Since $h'$ is continuous and never vanishes, it has a single sign on $[0,1]$
by the intermediate value theorem, so that
\[
|h(1) - h(0)| = \Bigl|\int_0^1 h'(t)\,\mathrm{d}t\Bigr|
= \int_0^1 |h'(t)|\,\mathrm{d}t
\;\ge\; |s|\,\varepsilon_{\mathrm{loc}}^{A,i}(\hat{w}).
\]
The values $h(1)$ and $h(0)$ are the outputs from the post-intervention states $z_1$ and $z_2$
at the same $u$, and $z_1, z_2 \in F^{\circ}_{x,c}$, so that the structural
condition fails.

(a2) Write $\Delta(u) := h(1) - h(0)$. By (iii$^{+}$),
\[
\zeta\,\mathrm{sign}(s)\,\Delta(u)
= \int_0^1 |s|\,\zeta\,\langle \partial_{X_A}\mathcal{M}_i(x, x_A^{(2)} + t s\hat{w}, u), \hat{w}\rangle\,\mathrm{d}t
\;\ge\; |s|\,\varepsilon_{\mathrm{loc}}^{A,i}(\hat{w})
\]
for $P_{U_{\mathrm{sys}}}$-almost every $u$, so that the differences share a sign. Because $u$
has the same distribution under both states and $\mathcal{M}_i$ is integrable,
taking expectations gives
\[
\zeta\,\mathrm{sign}(s)\,
\left(
\mathbb{E}_{U_{\mathrm{sys}}}\mathcal{M}_i(x,x_A^{(1)},U_{\mathrm{sys}})
-
\mathbb{E}_{U_{\mathrm{sys}}}\mathcal{M}_i(x,x_A^{(2)},U_{\mathrm{sys}})
\right)
\ge |s|\,\varepsilon_{\mathrm{loc}}^{A,i}(\hat{w}).
\]
The means differ, hence so do the output distributions, and the condition of
Theorem~\ref{thm:necessity}(i) fails.

(b) On $\tilde{Z}$, the output from the post-intervention state
$(z,\theta)$ is $\mathcal{M}_i(x_{\mathrm{PA}(i)},u;\theta)$,
where $x_{\mathrm{PA}(i)}$ is the parent value of $z$.
This value is common to both states because they share $z$. The arguments of (a1) and (a2) apply verbatim to
$h(t) := \mathcal{M}_i(x_{\mathrm{PA}(i)}, u; \theta^{(2)} + t s\hat{v})$, using
conditions (i) and (ii) of Theorem~\ref{thm:autonomy-app} and the
corresponding sign condition. On $\tilde{Z}$, $\theta_A$ varies across process
realizers of the same target, as in Theorem~\ref{thm:endo-autonomy-app}, which
is how an autonomy violation becomes visible to the fiber condition.

\emph{The sign condition cannot be dropped from (a2).} Let $u = \pm 1$ with
equal probability, $\mathcal{M}_i = u\,x_A$, $\hat{w} = 1$,
$x_A^{(1)} = 1$ and $x_A^{(2)} = -1$. The nonvanishing-sensitivity bound holds with
$\varepsilon_{\mathrm{loc}}^{A,i}(\hat{w}) = 1$, and the outputs $u$ and $-u$
differ by $2 = |s|\,\varepsilon_{\mathrm{loc}}^{A,i}(\hat{w})$ for every $u$.
Both outputs are nevertheless $\pm 1$ with equal probability, so that the
distributional alignment condition holds.
\end{proof}
}

{

\begin{corollary}[Output-Content Instability Under Observational Learning]
\label{cor:semantic-instability}
Assume the hypotheses of Corollary~\ref{cor:la-alignment}(a1), so that an
attainable post-intervention fiber $F^{\circ}_{x,c}$ contains states $z_1, z_2$ whose values of
$X_A$ differ by $s\hat{w}$ with $s \neq 0$. Write
$y_j(u) := \mathcal{M}_i(x, x_A^{(j)}, u)$ for the post-intervention outputs,
and let $\iota$ be a content map as in Section~\ref{app:comp-sem-la}.
\begin{enumerate}[leftmargin=*,itemsep=2pt]
\item[(a)] 
\emph{Failure of expressed-content invariance.} If $\iota(y_1(u)) \neq \iota(y_2(u))$ for
some $u$, expressed-content invariance (Definition~\ref{def:osl}) fails pathwise.
By Corollary~\ref{cor:la-alignment}(a1), this holds in particular
\begin{enumerate}[leftmargin=*,itemsep=2pt]
\item[(a1)] if $\iota$ is injective; or
\item[(a2)] if $\iota$ assigns distinct contents to outputs that differ by
more than some $\delta < |s|\,\varepsilon_{\mathrm{loc}}^{A,i}(\hat{w})$.
World-anchored content with prediction tolerance $\tau$
(Section~\ref{app:anchor}) is an instance with $\delta = 2\tau$ whenever one of
the two outputs succeeds in some candidate world, since outputs more than
$2\tau$ apart cannot both lie within $\tau$ of the same target.
\end{enumerate}
\item[(b)] \emph{No guaranteed correction by observational training.} The
observational loss depends on $\mathcal{M}_i$ only through its values on the
support of the observational distribution of $(X_{\mathrm{PA}(i)}, X_A)$. If
$(x, x_A^{(1)})$ or $(x, x_A^{(2)})$ lies outside that support, the
observational loss is unchanged by any modification of $\mathcal{M}_i$
confined to a neighbourhood of that point disjoint from the support, including
one that makes the two outputs agree. At the level of input--output mappings, without restrictions from a particular parameterization, observational loss is therefore indifferent to whether alignment holds there. If both points lie on the support and the hypotheses of Theorem~\ref{thm:locality-app} hold for their connecting segment and a parameter path through the current parameters, part (b) of that theorem gives a direction along which observational descent increases the expected magnitude of the response error between the two outputs.
\end{enumerate}

The conclusions about expressed-content invariance in part (a) also
hold for autonomy violations under Corollary~\ref{cor:la-alignment}(b),
with $\theta_A$ and $\varepsilon_{\mathrm{aut}}^{A,i}(\hat{v})$ in place
of $x_A$ and $\varepsilon_{\mathrm{loc}}^{A,i}(\hat{w})$.

 Under its own hypotheses,
including baseline correctness, Theorem~\ref{thm:autonomy-app}(b)
supplies a direction along which system observational loss decreases
while absolute mechanism output error increases.
Theorem~\ref{thm:autonomy-response-app}(b) instead supplies growth of
implementation-response error at fixed parent inputs and implementation
endpoints, under its additional sensitivity-growth assumption. 
The numerical growth concerns response error; failure of
expressed-content invariance requires the content separation in part (a).

\end{corollary}

\begin{proof}
(a) The outputs from the post-intervention states $z_1$ and $z_2$ at $u$ are $y_1(u)$ and
$y_2(u)$, and $z_1, z_2 \in F^{\circ}_{x,c}$. If their contents differ at
some $u$, the pathwise condition of Definition~\ref{def:osl} fails.

(a1) Corollary~\ref{cor:la-alignment}(a1) gives $y_1(u) \neq y_2(u)$ for
$P_{U_{\mathrm{sys}}}$-almost every $u$. An injective $\iota$ then gives
$\iota(y_1(u)) \neq \iota(y_2(u))$.

(a2) For $P_{U_{\mathrm{sys}}}$-almost every $u$, Corollary~\ref{cor:la-alignment}(a1) gives
$|y_1(u) - y_2(u)| \ge |s|\,\varepsilon_{\mathrm{loc}}^{A,i}(\hat{w}) > \delta$,
so that $\iota(y_1(u)) \neq \iota(y_2(u))$.

(b) With squared-error loss,
the observational loss is an integral against the observational distribution
of the mechanism's inputs, with $u$ drawn independently from $P_{U_{\mathrm{sys}}}$. Two
induced mappings that agree on the support of the observational distribution of
$(X_{\mathrm{PA}(i)}, X_A)$ therefore have the same loss. A point outside that
support has a neighbourhood of observational probability zero, so that
modifying $\mathcal{M}_i$ only there, for all $u$, leaves the loss unchanged.
Under the additional hypotheses stated in part (b), the on-support response-error claim is Theorem~\ref{thm:locality-app}(b). The autonomy mechanism-error claim follows from
Theorem~\ref{thm:autonomy-app}(b), under its hypotheses including baseline
correctness. The implementation-response claim follows from
Theorem~\ref{thm:autonomy-response-app}(b), under its own hypotheses
including sensitivity growth. 
The conclusion about expressed-content invariance uses the
content-separation argument in part (a).

\end{proof}

 The directional results alone do not
determine whether a full gradient-descent update worsens alignment.
Section~\ref{sec:gradient} states initial gradient-flow and full-step
conclusions under additional conditions on the learning problem;
Appendix~\ref{app:learning-conditions} supplies their proofs.
}

\clearpage
\section{Learning Conditions That Produce Computational L\&A Violations}
\label{app:learning-conditions}

This appendix supplies the proofs and further interpretation of the learning
results in Section~\ref{sec:gradient}.
We first prove the emergence and retained-improvement results and relate
their discrepancy to intervention-response error. We then prove the
autonomy-update result and discuss coupling conditions, before giving an
example of growing response to fixed implementations.
The gradient comparisons use population updates in the declared learned parameter
block, with other parameters, input distributions, and evaluation laws held fixed.

\subsection{Locality: learning from residual observational association}
\label{app:locality-emergence}

\noindent\textbf{Proof of Proposition~\ref{prop:locality-emergence}.}

\begin{proof}
Use the centering operator $C$ from Section~\ref{sec:gradient}.
At the baseline $h(P)$, the residual satisfies
$\mathbb E_{\mathrm{obs}}[r_i\mid P]=0$.
A square-integrable first-order prediction change using only $P$ has the
form $\delta(P)$, with loss derivative
\[
-2\mathbb E_{\mathrm{obs}}[r_i\delta(P)]
=-2\mathbb E_{\mathrm{obs}}\!\left[
\delta(P)\mathbb E_{\mathrm{obs}}[r_i\mid P]\right]=0.
\]
Thus a parent-only change cannot lower the loss to first order.

Subtracting the parent-conditional average of $J$ does not change its
correlation with the residual:
\[
\begin{aligned}
\mathbb E_{\mathrm{obs}}[r_iJ]
&=\mathbb E_{\mathrm{obs}}[r_iJ_c]
+\mathbb E_{\mathrm{obs}}\!\left[
\mathbb E_{\mathrm{obs}}[r_i\mid P]
\mathbb E_{\mathrm{obs}}[J\mid P]\right]\\
&=\mathbb E_{\mathrm{obs}}[r_iJ_c]
=\mathbb E_{\mathrm{obs}}[\gamma J_c]
=a_{\mathrm{obs}}.
\end{aligned}
\]
The second equality uses the zero conditional mean of $r_i$.
For the third, average at fixed $(P,X_A)$, where $r_i$ has mean $\gamma$.
Define $b:=J_c^{\mathsf T}a_{\mathrm{obs}}$. Its observational inner product
with the additional signal is
\[
\begin{aligned}
\langle\gamma,b\rangle
&=a_{\mathrm{obs}}^{\mathsf T}
  \mathbb E_{\mathrm{obs}}[\gamma J_c]\\
&=\|a_{\mathrm{obs}}\|^2>0,
\end{aligned}
\]
where the norm on parameter vectors is Euclidean.
Hence $b$ is nonzero: the update's nonparent component aligns with
predictive signal left unexplained by $P$.

Differentiating the squared-error loss gives
$\nabla L_i(\theta_i^0)=-2a_{\mathrm{obs}}$.
The smoothness assumption supplies a local gradient flow, whose initial
parameter velocity is $2a_{\mathrm{obs}}$. Therefore
\[
\theta_i(t)-\theta_i^0=2t\,a_{\mathrm{obs}}+o(t).
\]
Expanding the prediction around the baseline translates this parameter
movement into an output change:
\[
\mathcal M_i(\cdot;\theta_i(t))
=h(P)+2tJ^{\mathsf T}a_{\mathrm{obs}}+o_{L^2}(t).
\]
Here the remainder has $L^2(\rho)$ norm $o(t)$.
Applying the bounded linear operator $C$ removes $h(P)$ and replaces
$J$ by $J_c$, giving
\[
C\mathcal M_i(\cdot;\theta_i(t))=2tb+o_{L^2}(t).
\]
Taking norms yields
$\mathcal V_{\mathrm{loc}}(\theta_i(t))=2t\|b\|_{L^2(\rho)}+o(t)$.
The initial loss derivative is
\[
\left.\frac{\mathrm dL_i}{\mathrm dt}\right|_{t=0}
=(-2a_{\mathrm{obs}})^{\mathsf T}(2a_{\mathrm{obs}})
=-4\|a_{\mathrm{obs}}\|^2,
\]
which gives the claimed loss expansion.

To establish strict local growth of the discrepancy, put
$u(t):=C\mathcal M_i(\cdot;\theta_i(t))$.
Then $u(t)=2tb+o_{L^2}(t)$ and $u'(t)=2b+o_{L^2}(1)$, so
\[
\frac{\mathrm d}{\mathrm dt}\|u(t)\|_{L^2(\rho)}
=\frac{\langle u(t),u'(t)\rangle_{L^2(\rho)}}
       {\|u(t)\|_{L^2(\rho)}}
\longrightarrow 2\|b\|_{L^2(\rho)}>0.
\]
Continuity also keeps the loss derivative negative nearby.
For a single gradient step,
$\theta_i^\eta=\theta_i^0+2\eta a_{\mathrm{obs}}$.
The same first-order expansions hold with $t$ replaced by $\eta$.
\end{proof}

To connect this discrepancy to intervention responses, draw $P$ from its
observational marginal and draw $A_0,A_1$ independently from the conditional
law of $X_A$ given $P$. Assume these conditional laws exist, as for Euclidean
or discrete inputs, and let $\nu$ be the resulting fixed comparison law.
Write $Z_j:=\mathcal M_i(P,A_j;\theta_i)$.
Conditional on $P$, these predictions are independent and have the same law.
Expanding their squared difference and using conditional independence gives
\[
\begin{aligned}
\mathbb E_\nu[(Z_1-Z_0)^2\mid P]
&=\mathbb E_\nu[Z_1^2\mid P]+\mathbb E_\nu[Z_0^2\mid P]
  -2\mathbb E_\nu[Z_1Z_0\mid P]\\
&=2\mathbb E_\nu[Z_0^2\mid P]
  -2\bigl(\mathbb E_\nu[Z_0\mid P]\bigr)^2\\
&=2\mathbb E_\nu\!\left[
  \bigl(Z_0-\mathbb E_\nu[Z_0\mid P]\bigr)^2\mid P\right].
\end{aligned}
\]
The centered prediction in the last line is $C\mathcal M_i(P,A_0;\theta_i)$.
Since $(P,A_0)$ has law $\rho$, averaging over $P$ yields
\[
\mathbb E_\nu[(Z_1-Z_0)^2]
=2\|C\mathcal M_i(\cdot;\theta_i)\|_{L^2(\rho)}^2
=2\mathcal V_{\mathrm{loc}}(\theta_i)^2.
\]
Taking square roots therefore gives
\[
\left(\mathbb E_\nu\left|
\mathcal M_i(P,A_1;\theta_i)-\mathcal M_i(P,A_0;\theta_i)
\right|^2\right)^{1/2}
=\sqrt{2}\,\mathcal V_{\mathrm{loc}}(\theta_i).
\]
When these pairs describe admissible interventions in the evaluation domain,
the worldly response is zero, so this is an RMS intervention-response error. If the common worldly interventional mean is $h(P)$, then also
\[
\|\mathcal M_i(\cdot;\theta_i)-h\|_{L^2(\rho)}
\geq\mathcal V_{\mathrm{loc}}(\theta_i),
\]
because $C$ is an orthogonal projection and $Ch=0$. The same input law here
weights the evaluated intervention--context pairs. These comparisons draw
values from the observational conditional support of $X_A$ given $P$.
Behavior outside that support is not constrained by this identity;
Corollary~\ref{cor:semantic-instability}(b) describes the corresponding
limits of observational loss.

\Needspace{5\baselineskip}
\noindent\textbf{Proof of Lemma~\ref{lem:locality-gain}.}

\begin{proof}
Throughout this proof, inner products and norms use the joint observational
law of $(X_i,P,X_A)$, with
$\langle a,b\rangle=\mathbb E_{\mathrm{obs}}[ab]$.
For functions of $(P,X_A)$, they agree with those in $L^2(\rho)$.
With $r_i=X_i-h(P)$, write
\[
X_i-\mathcal M
=(X_i-h)-(\mathcal M-h)
=r_i-(\mathcal M-h).
\]
Squaring this identity gives
\[
(X_i-\mathcal M)^2
=r_i^2-2r_i(\mathcal M-h)+(\mathcal M-h)^2.
\]
Taking observational expectations yields the predictor's risk:
\[
L_i(\mathcal M)
=\|r_i\|^2-2\langle r_i,\mathcal M-h\rangle
+\|\mathcal M-h\|^2.
\]
The parent-only predictor has risk
$L_i(h)=\|r_i\|^2$.
Subtracting the predictor's risk from this baseline cancels the residual
squared term, leaving
\[
\begin{aligned}
G(\mathcal M)
&=L_i(h)-L_i(\mathcal M)\\
&=2\langle r_i,\mathcal M-h\rangle
-\|\mathcal M-h\|^2.
\end{aligned}
\]

To evaluate the two terms, split the predictor's departure from $h$ as
\[
\mathcal M-h=C\mathcal M+q(P),\qquad
q(P):=\mathbb E_{\mathrm{obs}}[\mathcal M\mid P]-h(P).
\]
Conditional centering gives
$\mathbb E_{\mathrm{obs}}[C\mathcal M\mid P]=0$.
Since $q$ depends only on $P$,
\[
\mathbb E_{\mathrm{obs}}[(C\mathcal M)q(P)]
=\mathbb E_{\mathrm{obs}}\!\left[
q(P)\,\mathbb E_{\mathrm{obs}}[C\mathcal M\mid P]\right]=0.
\]
The two components are therefore orthogonal, so
\[
\|\mathcal M-h\|^2
=\|C\mathcal M\|^2+\|q\|^2
=\mathcal V_{\mathrm{loc}}(\mathcal M)^2+\|q\|^2.
\]

For the cross term, $\mathbb E_{\mathrm{obs}}[r_i\mid P]=0$ implies
$\mathbb E_{\mathrm{obs}}[r_iq(P)]=0$.
Conditional averaging at fixed $(P,X_A)$ then gives
\[
\langle r_i,\mathcal M-h\rangle
=\langle r_i,C\mathcal M\rangle
=\langle\gamma,C\mathcal M\rangle,
\]
because $C\mathcal M$ is a function of $(P,X_A)$ and
$\mathbb E_{\mathrm{obs}}[r_i\mid P,X_A]=\gamma$.
Combining the two calculations yields the exact decomposition
\[
G(\mathcal M)
=2\langle\gamma,C\mathcal M\rangle
-\mathcal V_{\mathrm{loc}}(\mathcal M)^2-\|q\|^2.
\]
Only alignment of the predictor's nonparent variation with the additional
signal $\gamma$ can contribute positively. Both squared terms subtract
from that contribution.

Cauchy--Schwarz gives
$\langle\gamma,C\mathcal M\rangle\leq
\|\gamma\|\,\|C\mathcal M\|=g\,\mathcal V_{\mathrm{loc}}(\mathcal M)$.
Dropping the nonpositive term $-\|q\|^2$ therefore gives
\[
G(\mathcal M)
\leq 2g\,\mathcal V_{\mathrm{loc}}(\mathcal M)
-\mathcal V_{\mathrm{loc}}(\mathcal M)^2.
\]
If $g=0$, then $\gamma=0$ in $L^2(\rho)$ and the exact decomposition gives
$G(\mathcal M)=-\|\mathcal M-h\|^2\leq0$.
Thus positive gain requires $g>0$. Dropping the remaining negative square
then gives $G(\mathcal M)\leq2g\,\mathcal V_{\mathrm{loc}}(\mathcal M)$,
which rearranges to
$\mathcal V_{\mathrm{loc}}(\mathcal M)\geq G(\mathcal M)/(2g)>0$.
\end{proof}

The bound holds for any parameterization and learning rule. Retaining a fixed
positive gain requires retaining a positive locality discrepancy on this input
law. The gradient flow of Proposition~\ref{prop:locality-emergence} retains
its early gain for as long as it exists, since its loss is nonincreasing.
The discrepancy persists, although its magnitude may vary.

Under the same conditions of admissibility and worldly invariance used for
the comparison law $\nu$, positive gain also gives the response-error bound
\[
\left(\mathbb E_\nu
|\mathcal M(P,A_1)-\mathcal M(P,A_0)|^2\right)^{1/2}
\geq \frac{G(\mathcal M)}{\sqrt{2}\,g}.
\]

\paragraph{Connection to the amplification theorem.}
Proposition~\ref{prop:locality-emergence} establishes initial growth of an
averaged locality discrepancy. Theorem~\ref{thm:locality-app} additionally
requires nonvanishing sensitivity whose magnitude increases uniformly along
a specified intervention segment. These stronger conditions require separate
verification.

Residual association must be accessible to the parameter gradient. For example,
$\mathcal M_i=h(P)+\beta^2X_A$, learning only $\beta$, can express dependence
on $X_A$ but has zero parameter derivative at $\beta=0$.
The nonzero-gradient assumption distinguishes
such a stationary parameterization from one in which observational learning
initiates the violation.

\subsection{Autonomy: improving other predictions damages a correct mechanism}
\label{app:autonomy-learning-conditions}

\noindent\textbf{Proof of Proposition~\ref{prop:autonomy-gradient-damage}.}

\begin{proof}
At the baseline, $\nabla_{\theta_A}L_i(\theta_A^0)=0$, so the system gradient
is $\nabla_{\theta_A}L_{\mathrm{obs}}^{\mathrm{sys}}(\theta_A^0)=g_A$.
Gradient flow therefore has initial parameter velocity $-g_A$, giving
\[
\theta_A(t)-\theta_A^0=-t\,g_A+o(t).
\]
The chain rule translates this movement into the initial output velocity:
\[
\left.\frac{\mathrm d}{\mathrm dt}
\mathcal M_i(\cdot;\theta_A(t))\right|_{t=0}
=J_A^{\mathsf T}(-g_A)=-J_A^{\mathsf T}g_A.
\]
Since the baseline equals $f_i^*$, the output change is also the error.
The $L^2(Q)$ expansion is consequently
\[
\mathcal M_i(\cdot;\theta_A(t))-f_i^*
=-tJ_A^{\mathsf T}g_A+o_{L^2}(t).
\]

To compute the size of this error, expand the squared norm of its initial
velocity:
\[
\begin{aligned}
\|J_A^{\mathsf T}g_A\|_{L^2(Q)}^2
&=\mathbb E_Q[(J_A^{\mathsf T}g_A)^2]\\
&=\mathbb E_Q[g_A^{\mathsf T}J_AJ_A^{\mathsf T}g_A]\\
&=g_A^{\mathsf T}\mathbb E_Q[J_AJ_A^{\mathsf T}]g_A\\
&=g_A^{\mathsf T}G_Ag_A>0.
\end{aligned}
\]
Thus the coupling condition says that the observational update produces
nonzero output velocity under the evaluation law $Q$.
For $t>0$, taking norms in the error expansion gives
\[
\|\mathcal M_i(\cdot;\theta_A(t))-f_i^*\|_{L^2(Q)}
=t\sqrt{g_A^{\mathsf T}G_Ag_A}+o(t).
\]
Strict local growth follows by the norm-derivative argument used in the
proof of Proposition~\ref{prop:locality-emergence}.

The initial system-loss derivative is
\[
\left.\frac{\mathrm dL_{\mathrm{obs}}^{\mathrm{sys}}}{\mathrm dt}\right|_{t=0}
=g_A^{\mathsf T}(-g_A)=-\|g_A\|^2<0.
\]
Continuity preserves its negativity nearby.
The gradient step $\theta_A^\eta=\theta_A^0-\eta g_A$ gives the same
first-order expansions with $t$ replaced by $\eta$.
\end{proof}

\paragraph{Interpreting the coupling condition.}
Positivity concerns the actual update direction.
Parameter dependence alone is insufficient: $\mathcal M_i=\theta_1+\theta_2$,
for example, is unchanged along $(1,-1)$.

In a scalar block, nonzero mechanism sensitivity and a nonzero observational
gradient have a nonzero product:
\[
\left.\frac{\mathrm d\mathcal M_i}{\mathrm dt}\right|_{t=0}
=-\partial_{\theta_A}\mathcal M_i\,g_A.
\]
Under the uniform sensitivity and baseline conditions of
Theorem~\ref{thm:autonomy-app}, the direction
$\hat v=-\operatorname{sign}(g_A)$ therefore has both the required spurious
sensitivity and negative system-loss derivative. This supplies that theorem's
observational-pressure condition through improvement of other predictions.

For a multidimensional block, positive definiteness of $G_A$ is a sufficient
structural condition: every nonzero update changes the evaluated mechanism
outputs. More generally, positive definiteness on a specified subspace containing
the update is sufficient. The proposition requires only that the actual update
affect the output, as checked by $g_A^{\mathsf T}G_Ag_A>0$.
An existing spurious sensitivity suffices; the sensitivity itself need not grow.

\subsection{Autonomy: learning strengthens response to fixed implementations}
\label{app:autonomy-amplification-example}

Here the two implementations remain fixed. Learning changes how strongly
the outcome mechanism responds to their differences, even though their worldly
outcomes agree. 

Fix $A\notin\mathrm{MPA}^*(i)$, and let two implementations of the same target and context
have fixed scalar parameters $\vartheta^0\ne\vartheta^1$. Assume they remain
admissible throughout learning, their worldly outcomes agree, their model
parent inputs are the same fixed $p$, and their
mechanism-induced mapping is
\[
\mathcal M_i(p;\vartheta,\lambda)=h_i(p)+\lambda\vartheta.
\]
The implementation parameter $\vartheta$ stays fixed within each comparison
run. The disjoint learned parameter $\lambda$ controls sensitivity to it.
The true output is $h_i(p)$, so at $\lambda=0$ both implementations give
the correct prediction. For $\lambda\geq0$, their spurious response difference is
\[
V_{\mathrm{aut}}=|\vartheta^1-\vartheta^0|\lambda.
\]
Because the implementation difference is fixed, increasing $\lambda$ directly
increases this error.

Training can reward this increase when $\lambda$ is shared with another
prediction. Let that prediction's output be $\lambda$ and its target $d>0$.
Training uses one
fixed implementation $\bar\vartheta$, giving the system loss
\[
L_{\mathrm{obs}}^{\mathrm{sys}}(\lambda)
=(\lambda-d)^2+\bar\vartheta^{\,2}\lambda^2.
\]
The first term rewards moving $\lambda$ toward $d$; the second measures the
resulting error in mechanism $i$ under the training implementation.
Gradient flow balances these pressures:
\[
\dot\lambda=2\bigl[d-(1+\bar\vartheta^{\,2})\lambda\bigr].
\]
Starting at zero, $\lambda$ increases toward $d/(1+\bar\vartheta^{\,2})$
while system loss decreases and the spurious implementation response grows.

At initialization, sensitivity to $\vartheta$ is zero, so the amplification
theorem's nonvanishing-sensitivity condition does not yet hold. Once learning
makes it positive, the theorem applies locally. At any
$\lambda_0\in(0,d/(1+\bar\vartheta^{\,2}))$, take the learning path
$\lambda=\lambda_0+\phi$. On a sufficiently small neighborhood, directional
sensitivity to $\vartheta$ is at least $\lambda_0/2$, its magnitude increases
at rate one, and the system-loss derivative is negative. The linear mechanism
also satisfies the segment and domination conditions. With the stipulated
admissibility and common inputs, this verifies the hypotheses of
Theorem~\ref{thm:autonomy-response-app} even though training observes only one
implementation. The construction demonstrates how the theorem's assumptions
can arise; it does not establish that sensitivity generally grows during training.

\paragraph{Scope of the learning results.}

 Residual association can reward a new nonparent dependence;
improvement in other predictions can act through an existing parameter coupling
or strengthen the response to implementation differences. The general propositions
establish initial behavior under their stated conditions. Their application to
later training stages requires the corresponding conditions to hold there as well.
Lemma~\ref{lem:locality-gain} applies throughout training: retaining an
observational advantage over the best local predictor requires retaining
nonlocal dependence.

\clearpage
\section{Proposed Tests in an Implemented Language Model}
\label{app:llm-witness-tests}

\subsection{Purpose and choice of task}

Applying the proposed Lie derivative witnesses to a learned system requires identifying
interventions that vary the assessed quantity while preserving the conditions
required by the comparison. This requirement becomes especially demanding for
autonomy: an intervention must alter another mechanism's rule while preserving
the legitimate inputs to the mechanism being assessed. We therefore propose
beginning with a controlled arithmetic task in an implemented language model.
Summation provides an exactly specified mechanism-induced input--output mapping.
Pairing it with an independently specified affine computation supplies inputs
and rule parameters whose variation should preserve the sum. This setting makes
the intervention correspondence easier to formulate and test, while leaving
its computational realization open to investigation. The experiment isolates
a difficult attribution problem that a richer task, such as a medical or
linguistic application, would compound with uncertainty about the worldly
causal structure and relevant context.

The task's simplicity allows the investigation to concentrate on whether an
internal intervention changes a represented input or rule, whether other
required quantities remain fixed, and whether the model uses those
representations in producing its answer. The following nine steps propose
an operational test of computational locality and autonomy using the local
witnesses of Proposition~\ref{prop:local-witness}.

\subsection{Task specification and proposed mechanism attributions}

\paragraph{Step 1: Establish the task and baseline performance.}
Present two computations,
\[
S=a+b,\qquad R=cx+d,
\]
in short, consistently formatted prompts. Vary their inputs and affine-rule
parameters independently. Begin with small integers and unambiguous arithmetic,
and assess the two answers in separate continuations. Fix the model checkpoint,
execution settings, and evaluation domain. Confirm that the model reliably
performs both computations and responds appropriately to changes in operands
and rule parameters. This establishes the baseline against which the proposed
interventions are assessed.

A candidate partial attribution proposes that the model represents the
operands and rules and uses them through the specified dependencies to execute
the requested computation. Its summation mechanism $\mu_S$ induces the
mapping being assessed, while a second mechanism $\mu_R$ applies the affine
rule. Evidence about these operations can constrain the attribution without
yet identifying the complete algorithm for addition.

\paragraph{Step 2: Specify the invariance comparisons.}
For locality, vary the represented input $x$ while preserving $a,b$, both
rules, and the relevant context. For autonomy, vary the represented affine
rule while preserving $a,b,x$, the addition operation, and preferably the
current value of $R$. In both comparisons, the correct sum remains $a+b$.
The second computation thus supplies a nonparent input for the locality test
and a separately variable rule for the autonomy test.

\begin{center}
\small
\begin{tabular}{|p{1.6cm}|p{4.1cm}|p{6.6cm}|}
\hline
\textbf{Test} & \textbf{Intended change} & \textbf{Preserved quantities} \\
\hline
Locality & Represented value of $x$ & $a,b,c,d$, the addition operation, and relevant context \\
\hline
Autonomy & Parameters of the represented rule $R=cx+d$ & $a,b,x$, the addition operation, relevant context, and the current value of $R$ \\
\hline
\end{tabular}
\end{center}

Preserving the current affine output strengthens the autonomy comparison by
separating a change in its rule from a change in the value it currently
produces. For example, at $x=2$, the rules $R=x+2$ and $R=2x$ both give $4$,
although they disagree at other inputs. The autonomy attribution requires
that the sum's induced rule remain unchanged when the affine rule changes;
the locality attribution requires the sum to ignore variation in $x$.

\subsection{Identifying and validating intervention directions}

\paragraph{Step 3: Discover candidate sites and directions.}
Compare internal activations for paired prompts differing only in $x$ to
propose sites representing that input. For autonomy, compare prompts with
different affine rules but identical operands and current affine outputs.
Activation comparisons, probes, and circuit analyses can supply candidate
sites and intervention directions. Record the proposed role of each site
and the quantities each intervention is intended to preserve. Separate the
examples used for discovery from those used to confirm the attribution.

\paragraph{Step 4: Check the intervention's interpretation.}
At a specified processing stage, apply the intervention and branch from the
same intervened computation into separate continuations asking for $a$, $b$,
$x$, the affine rule or its parameters, $R$, and $S$. Use an unchanged
comparison branch for each question. For locality, the intended pattern is
a changed report of $x$ with stable reports of $a,b,c,d$. For autonomy, the
reports should indicate a changed affine rule with stable $a,b,x,R$.
Specify how the intervention is applied to the common computational state
before the branches diverge.

These reports serve as manipulation checks. Correct reports support preservation of the quantities being reported. They do not by themselves establish that the computation producing the sum uses the checked representations: a follow-up answer may reconstruct a value from the original prompt. The witness test therefore also requires evidence connecting those representations to the computation producing the sum.

\paragraph{Step 5: Test the computational use of the representations.}
Patch a different value into a candidate representation of $a$ or $b$ and
check whether the sum changes accordingly across held-out examples. Where
feasible, use interventions on the proposed computational paths to assess
whether summation reads the representations checked in Step~4.

For a candidate affine-rule representation, test its application at several
other $x$-values. An edit from $x+2$ to $2x$ should change the affine answer
where those rules disagree, while preserving it at $x=2$. This tests whether
the edited quantities function as parameters specifying a rule. Merely
eliciting a different verbal description of the rule would leave its use
in computation unresolved. The attribution and its supporting checks remain
fixed during the subsequent witness comparison.

\subsection{Intervention paths and witness measurements}

\paragraph{Step 6: Construct and validate an intervention path.}
Vary a candidate intervention's strength over a small interval around the
baseline, producing states
\[
z(t)=\gamma(t),\qquad \gamma(0)=z_0.
\]
Here $z$ denotes the system's computational state at the selected stage,
including the components retained for continuation. At each tested strength,
repeat the checks on the changed and preserved quantities. For locality,
seek a path changing represented $x$ at fixed operands and addition rule.

For autonomy, a useful family of affine rules is
\[
c(t)=c_0+t,\qquad d(t)=d_0-x_0t.
\]
At the evaluated input $x_0$, these rules satisfy
\[
c(t)x_0+d(t)=c_0x_0+d_0,
\]
so their current output remains fixed. The internal intervention must realize
this coordinated rule change while preserving the sum's legitimate inputs.
The corresponding parameter-space direction is
$\partial_c-x_0\partial_d$; its computational realization must be established
through the proposed attribution and checks.

Specify the intervention interface and the states attainable through it.
An admissible path must preserve the required target and context throughout
the comparison. Interpolating between two activation vectors supplies a
candidate path whose intermediate interpretation requires validation. If
only the endpoints support the intended comparison, their output disagreement
can establish a finite invariance failure; a continuous witness additionally
requires a justified path. This distinction implements the restriction to
attainable fiber directions in Proposition~\ref{prop:local-witness}.

\paragraph{Step 7: Measure the output response.}
For a fixed summation question, define
\[
g(z):=P_{\mathrm{model}}(\text{answer}=a+b\mid z).
\]
The conditional law here is induced by continuing the actual model from
the selected stage, with the summation question fixed. Define the answer
event precisely, including formatting and termination. Score its complete
sequence probability, or use a specified categorical readout with verified
single-token labels. This probability is an expected indicator of an output
event, of the form $g_\chi$.

Where the continuation probability and intervention path are differentiable,
calculate
\[
\mathcal L_Vg(z_0)
=\left.\frac{d}{dt}g(\gamma(t))\right|_{t=0},
\qquad V=\gamma'(0).
\]
For a straight activation path of direction $v$, the derivative is
$\nabla_z g(z_0)\cdot v$. Differentiate through the original model and
check the result against actual small interventions using
\[
\frac{g(\gamma(\epsilon))-g(\gamma(-\epsilon))}{2\epsilon}
\]
at several perturbation sizes. Check numerical precision and local agreement
between the derivative and measured responses. Output probabilities can
reveal a change even when the most probable answer remains the same.

\paragraph{Step 8: Apply the witness criteria.}
A locality witness is a nonzero derivative along a supported change in
represented $x$, with the sum's legitimate inputs, operation, and comparison
conditions preserved. An autonomy witness is a nonzero derivative along a
supported change to the affine rule under the corresponding fixed-input
comparison. Step~5's evidence that the edited quantities specify that rule
is essential to the autonomy interpretation.

Under these intervention correspondences, the task requires an unchanged
sum, so a nonzero derivative of its answer probability witnesses failure
of the required distributional invariance. The conclusion is conditional
on the mechanism attribution and validated intervention domain. If an edit
also changes a legitimate operand, the changed sum probability must instead
be assessed under that altered comparison. A zero derivative for this one
answer event or direction leaves other events and directions to be tested.

\subsection{Confirmation and interpretation}

\paragraph{Step 9: Confirm the result on held-out comparisons.}
Freeze the sites, directions, manipulation checks, and output measure before
testing new operand values and prompt formulations. Include genuine operand
changes that should alter the sum, similarly sized perturbations at unrelated
sites, and restoration of the original state. Check for changes in formatting
or general arithmetic performance that could explain the measured response.
For autonomy, retain the rule-application checks at inputs away from the
point where the affine outputs agree.

Report individual confirmed cases and their frequency over a declared
evaluation measure. Quantify variation across cases separately from numerical
uncertainty in the derivative calculation, and report the tolerances used
to assess preservation of the required quantities. Each supported nonzero
derivative supplies a local witness at its comparison; its frequency addresses
the breadth of failure over the evaluation domain.

The report should distinguish output sensitivity to an activation edit,
failure of a required functional invariance under an established intervention
correspondence, and evidence for the proposed computational organization.
Beginning with locality allows its operand checks to be reused in the more
demanding autonomy experiment. Extending the protocol to medicine would then
require establishing the additional worldly causal structure, relevant
context, and intervention correspondences for that domain.

\clearpage
{
\section{Notation and Conventions}
\label{app:notation}

Symbols used only within individual results are defined locally. Those definitions govern their scope. A star marks a worldly or correct reference. The subscript $\mathrm{sys}$ identifies system exogenous conditions. Probability laws describe variation in exogenous conditions and resulting outputs rather than acting as causal mechanisms.

\subsection*{World, model, and computational realization}
\begin{description}[style=nextline,leftmargin=1.5em,labelindent=0pt,font=\normalfont,itemsep=5pt]
\item[$M$, $M^*$; $\mu_i$, $\mathcal{M}_i$, $f_i^*$]
A candidate model and the true model; an attributed computational mechanism $\mu_i$, its mechanism-induced mapping $\mathcal M_i$, and the worldly mechanism mapping $f_i^*$. A mechanism specifies operations and dependencies; its induced mapping specifies the resulting outputs for its arguments. Distinct mechanisms may induce the same mapping. The true model belongs to the learner's hypothesis class only where this is assumed. The worldly mechanism mapping $f_i^*$ is distinct from the predicted mean $f(e,c)$.

\item[$\Phi$, $P_{\Phi}(\cdot\mid z)$]
The complete computational map $\Phi:Z\times\mathcal{U}_{\mathrm{sys}}\to Z$
and its induced state-transition law given internal state $z$. The map retains dependence on the system's actual exogenous conditions. The law averages over their variation
(Definition~\ref{def:alignment}).

\item[$\mathrm{In}(i)$, $\mathrm{PA}(i)$, $\mathrm{MPA}^*(i)$]
Computational inputs, worldly state-variable parents, and legitimate worldly
mechanism-parameter dependencies. Locality concerns dependence on inputs outside
$\mathrm{PA}(i)$; autonomy concerns parameter changes outside
$\mathrm{MPA}^*(i)$. By convention $i\in\mathrm{MPA}^*(i)$, and in the ordinary
modular case $\mathrm{MPA}^*(i)=\{i\}$ (Section~\ref{sec:lap}).

\item[$U_{\mathrm{sys}}$, $u_{\mathrm{sys}}$; $U^*$, $u^*$]
System and world exogenous conditions, with lowercase letters denoting fixed values. Their laws $P_{U_{\mathrm{sys}}}$ and $P_{U^*}$ describe variation across the evaluated cases. Their spaces need not be identified.
In response and policy comparisons, $u$ abbreviates $u_{\mathrm{sys}}$.
In Theorem~\ref{thm:autonomy-app}, $u=(u_{\mathrm{sys}},u^*)$ instead indexes the
specified joint evaluation used to assess baseline correctness.

\item[$\theta_A$, $\eta$, $\phi$]
Implementation parameters associated with mechanism $A$, a separate learned
parameter block, and displacement along a specified learning path. In
Theorem~\ref{thm:autonomy-response-app}, learning varies $\eta$ while the compared
$\theta_A$ endpoints stay fixed. In Theorem~\ref{thm:autonomy-app}, learning
varies $\theta_A$ itself. These directional comparisons do not describe every
possible training update.
\end{description}

\subsection*{Interventions, realizers, and output levels}
\begin{description}[style=nextline,leftmargin=1.5em,labelindent=0pt,font=\normalfont,itemsep=5pt]
\item[$\pi_X$, $\pi_Y$, $\pi_C$ versus $\pi$, $\pi_A$]
Projections of internal states onto task target, output, and relevant context,
versus intervention-generating policies. The joint projection is
$\pi_{X,C}=(\pi_X,\pi_C)$. Task variable $X$ is distinct in scope from $X$
denoting all endogenous variables in expressions for joint model distributions.

\item[$e$, $R$, $r$]
An intervention specification, a process realizer, and an eligible realization
in Definition~\ref{def:int-comp}. For a fixed target, $r=(R,z)$ specifies the
process and starting state; $e$ may also specify a policy. Elsewhere $r$ is a
scalar policy displacement and $r_t$ an occurrence probability.
$\mathrm{do}_R(X=x)$ specifies the implementing process;
$\mathrm{do}(X=x)$ has a common worldly target under world realizer invariance.

\item[$F_{x,c}$, $F^\circ_{x,c}$, $\tilde Z$]
All internal states with target $x$ and context $c$, the attainable
post-intervention subset, and the extended space $\tilde Z=Z\times\Theta_A$
used for implementation-parameter comparisons. The circle denotes
attainability, not topological interior. States in a fiber are constitutive
realizers; $R$ denotes a process realizer.

\item[$I_x^R$]
The internal-state transformation induced by process $R$, setting the target
to $x$ and preserving relevant context on its eligible domain
(Definition~\ref{def:alignment}). The functions $y_x$ and $K_x$ below are
evaluated at already-intervened states; they do not apply $I_x^R$ a second time.

\item[$\widehat Y_{e,r}(c,u_{\mathrm{sys}})$, $Y^{*,e}(c,u^*)$]
The system output under eligible realization $r$, and the worldly outcome under
specification $e$. Correct means and the two required kinds of structural
agreement are separate conditions in Definition~\ref{def:int-comp}.
\end{description}

The output notation distinguishes individual values, laws, and means:
\begin{center}
\small
\begin{tabular}{|p{4.4cm}|p{8.4cm}|}
\hline
\textbf{Notation} & \textbf{Meaning} \\
\hline
$y_x(z,u_{\mathrm{sys}})$ & Computed output $\pi_Y(\Phi(z,u_{\mathrm{sys}}))$ at fixed system exogenous conditions. \\
\hline
$K_x(z)$ & Output law $(\pi_Y)_*P_{\Phi}(\cdot\mid z)$. \\
\hline
$\psi(x,c,u_{\mathrm{sys}})$ & Common output map supplied by structural alignment. \\
\hline
$\Psi(x,c)$ & Common output law supplied by distributional alignment. \\
\hline
$f(e,c)$ & Predicted mean; $f(x,c)$ and $f(\pi,c)$ are target and policy specializations. \\
\hline
$f_M(e,c)$ & Interventional mean function induced by candidate
model $M$ under specification $e$. \\
\hline
\end{tabular}
\end{center}
For a measurable map $T$, $(T)_*P$ denotes the law of $T(U)$ when $U$ has law
$P$. A common output map or law expresses alignment, not worldly correctness.

\subsection*{Error quantities, hypothesis classes, and overloaded symbols}
\begin{description}[style=nextline,leftmargin=1.5em,labelindent=0pt,font=\normalfont,itemsep=5pt]
\item[$\rho$ in $L^2(\rho)$]
The observational law of $(P,X_A)$ in Section~\ref{sec:gradient}, where $P=X_{\mathrm{PA}(i)}$. It differs from the intermediate-state map $\rho(z,u)$ in Appendix~\ref{app:comp-sem-la}.

\item[$G(\mathcal M)$]
The observational improvement $L_i(h)-L_i(\mathcal M)$ under squared-error loss, relative to the best parent-based predictor $h(P)=\mathbb E_{\mathrm{obs}}[X_i\mid P]$ (Lemma~\ref{lem:locality-gain}). This use differs from the causal graph $G=(V,E)$ and the output-generating computation $G(z_{\mathrm{rep}},v,u)$.

\item[$e(\phi;p,u)$; $e_\phi^{\mathrm{aut}}(u)$, $e_\phi^{\mathrm{loc}}(u)$]
The signed departure from a correct reference in Theorem~\ref{thm:autonomy-app},
versus response errors comparing two implementations or intervention values.
The latter subtract the worldly change, which is zero under their assumptions;
neither endpoint prediction need be correct. These errors differ from an
intervention specification $e$.

\item[$V_{\mathrm{aut}}(\phi)$, $V(\phi)$]
Expected absolute response errors under a fixed common law for system exogenous conditions. They average the magnitudes of discrepancies compared at fixed exogenous values. They are not absolute differences of means or distances between output distributions. Here $V$ is unrelated to the vertex
set in $G=(V,E)$.

\item[$\varepsilon_{\mathrm{aut}}^{A,i}$, $\varepsilon_{\mathrm{loc}}^{A,i}$]
Positive lower bounds on directional sensitivity over the specified comparison
domain. They are not output-error magnitudes. By contrast, $\varepsilon_t$ in
the retention results is a conditional mean initial discrepancy.

\item[$\mathfrak E$, $\mathfrak E_{\mathrm{int}}$; $\mathcal E_{\mathcal H}(P_{\mathrm{obs}})$]
An intervention class and the domain evaluated for competence, versus the
observational equivalence class of models in $\mathcal H$.
$\mathcal B\subseteq\mathcal H$ is an inductive-bias class; $\mathcal H_t$ denotes the admitted class at inquiry stage $t$;
in Appendix~\ref{app:candidate-elimination} it is the surviving candidate
set under elimination. $\mathcal H$ denotes a fixed class, including
$\mathcal H_t$ at a specified inquiry stage. In $P_M(X\mid e)$, the bar denotes
intervention, not conditioning on an ordinary observed event.

\item[$\Delta(M^*,\mathcal H;\mu_{\mathrm{eval}})$]
The model-class approximation gap
$\inf_{M\in\mathcal H}\|f_M-f_{M^*}\|_{L^2(\mu_{\mathrm{eval}})}$, where $\mu_{\mathrm{eval}}$ weights
intervention/context pairs. It concerns mean functions; excluding $M^*$ from
$\mathcal H$ need not make it positive. Local quantities such as
$\Delta_{\mathrm{model}}$, $\Delta_{\mathrm{world}}$, and $\Delta_i(u)$ instead
denote response differences.

\item[$D_s$, $E_s$, $q_s$]
Paired-trajectory discrepancy, error against a correct reference, and an
expected-retention lower bound satisfying
$\mathbb E[D_{s+1}]\geq q_s\mathbb E[D_s]$.
A discrepancy is an error only when its reference supplies correct task behavior.
The coefficient $q_s$ differs from the directional sensitivity $q_\phi(t,u)$ in
the response-growth proofs.

\item[$\delta$, $\tau$, $\delta_t$]
In practical adequacy, $\delta$ bounds mean-prediction error and $\tau$ bounds
fixed-randomness disagreement. In Corollary~\ref{cor:semantic-instability},
$\delta$ is a content-separation threshold and $\tau$ is the prediction tolerance
used to define success in candidate worlds. In propagation, $\delta_t=D_t$ is
the initial discrepancy. Each use retains its locally specified meaning.
\end{description}

\subsection*{Semantic content and comparison conventions}
\begin{description}[style=nextline,leftmargin=1.5em,labelindent=0pt,font=\normalfont,itemsep=5pt]
\item[$\mathcal C_{\mathrm{out}}$, $\mathcal C_{\mathrm{rep}}$, $\mathcal W$]
The spaces of expressed and represented content, and the possible worldly
outcomes or conditions about which those contents make claims
(Appendix~\ref{app:content-attribution}).

\item[$\iota$, $\sigma$]
The content assignments $\iota:\mathcal Y\to\mathcal C_{\mathrm{out}}$ and
$\sigma:Z\to\mathcal C_{\mathrm{rep}}$ under the specified interpretation.
Expressed-content invariance and represented-content invariance compare
these contents across the evaluated realizations
(Definitions~\ref{def:osl} and~\ref{def:rsl}).

\item[$\rho(z,u)$, $P_\rho(\cdot\mid z)$]
The internal state at the identified stage before output generation, and
its law under the common law for system exogenous conditions. Represented content at that stage is $\sigma(\rho(z,u))$, with law $\sigma_*P_\rho(\cdot\mid z)$. This map $\rho$ differs from the observational law used in $L^2(\rho)$.

\item[$G(z_{\mathrm{rep}},v,u)$]
The schematic output-generating computation, with
$z_{\mathrm{rep}}=\rho(z,u)$ and other relevant inputs $v$.
Here $G$ denotes a computation, rather than the causal graph $G=(V,E)$ or the observational improvement $G(\mathcal M)$.

\item[$\mathrm{R}_{x,c}$]
The specified relation between expressed and represented content for
target $x$ and context $c$. Semantic faithfulness requires
$\mathrm{R}_{x,c}(\iota(y_x(z,u)),\sigma(\rho(z,u)))$ on every evaluated
run. Causal use and worldly correctness are
assessed separately.

\item[$p_{x,c}$]
A map from represented to expressed content under the additional
assumption that $\mathrm{R}_{x,c}(a,b)$ implies $a=p_{x,c}(b)$.

Under semantic faithfulness, this restriction makes represented-content invariance imply expressed-content invariance pathwise, and also in distribution when $p_{x,c}$ is measurable (Lemma~\ref{lem:sem-relations}).
\end{description}

Pathwise comparisons fix exogenous conditions. Distributional comparisons concern laws. Mean comparisons concern expectations. For the same comparisons, pathwise
agreement implies distributional agreement, which implies mean agreement when
expectations exist; the converses need not hold.
Definition~\ref{def:int-comp} separately specifies correctness and the required
agreement across realizations and interventions. Response-growth comparisons
hold parent inputs, intervention or implementation endpoints, and the comparison
noise law fixed while learning varies the specified parameters.
}

\end{document}